\documentclass{sig-alternate}
\usepackage{hyperref}
\usepackage{graphicx}
\usepackage{rotating}
\usepackage[dvipsnames]{xcolor}  % color is sufficient

\newcommand{\bbobdatapath}{} % default output folder of rungeneric.py

\providecommand{\algfolder}{}
\providecommand{\bbobecdfcaptionallgroups}{
Empirical cumulative distribution of simulated (bootstrapped) runtimes, measured in number
         of objective function evaluations, divided by dimension (FEvals/DIM) for the $58$ targets $\{-10^{-4}, -10^{-4.2}, $ $-10^{-4.4}, -10^{-4.6}, -10^{-4.8}, -10^{-5}, 0, 10^{-5}, 10^{-4.9}, 10^{-4.8}, \dots, 10^{-0.1}, 10^0\}$ for all function groups and all dimensions. The aggregation over all 55 functions is shown in the last plot.
}
\providecommand{\bbobecdfcaptionsinglefcts}{
Empirical cumulative distribution of simulated (bootstrapped) runtimes in number
         of objective function evaluations divided by dimension (FEvals/DIM) for the $58$ targets $\{-10^{-4}, -10^{-4.2}, $ $-10^{-4.4}, -10^{-4.6}, -10^{-4.8}, -10^{-5}, 0, 10^{-5}, 10^{-4.9}, 10^{-4.8}, \dots, 10^{-0.1}, 10^0\}$ for functions $f_1$ to $f_{16}$ and all dimensions. 
}
\providecommand{\bbobpptablecaption}[1]{
                Average runtime (\aRT) to reach given targets, measured
                in number of function evaluations. For each function, the \aRT\ 
                and, in braces as dispersion measure, the half difference between 10 and 
                90\%-tile of (bootstrapped) runtimes is shown for the different
                target \Df-values as shown in the top row. 
                \#succ is the number of trials that reached the last target 
                $\hvref + 10^{-5}$.
                The median number of conducted function evaluations is additionally given in 
                \textit{italics}, if the target in the last column was never reached. 
                
}
\providecommand{\bbobppfigdimlegend}[1]{
        Scaling of runtime to reach $\hvref+10^{\#}$ with dimension;
        runtime is measured in number of $f$-evaluations and $\#$ is given in the legend;
        Lines: average runtime (\aRT);
        Cross (+): median runtime of successful runs to reach the most difficult
        target that was reached at least once (but not always);
        Cross ({\color{red}$\times$}): maximum number of
        $f$-evaluations in any trial. Notched
        boxes: interquartile range with median of simulated runs;
        All values are divided by dimension and 
        plotted as $\log_{10}$ values versus dimension. %
        Numbers above \aRT-symbols (if appearing) indicate the number of trials
        reaching the respective target. Horizontal lines mean linear scaling, slanted
        grid lines depict quadratic scaling.  
        
}

\providecommand{\bbobloglossfigurecaption}[1]{
    \aRT\ loss ratios (see Figure~\ref{tab:ERTloss} for details).  
    Each cross ({\color{blue}$+$}) represents a single function, the line
    is the geometric mean.
    
}
\providecommand{\bbobloglosstablecaption}[1]{
    \aRT\ loss ratio versus the budget in number of $f$-evaluations
    divided by dimension.
    For each given budget \FEvals, the target value \ftarget\ is computed
    as the best target $f$-value reached within the
    budget by the given algorithm.
    Shown is then the \aRT\ to reach \ftarget\ for the given algorithm
    or the budget, if the GECCO-BBOB-2009 best algorithm
    reached a better target within the budget,
    divided by the best \aRT\
    seen in GECCO-BBOB-2009 to reach \ftarget.
    Line: geometric mean. Box-Whisker error bar: 25-75\%-ile with median
    (box), 10-90\%-ile (caps), and minimum and maximum \aRT\ loss ratio
    (points). The vertical line gives the maximal number of function evaluations
    in a single trial in this function subset. See also
    Figure~\ref{fig:ERTlogloss} for results on each function subgroup.
    
}
\graphicspath{{\bbobdatapath\algfolder}}

\newcommand{\aRT}{\ensuremath{\mathrm{aRT}}}
\newcommand{\FEvals}{\ensuremath{\mathrm{FEvals}}}

\newcommand{\Df}{\ensuremath{\Delta f}}

\newcommand{\hvref}{\ensuremath{HV_\mathrm{ref}}}

\newcommand{\ftarget}{\ensuremath{f_\mathrm{t}}}

\begin{document}
%
% --- Author Metadata here ---
\crdata{TBA}
\clubpenalty=10000
\widowpenalty = 10000
% --- End of Author Metadata ---

\newcommand{\vc}[1]{\textit{\textbf{#1}}}

\title{Anytime Bi-Objective Optimization with a Hybrid Multi-Objective CMA-ES (HMO-CMA-ES)
% \titlenote{If needed}
}

% Camera-ready paper due by May 4th.


\numberofauthors{2} %  in this sample file, there are a *total*
\author{
% 1st. author
\alignauthor Ilya Loshchilov\\
       \affaddr{University of Freiburg}\\
       \affaddr{Freiburg, Germany}\\
       \email{ilya.loshchilov@gmail.com}
%% 2nd. author
\alignauthor Tobias Glasmachers\\
       \affaddr{Institut f\"{u}r Neuroinformatik,}\\
       \affaddr{Ruhr-Universit\"{a}t Bochum}\\
			\affaddr{Bochum, Germany}\\
       \email{tobias.glasmachers@ini.rub.de}
} % author
%% There's nothing stopping you putting the seventh, eighth, etc.
%% author on the opening page (as the 'third row') but we ask,
%% for aesthetic reasons that you place these 'additional authors'
%% in the \additional authors block, viz.
%\additionalauthors{Additional authors: John Smith (The Th{\o}rv{\"a}ld Group,
%email: {\texttt{jsmith@affiliation.org}}) and Julius P.~Kumquat
%(The Kumquat Consortium, email: {\texttt{jpkumquat@consortium.net}}).}
%% Just remember to make sure that the TOTAL number of authors
%% is the number that will appear on the first page PLUS the
%% number that will appear in the \additionalauthors section.

\maketitle
\begin{abstract}
We propose a multi-objective optimization algorithm aimed at achieving
good anytime performance over a wide range of problems. Performance is
assessed in terms of the hypervolume metric.
The algorithm called HMO-CMA-ES represents a hybrid of several old and
new variants of CMA-ES, complemented by BOBYQA as a warm start.
We benchmark HMO-CMA-ES on the recently introduced bi-objective problem
suite of the COCO framework (COmparing Continuous Optimizers),
consisting of 55 scalable continuous optimization problems, which is
used by the Black-Box Optimization Benchmarking (BBOB) Workshop 2016.
\end{abstract}

% Add any ACM category that you feel is needed

% Complete with anything that is needed

% Complete with anything that is needed

\section{Introduction}

The design of anytime optimizers is targeted at achieving good
performance for different budgets of computational resources, e.g.,
ranging from $n$ to $10^6 n$ function evaluations, where $n$ denotes
the dimension of the search space. At the same time the black-box
optimization paradigm mandates robustness towards problems with vastly
differing characteristics.
In this work, we followed the approach of~\cite{loshchilov2013bi} where
a set of well-performing algorithms was combined to target different
classes of problems to achieve good overall anytime performance for
single-objective optimization. Here the approach is transferred to
multi-objective optimization. This effort requires a careful selection
of algorithm components, tuning parameters, and combination strategies.

\begin{figure}
\includegraphics[width=0.5\textwidth]{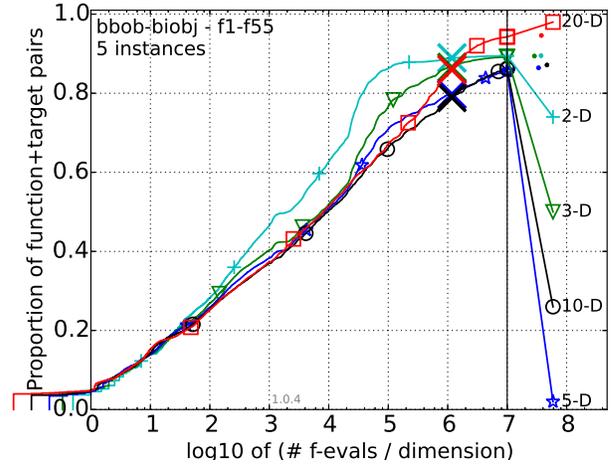}
 \caption{\label{fig:all}
  The aggregated results over all 55 functions for different problem dimensions.
  See Figure 2 for a more detailed description.
 }
\end{figure} 

The proposed Hybrid Multi-Objective Covariance Matrix Adaptation
Evolution Strategy (HMO-CMA-ES) consists of the following components:
\begin{itemize}
\item
	BOBYQA~\cite{powell2009bobyqa} on a scalarized objective function
	as a warm start,
\item
	steady-state multi-objective CMA-ES~\cite{igel2007steady} in our
	version with increasing population size (ss-MO-CMA-ES),
\item
	our version of CMA-ES with restarts on different scalarized
	objectives (restart-CMA-ES), and
\item
	generational multi-objective CMA-ES~\cite{igel2007covariance}
	in our version with restarts denoted as IPOP-MO-CMA-ES.
\end{itemize}

The bi-objective problem suite \cite{biobj2016func} of the COCO
framework consists of $55$ classes of bi-objective functions
$f_k : \mathbb{R}^n \to \mathbb{R}^2$, $k \in \{1, \dots, 55\}$,
scalable to any input space dimension $n \geq 2$. Common dimensions for
evaluation are $n = 5$ and $n = 20$. The bi-objective functions are
formed by combining all $55$ combinations of $10$ single-objective
functions, representing different challenges such as high conditioning
number and multi-modality. Five differently parameterized instances of
each problem are available for benchmarking, resulting in a total of
$275$ optimization problems. HMO-CMA-ES is evaluated on
this benchmark suite.

\section{The HMO-CMA-ES Algorithm}

In this section we describe the individual components, their final
integration in the HMO-CMA-ES algorithm, and a rationale for the
specific design choices. The source code is available at 
\url{https://sites.google.com/site/hmocmaes/}.

\subsection{BOBYQA as a Warm Start}

BOBYQA is a well-known trust-region method by Michael J.\ D.\
Powell~\cite{powell2009bobyqa}. It is well suited for uni-modal problems.
It (more exactly, its unconstrained and less advanced variant NEWUOA) is a part of the HCMA
algorithm for single-objective optimization \cite{loshchilov2013bi} that
served as inspiration for this work. Its role in HCMA is to solve simple
convex-quadratic problems at low cost in the initial phase.
In the context of multi-objective optimization we use BOBYQA for a fast
approach of the Pareto front. More specifically, we optimize a linear
aggregation function $g_\alpha(\vc{x}) = \alpha f_1(\vc{x}) + (1 - \alpha) f_2(\vc{x})$
where $f_1$ and $f_2$ are the components of the bi-criteria objective
function (the two objectives) and $\alpha \in [0, 1]$ is an aggregation
coefficient. We start BOBYQA with the initial solution
$\vc{x}_\text{init} = 0$ in the center of the suggested range for all
biobj-BBOB problems. In order to correct for a possible mis-scaling of
the objectives we normalize the components of subsequent objective
function evaluations:
\begin{align*}
	g_\alpha(\vc{x}) = \alpha \frac{f_1(\vc{x})}{f_1(\vc{x}_\text{init})} + (1 - \alpha) \frac{f_2(\vc{x})}{f_2(\vc{x}_\text{init})}
\end{align*}
The first run of BOBYQA stops after at most $5n$ function evaluations
or if BOBYQA's relative objective function improvement ratio $f_{tol}$
drops below $10^{-3}$. The remaining runs/restarts are launched with
different values of $\alpha$ in the following order:
$0.5, 0.0, 1.0, 0.95, 0.9, 0.85, \ldots, 0.05, 0.0$.
This chain represents a sweep along (convex parts of) the Pareto front,
hence the procedure yields a first rough approximation of the front.
Further restarts are conducted with a smaller stopping tolerance of
$f_{tol} = 10^{-4}$ to improve the approximation. We also decrease the
radius of the initial trust-region from $6$ (a rather global search in
$[-5,5]^n$) to $2$ (a rather local search) as each restart with a new
value of $\alpha$ is initialized in the best solution of the previous
run. These settings for BOBYQA are designed for budgets of up to $100n$
function evaluations. Most of these settings are irrelevant for
HMO-CMA-ES, where BOBYQA is run only for $10n$ function evaluations and
hence performs only few restarts within this very low budget.

\subsection{Steady-state MO-CMA-ES with Increasing Population Size}

The initial runs of BOBYQA are expected to find a better-than-random
approximates of the Pareto front. We collect all solutions generated by
BOBYQA and apply non-dominated sorting with the hypervolume metric as
secondary sorting criterion~\cite{igel2007covariance}. The five best
solutions form the initial population of the steady-state MO-CMA-ES
\cite{igel2007steady}, which is started with an initial step size of
$\sigma = \frac{1}{2}$. The population size is increased by one every
$50n$ iterations. This mechanism achieves a fast approach and a good
coverage of the Pareto front. A very similar idea was introduced
recently in \cite{glasmachers2014start}. In addition we employ a
crossover procedure with probability 10\%. It randomly selects two
solutions $\vc{x}_1$ and $\vc{x}_2$ and generates an offspring
$\vc{x}_3 \leftarrow \vc{x}_1 + a (\vc{x}_2 - \vc{x}_1)$ with blending
coefficient $a \sim \mathcal{N}(\frac{1}{2}, \frac{1}{4})$.
The offspring inherits the averaged step-size and covariance matrix
from its parents.

\subsection{Generational MO-CMA-ES with Restarts}

We use a version of generational MO-CMA-ES \cite{igel2007covariance}
where we double the population size (initially set to 10) after each
restart happening every $50n$ iterations of the algorithm. 
We denote it as IPOP-MO-CMA-ES (the idea first appeared in~\cite{commun2013}).
The initial step size is set to $\sigma = 2$.

\subsection{CMA-ES with Restarts}

We apply CMA-ES with a new restart variant to a linear aggregation
of the objective function. In each restart, a new aggregation
coefficient $\alpha$ is sampled uniformly from $[0, 1]$.
The first population consists of $\lambda = \lambda_{\min} = 50$
individuals. At the $i$-th restart, the population size is sampled as
$\lambda \leftarrow \lambda_{\min} (\lambda_{\max} / \lambda_{\min})^{b}$,
where $b$ is drawn from a uniform distribution on $[0,2]$ and
$\lambda_{\max} = \lambda_{\min} 1.02^i$. This proceeding is inspired by
BIPOP-CMA-ES but with a far smaller increase factor of $1.02$ compared
to $2$ in standard BIPOP-CMA-ES. We set the maximum number of
iterations to $100 \times 1.02^i$.

We had initially planned to use multiple BIPOP-CMA-ES instances
\cite{hansen2009benchmarking}, each optimizing a different aggregated
objective function. This approach would guarantee very good performance
for large budgets, but the initialization phase of multiple BIPOP-CMA-ES
takes a while and this would negatively impact the anytime performance
of the algorithm. The above proposal of a restart CMA-ES with random
aggregation coefficient $\alpha$ acts as a replacement.

\subsection{HMO-CMA-ES}

The proposed Hybrid Multi-objective CMA-ES algorithm is
designed to achieve best anytime performance. It has four phases, with
a different set of algorithms running. If multiple algorithms are active
at the same time then they are running in parallel, in a round-robin
fashion.

We start with BOBYQA for the first $10n$ function evaluations (phase~1).
The best solutions of BOBYQA are used to initialize the ss-MO-CMA-ES.
This algorithm runs until $1,000 n$ function evaluations (phase~2). Then
we launch restart-CMA-ES to run in parallel to ss-MO-CMA-ES (phase~3)
such that the best solution found by each run of restart-CMA-ES is
injected as a candidate solution for the next iteration of ss-MO-CMA-ES.
After $20,000 n$ function evaluations we also launch IPOP-MO-CMA-ES to
run in parallel to ss-MO-CMA-ES and restart-CMA-ES (phase~4). With a
probability of 10\% a random solution from the current population of
IPOP-MO-CMA-ES is injected into ss-MO-CMA-ES. The role of ss-MO-CMA-ES
is to fine-tune the hypervolume metric. The three algorithms are running
for $400,000 n$ function evaluations each, comprising the total budget
of $1.2 \times 10^6n$ function evaluations.

% \section{Algorithm Presentation}
%
% \section{Experimental Procedure}
%
%%%%%%%%%%%%%%%%%%%%%%%%%%%%%%%%%%%%%%%%%%%%%%%%%%%%%%%%%%%%%%%%%%%%%%%%%%%%%%%
\section{CPU Timing}
%%%%%%%%%%%%%%%%%%%%%%%%%%%%%%%%%%%%%%%%%%%%%%%%%%%%%%%%%%%%%%%%%%%%%%%%%%%%%%%
% note that the following text is just a proposal and can/should be changed to your needs:
In order to evaluate the CPU timing of the algorithm, we have run
HMO-CMA-ES with restarts on the entire bbob-biobj test suite for
$1000 n$ function evaluations. The C++ code (called from Matlab)
was run on one core of Intel(R) Core(TM) i5-4690 CPU @ 3.50GHz. 
The time per function
evaluation for dimensions 2, 3, 5, 10 and 20 equals
40, 37, 36, 39, and 51 microseconds, respectively.

%%%%%%%%%%%%%%%%%%%%%%%%%%%%%%%%%%%%%%%%%%%%%%%%%%%%%%%%%%%%%%%%%%%%%%%%%%%%%%%
\section{Results}
%%%%%%%%%%%%%%%%%%%%%%%%%%%%%%%%%%%%%%%%%%%%%%%%%%%%%%%%%%%%%%%%%%%%%%%%%%%%%%%

\begin{figure*}
\centering
\begin{tabular}{@{\hspace*{-0.018\textwidth}}l@{\hspace*{-0.02\textwidth}}l@{\hspace*{-0.02\textwidth}}l@{\hspace*{-0.02\textwidth}}l@{\hspace*{-0.02\textwidth}}}
\includegraphics[width=0.25\textwidth]{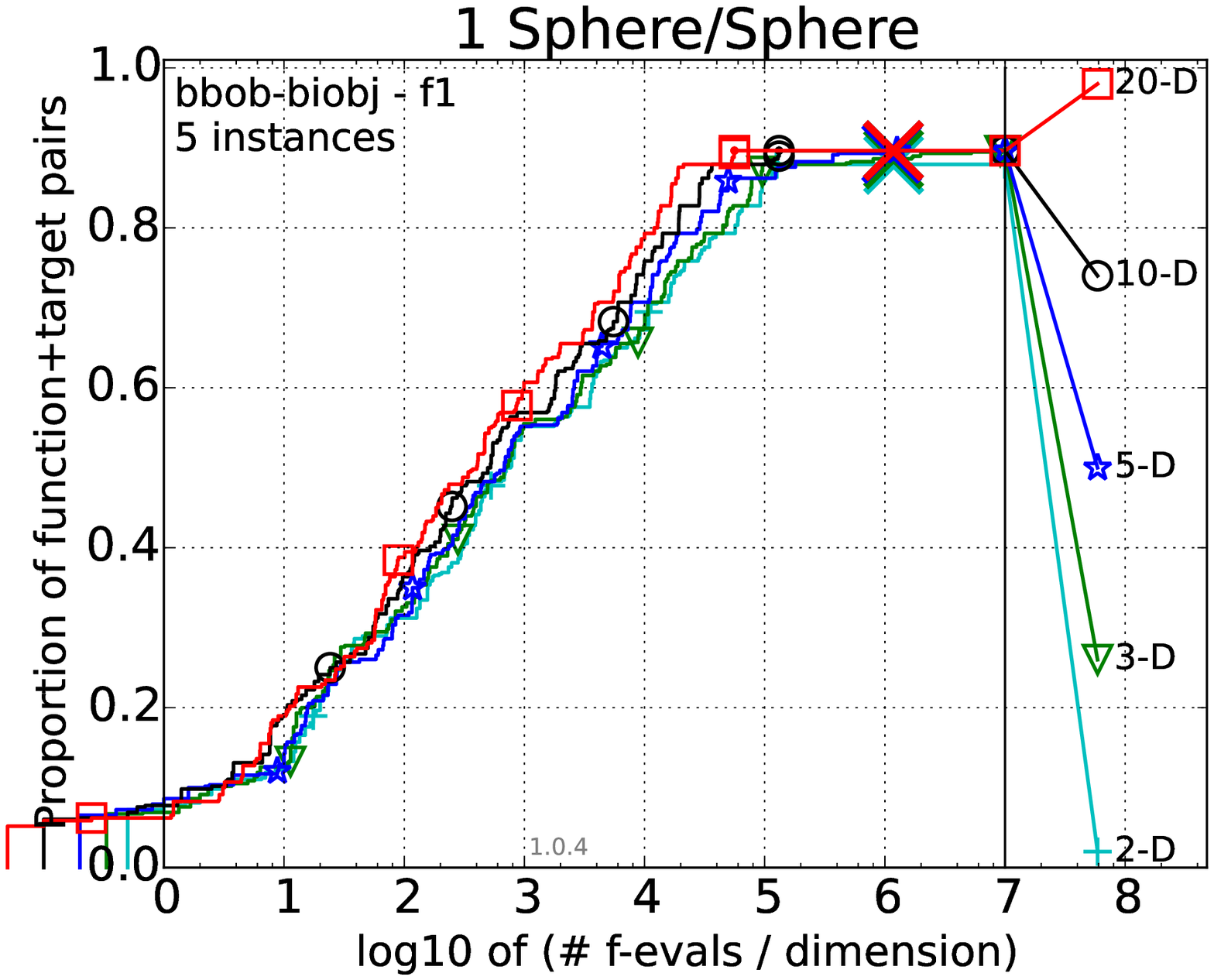}&
\includegraphics[width=0.25\textwidth]{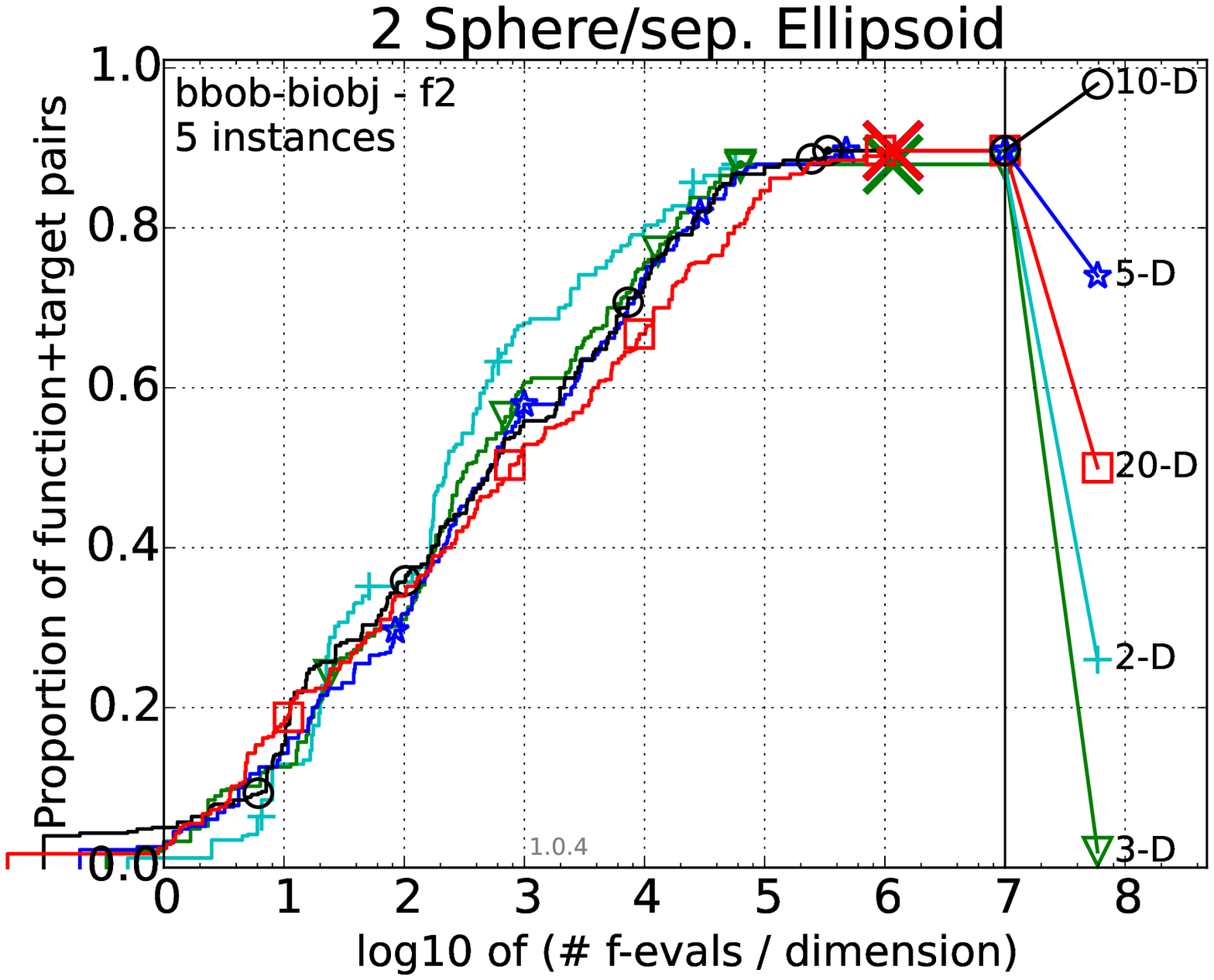}&
\includegraphics[width=0.25\textwidth]{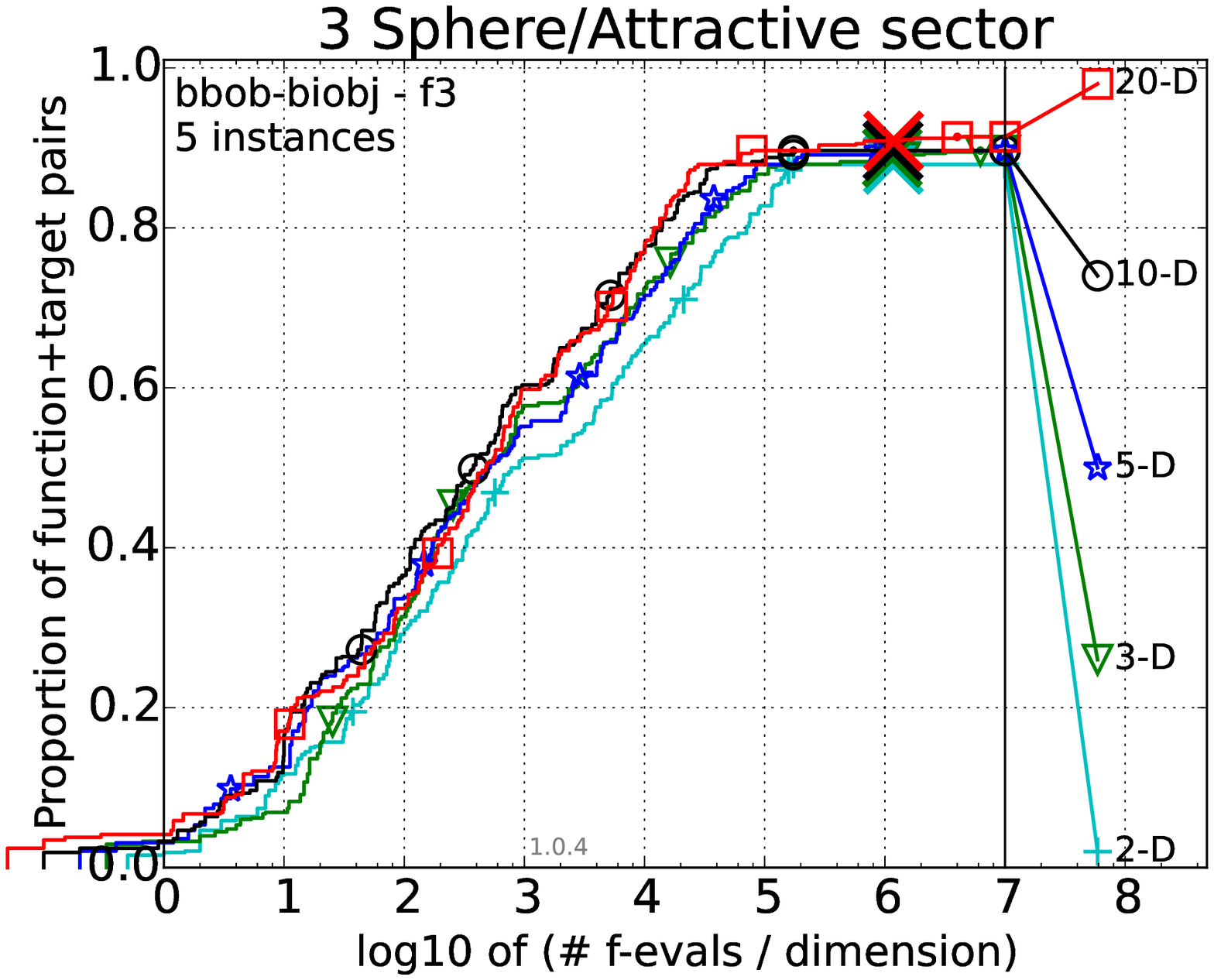}&
\includegraphics[width=0.25\textwidth]{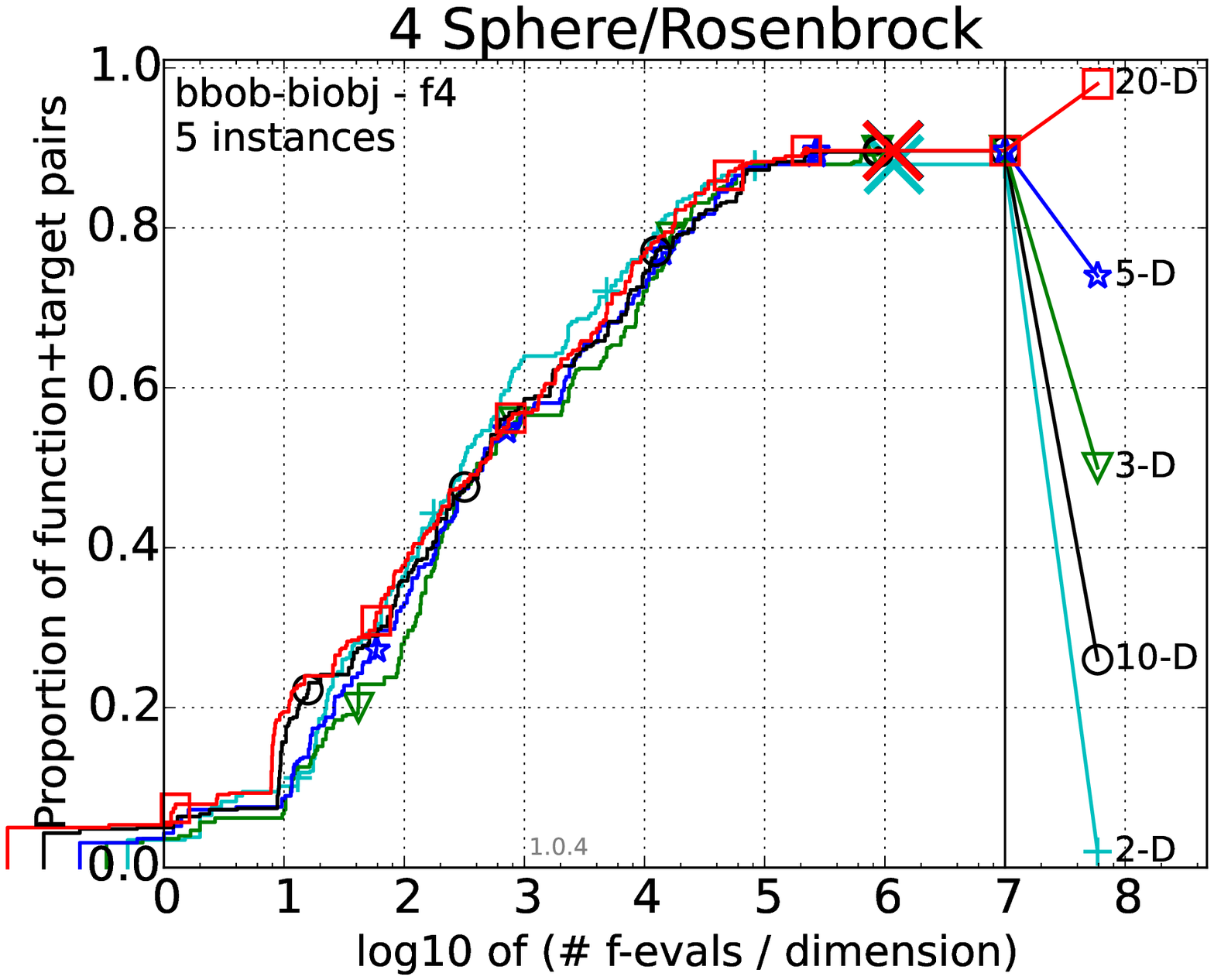}\\[-1.8ex]
\includegraphics[width=0.25\textwidth]{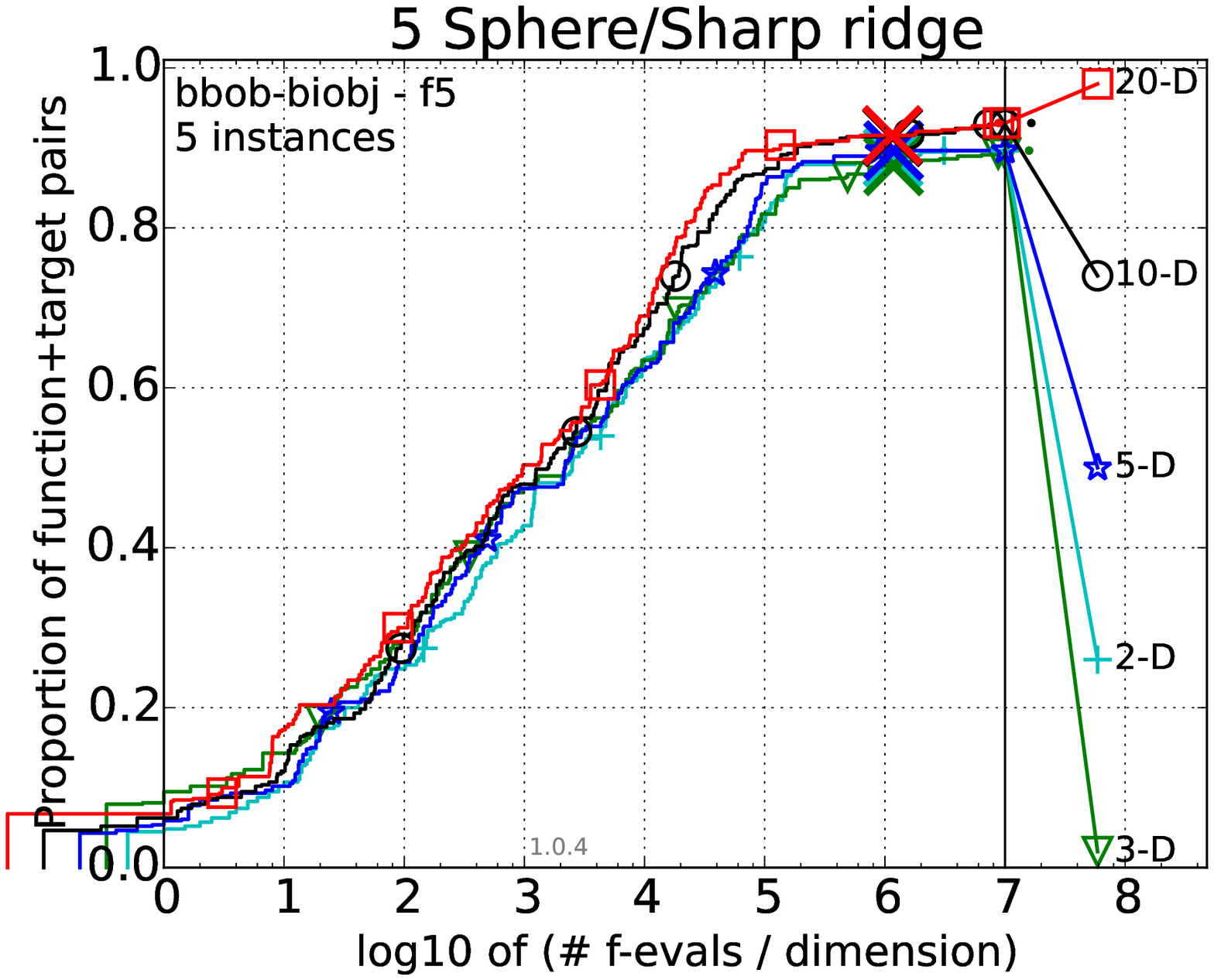}&
\includegraphics[width=0.25\textwidth]{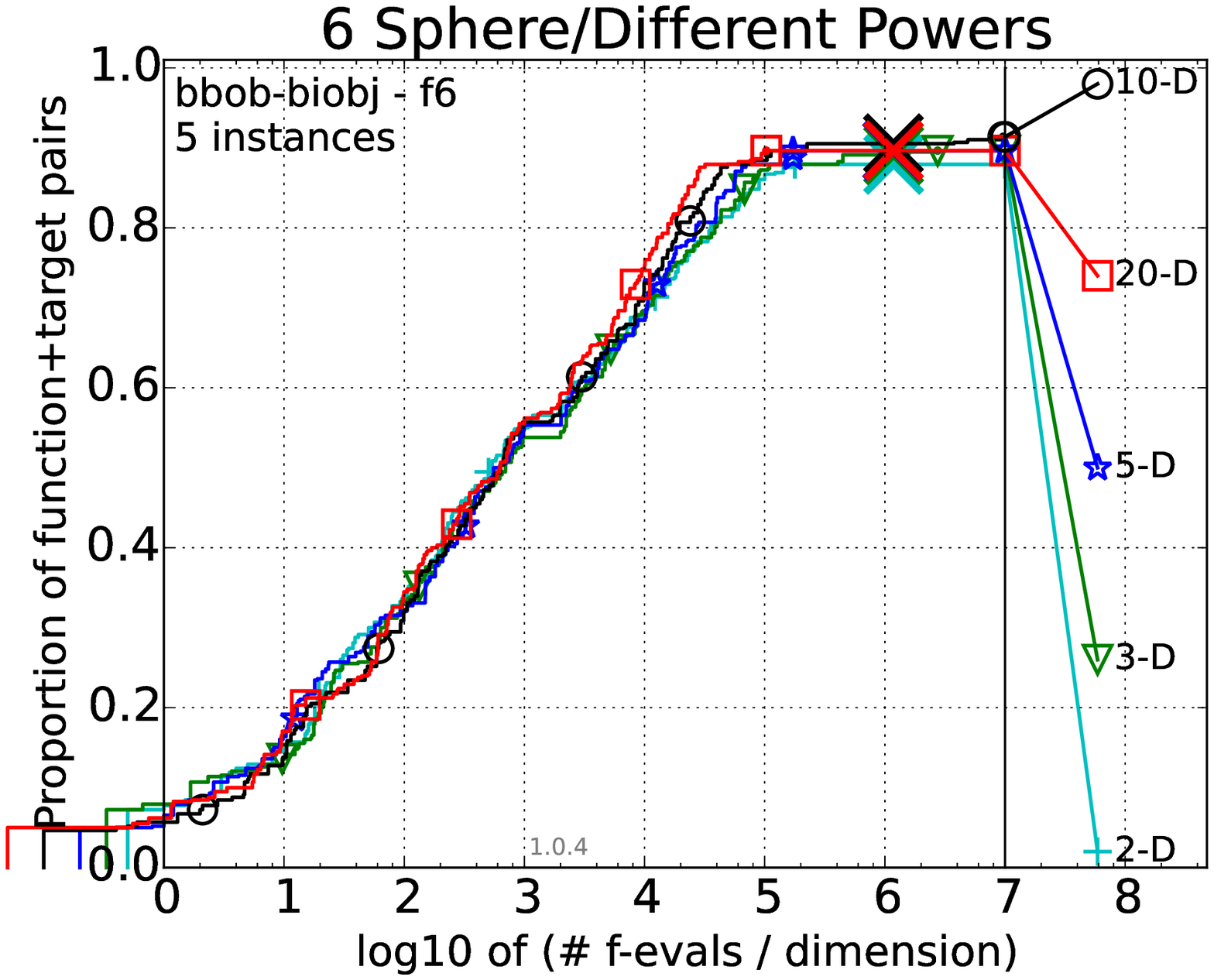}&
\includegraphics[width=0.25\textwidth]{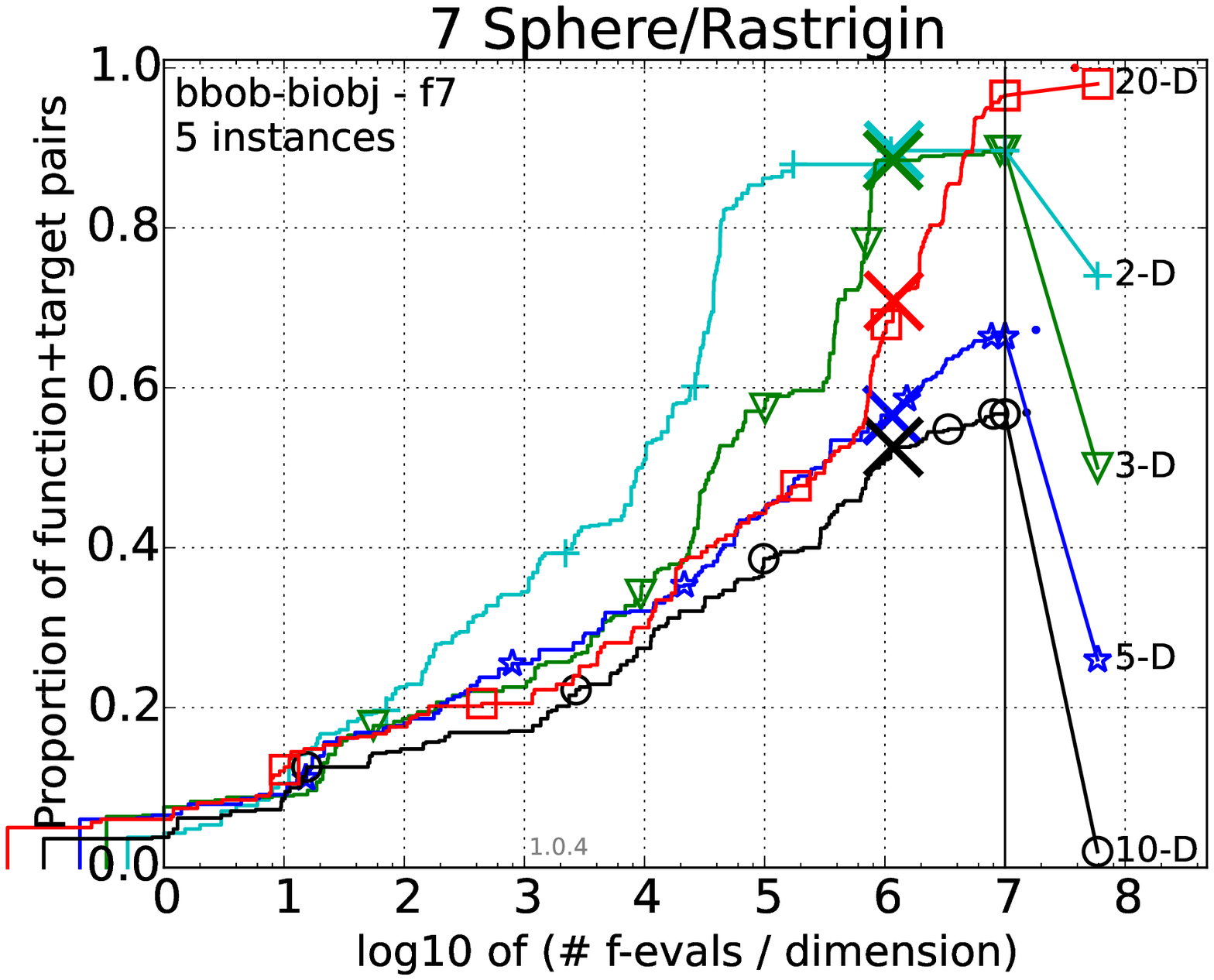}&
\includegraphics[width=0.25\textwidth]{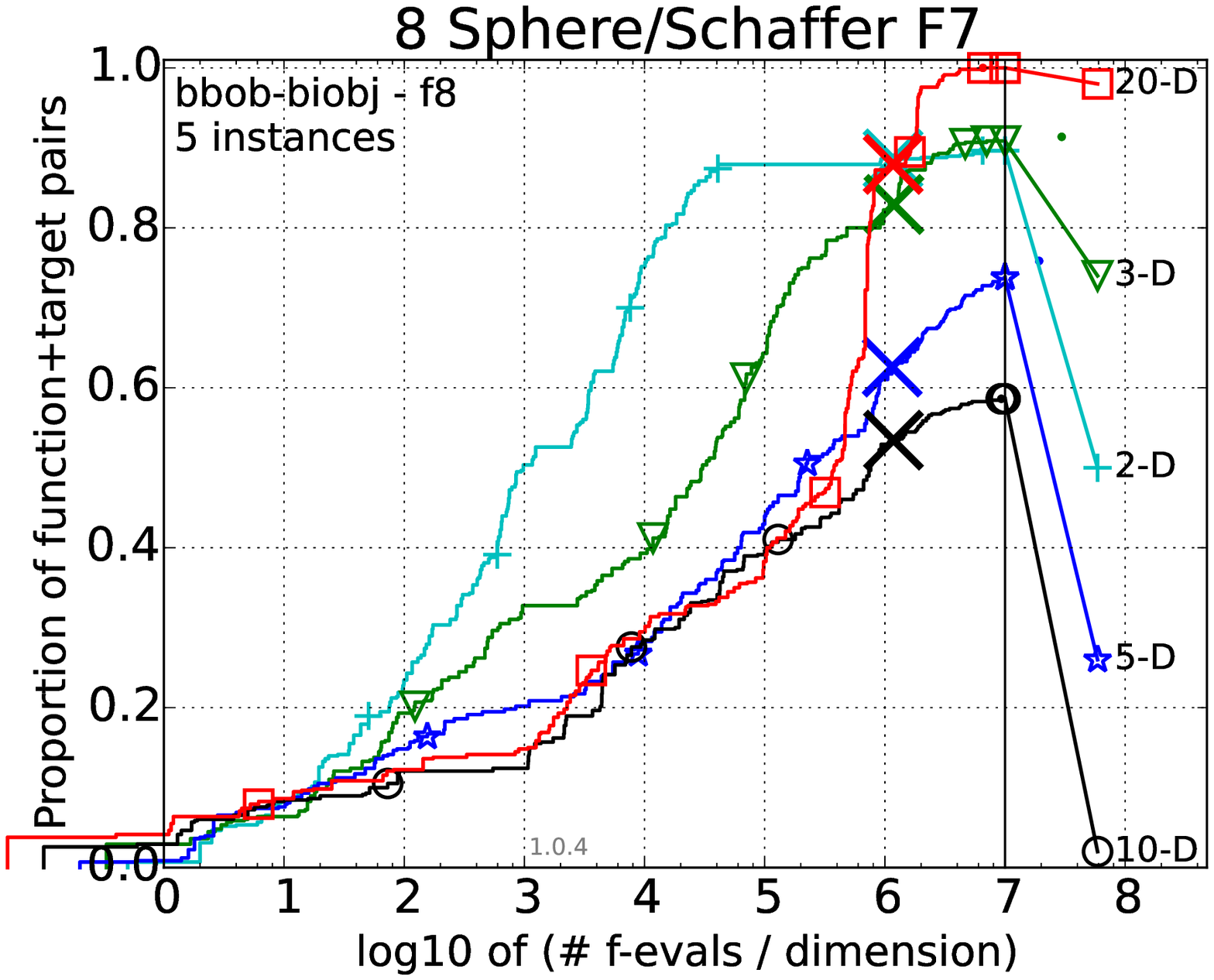}\\[-1.8ex]
\includegraphics[width=0.25\textwidth]{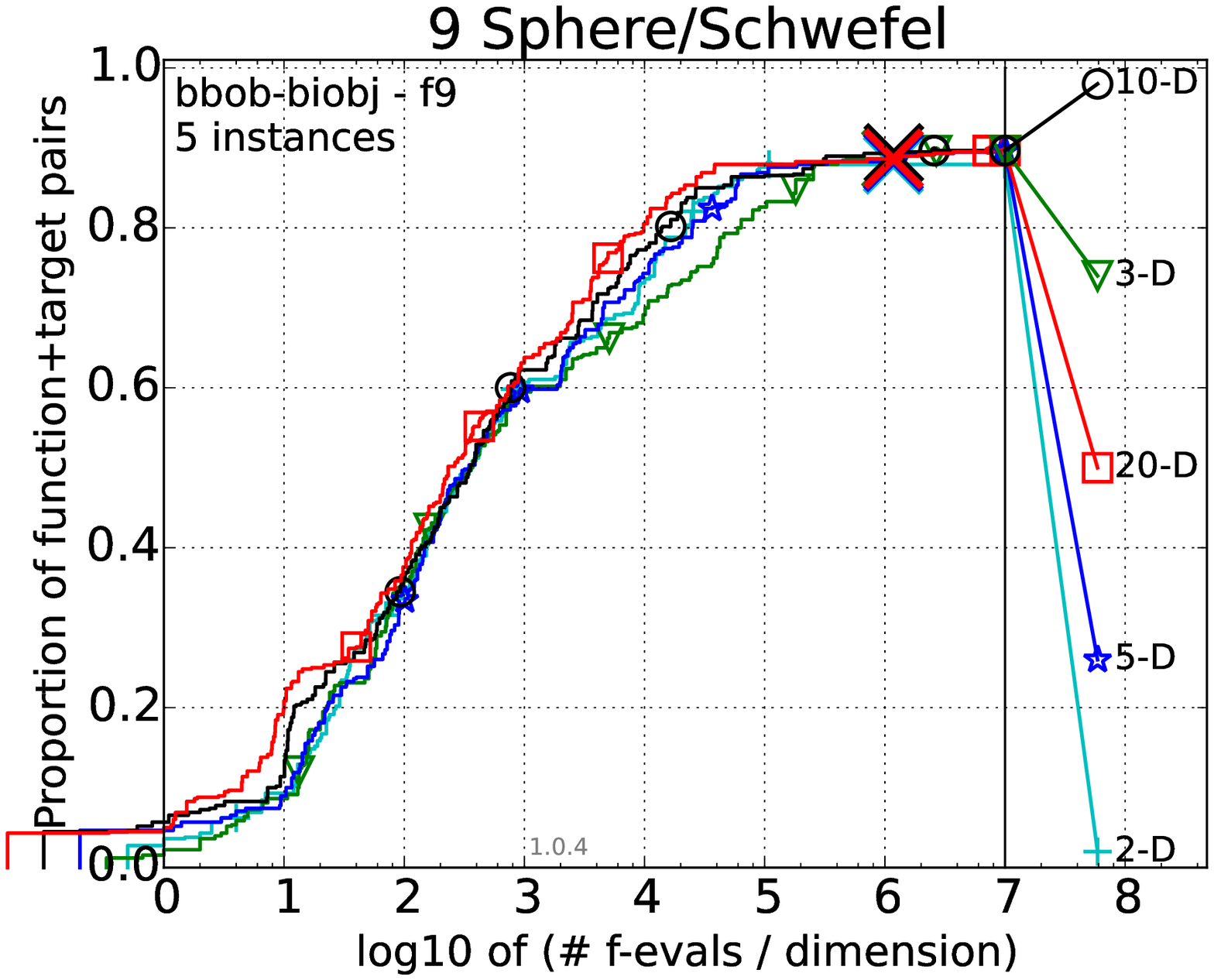}&
\includegraphics[width=0.25\textwidth]{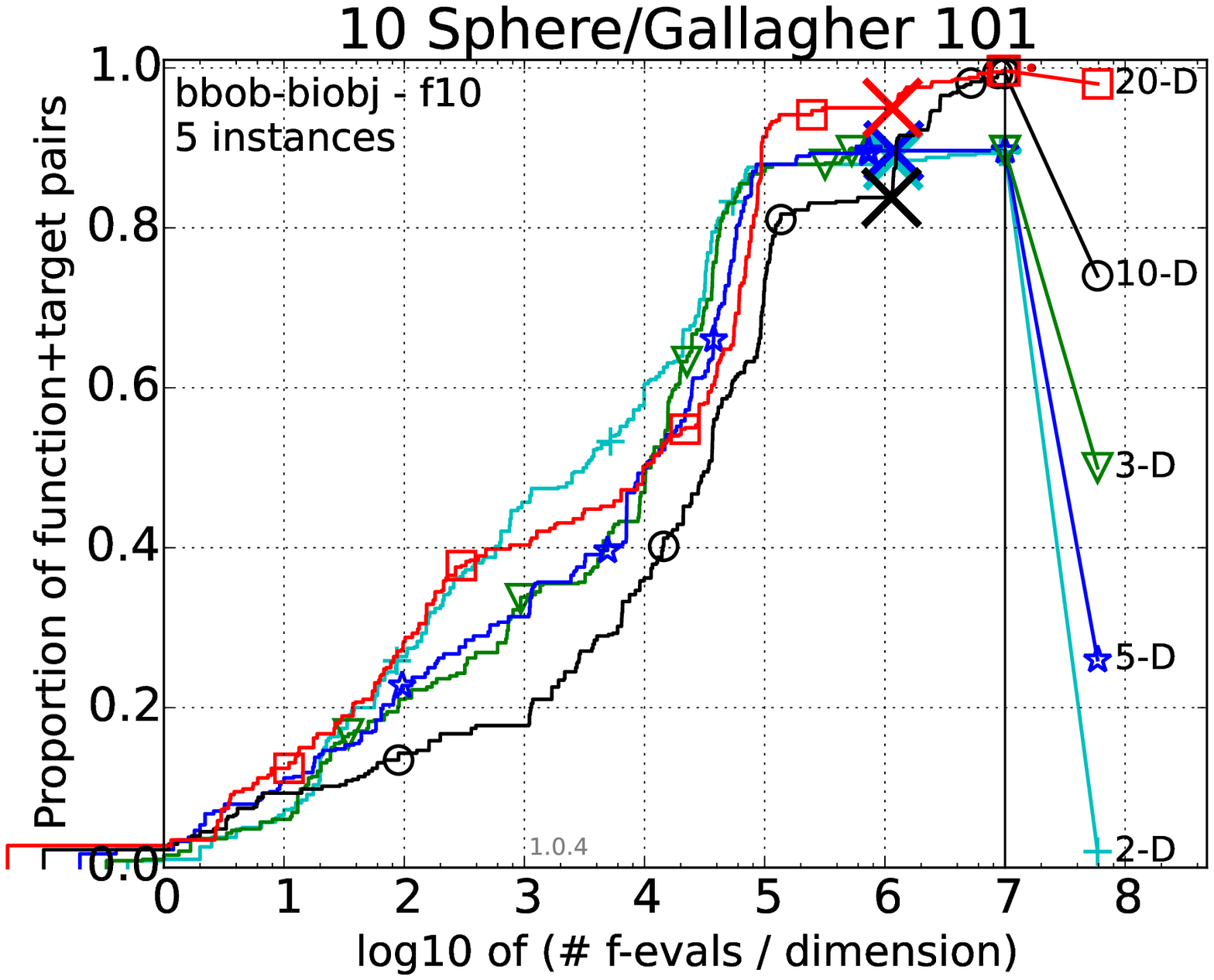}&
\includegraphics[width=0.25\textwidth]{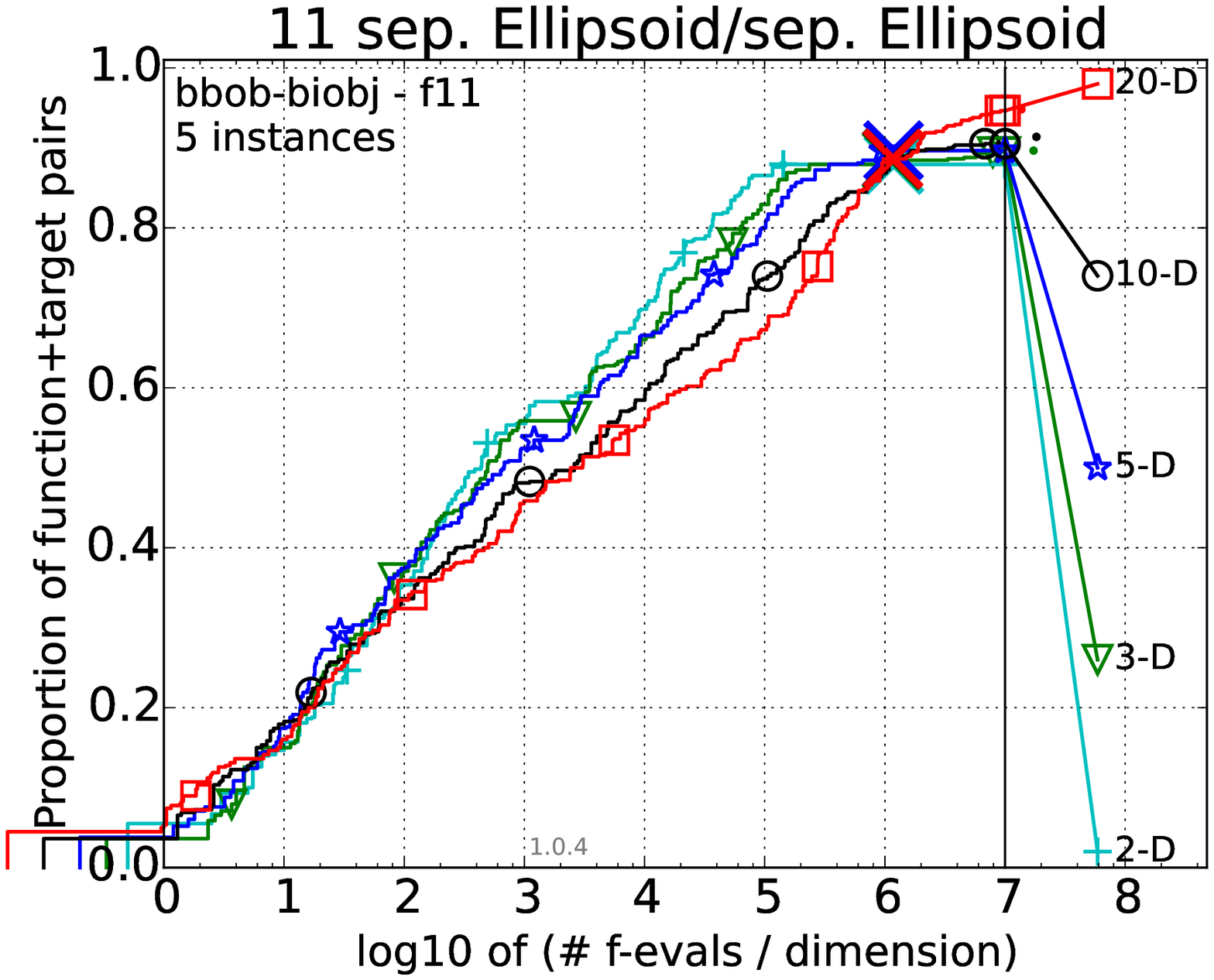}&
\includegraphics[width=0.25\textwidth]{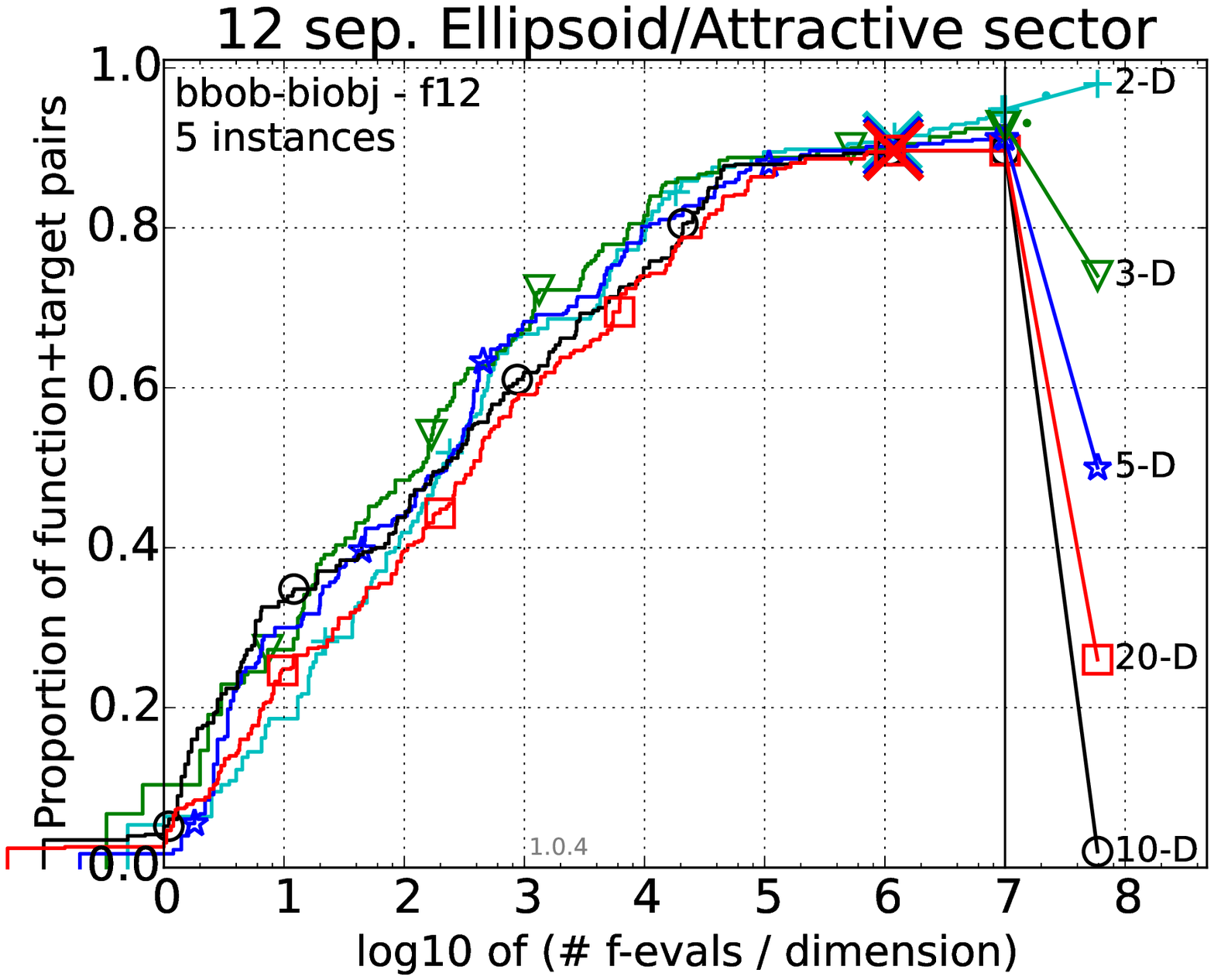}\\[-1.8ex]
\includegraphics[width=0.25\textwidth]{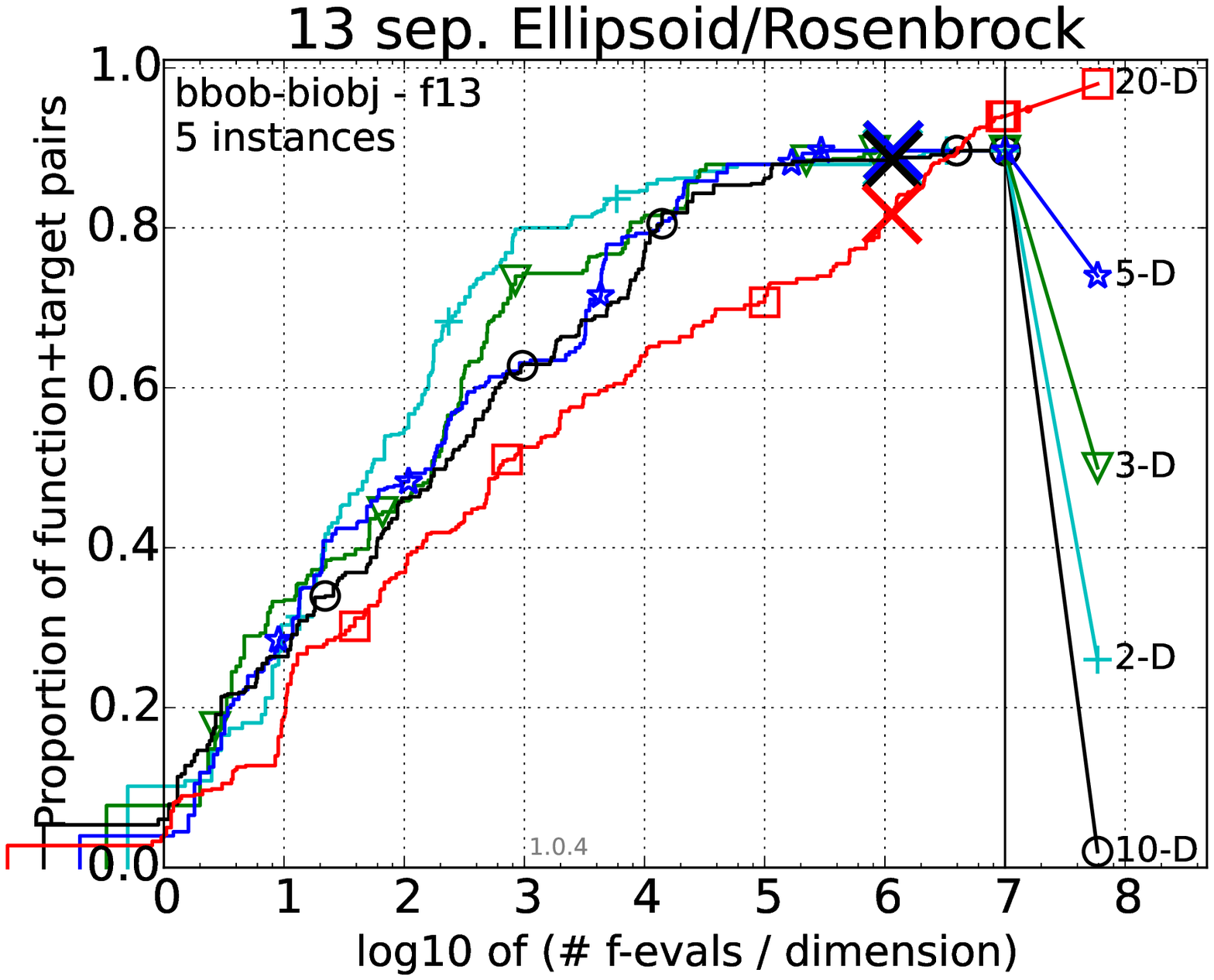}&
\includegraphics[width=0.25\textwidth]{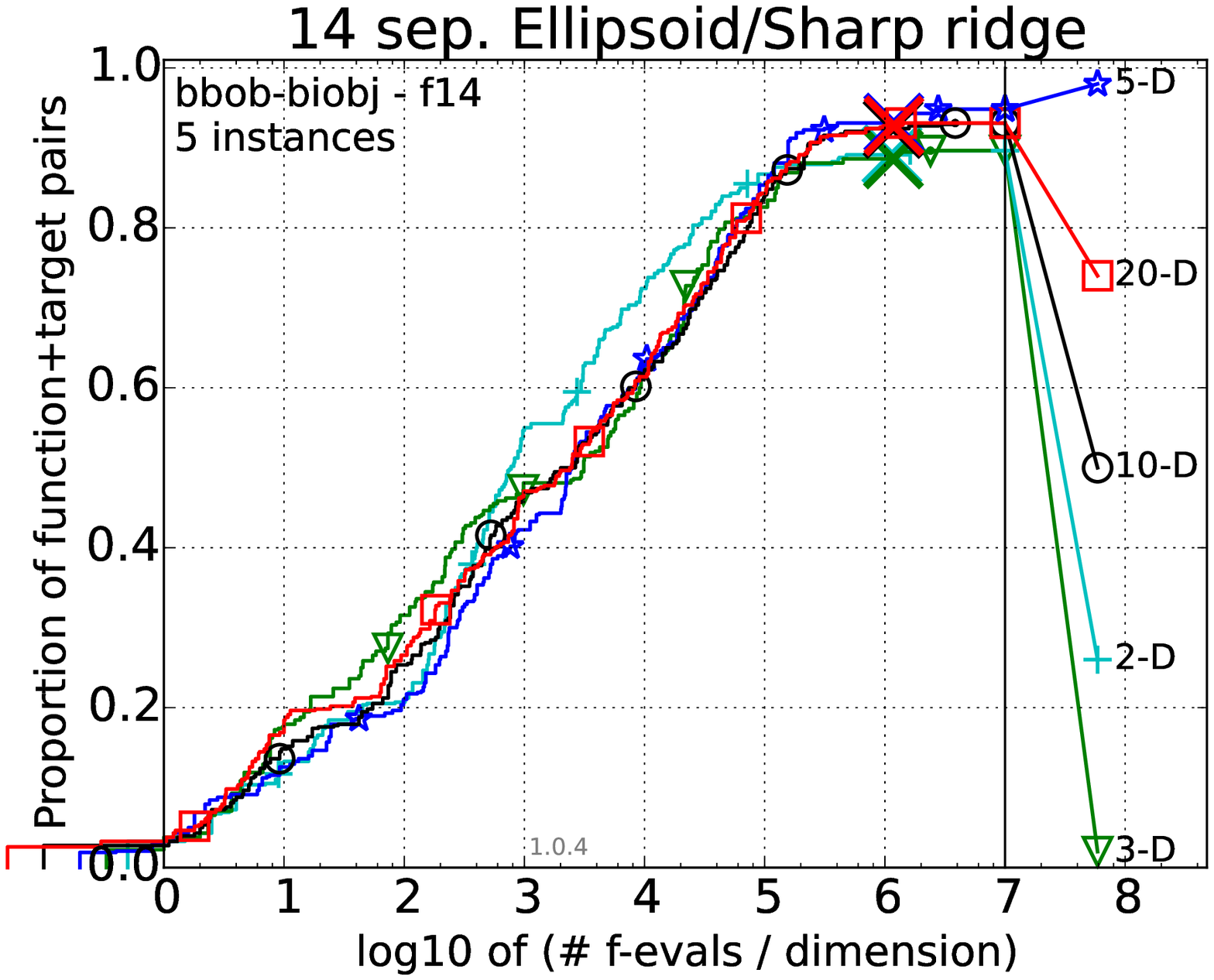}&
\includegraphics[width=0.25\textwidth]{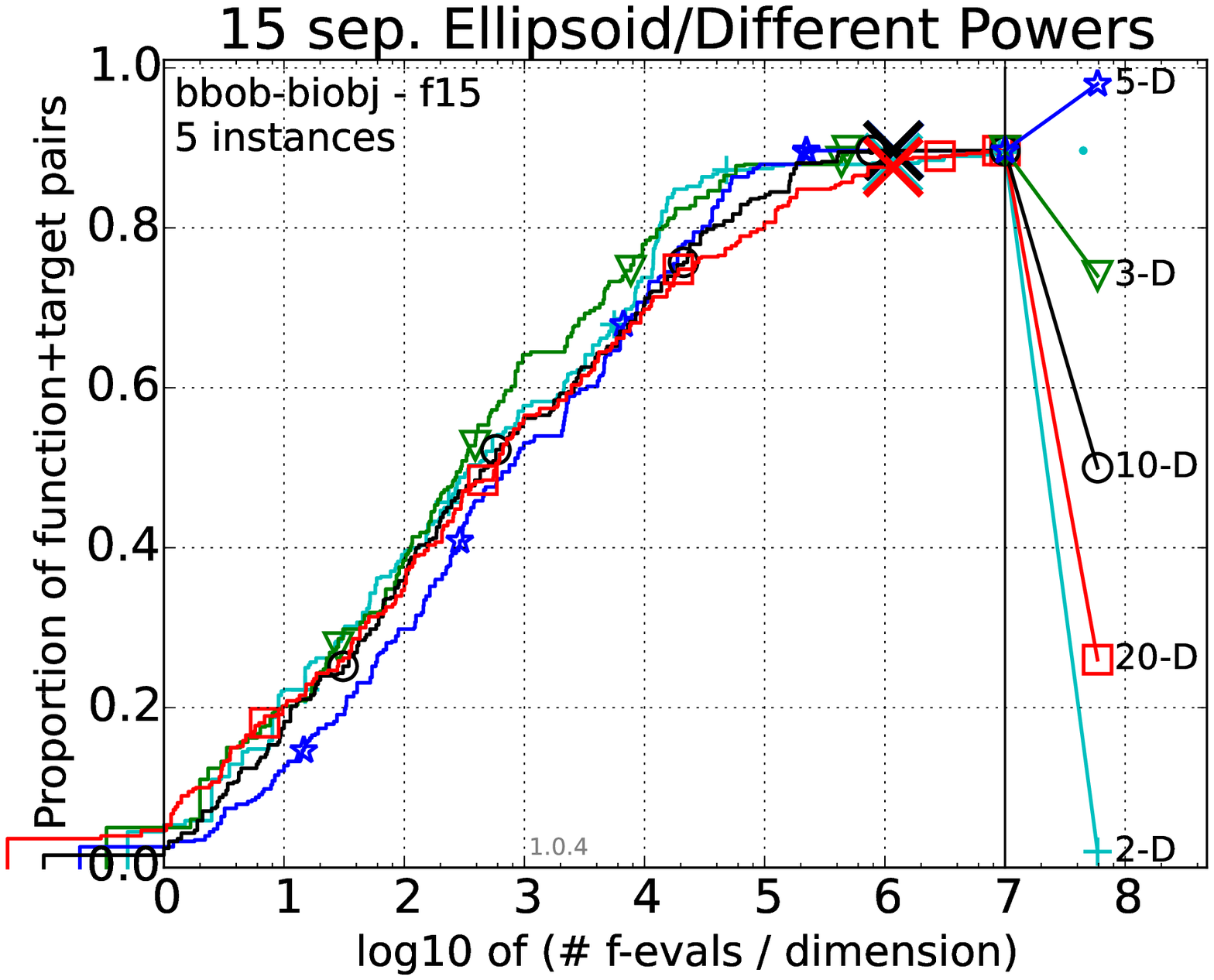}&
\includegraphics[width=0.25\textwidth]{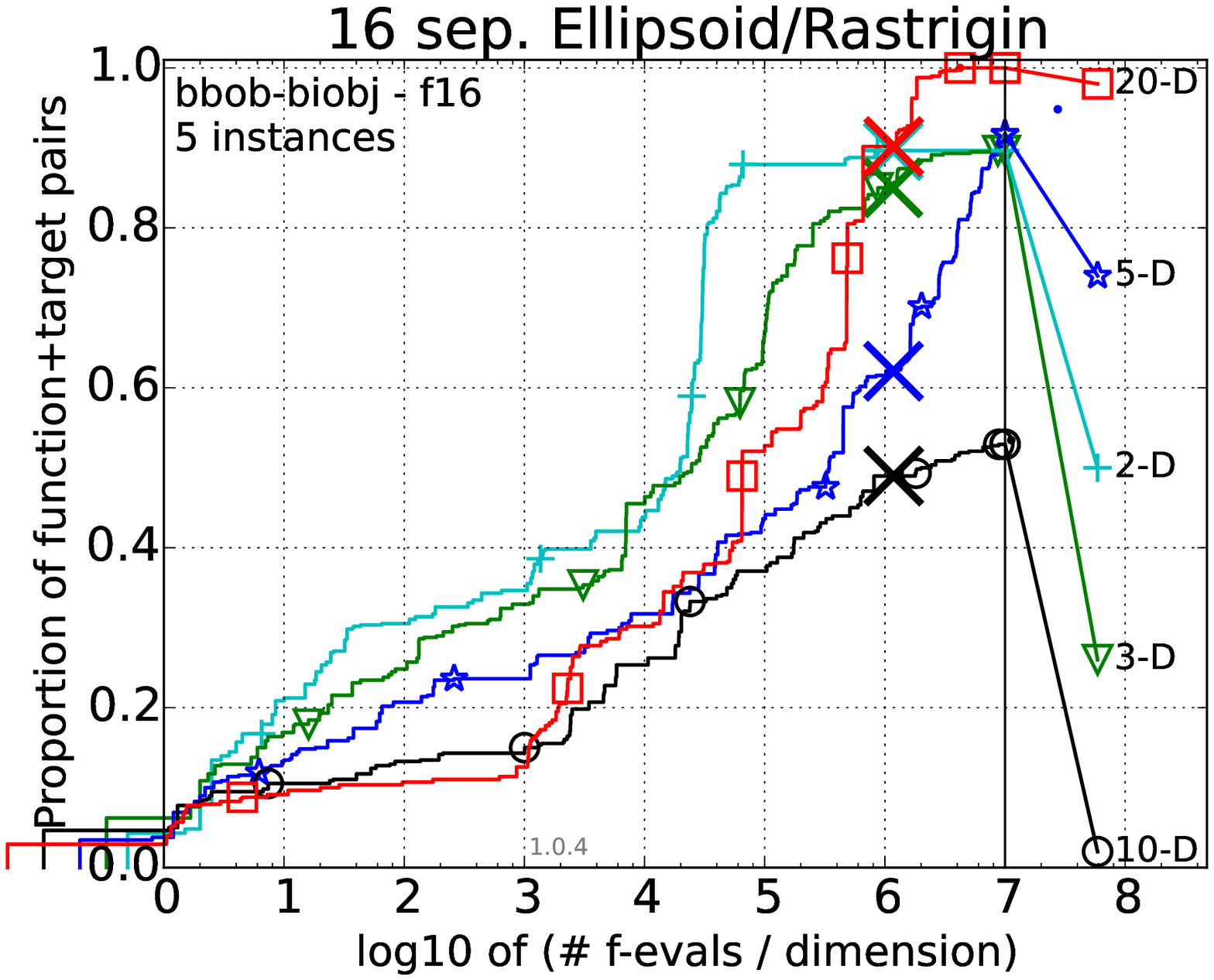}\\[-1.8ex]
\end{tabular}
 \caption{\label{fig:ECDFsingleOne}
 \bbobecdfcaptionsinglefcts{}
}

\end{figure*}

\begin{figure*}
\centering
\begin{tabular}{@{\hspace*{-0.018\textwidth}}l@{\hspace*{-0.02\textwidth}}l@{\hspace*{-0.02\textwidth}}l@{\hspace*{-0.02\textwidth}}l@{\hspace*{-0.02\textwidth}}l@{\hspace*{-0.02\textwidth}}}
\includegraphics[width=0.25\textwidth]{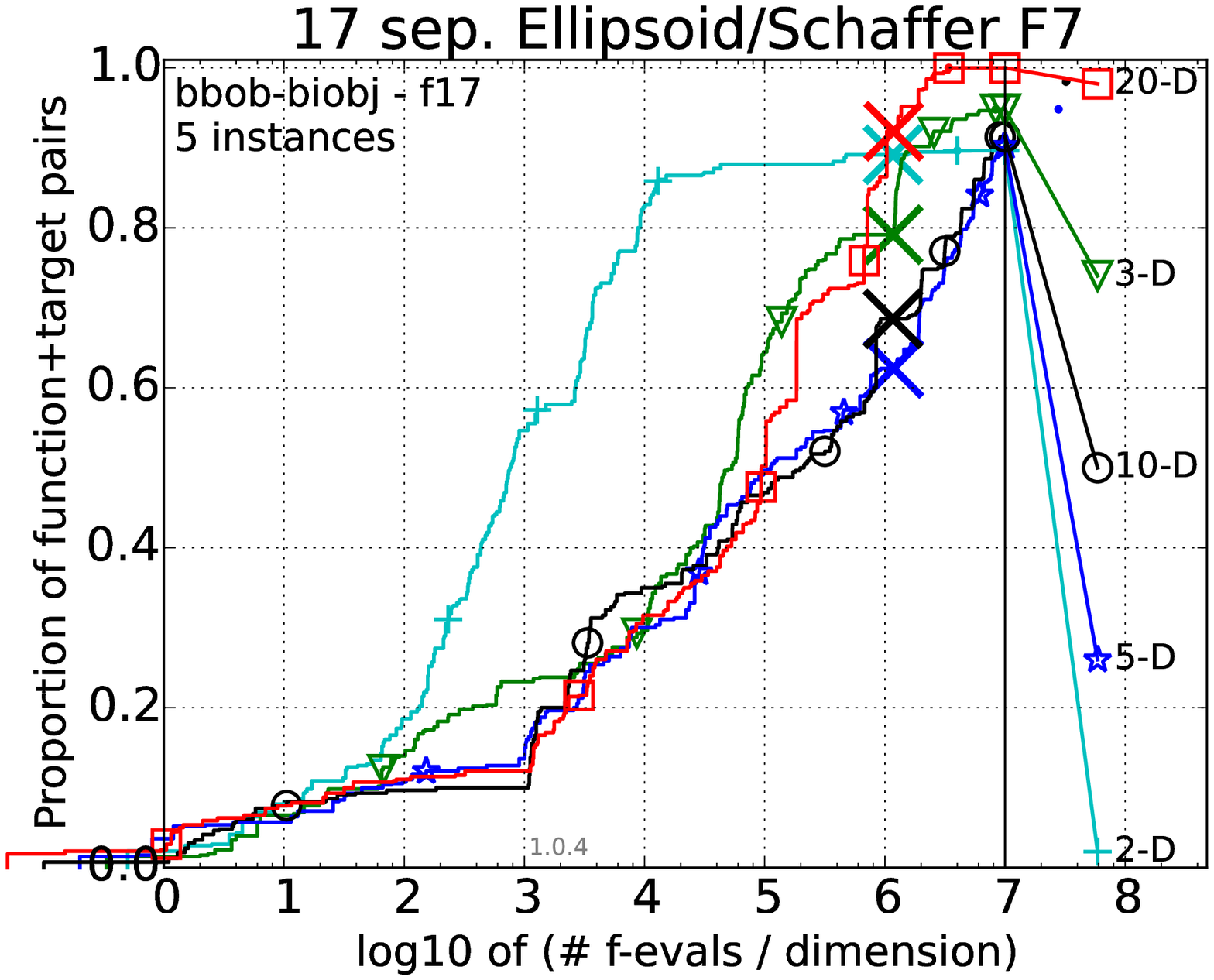}&
\includegraphics[width=0.25\textwidth]{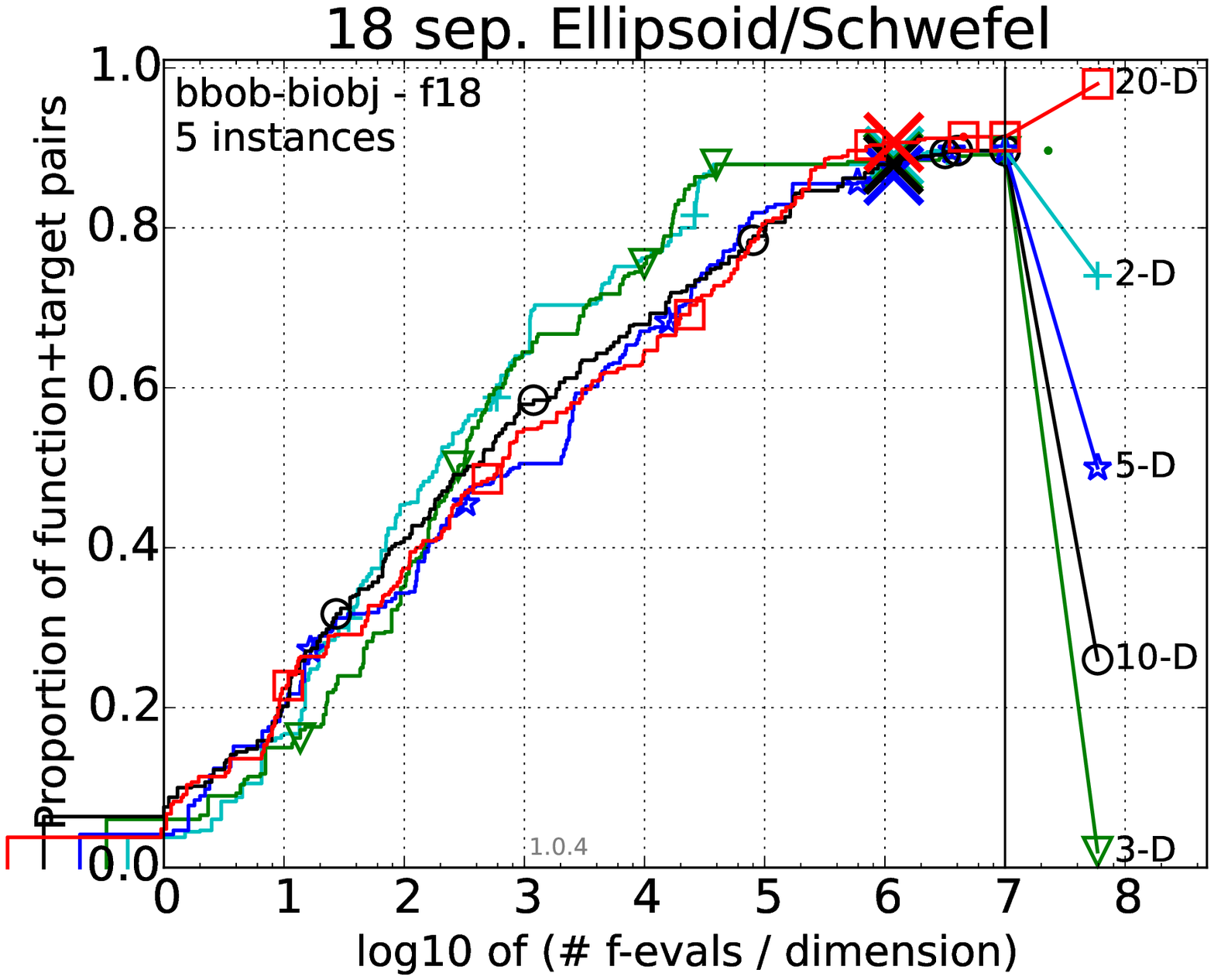}&
\includegraphics[width=0.25\textwidth]{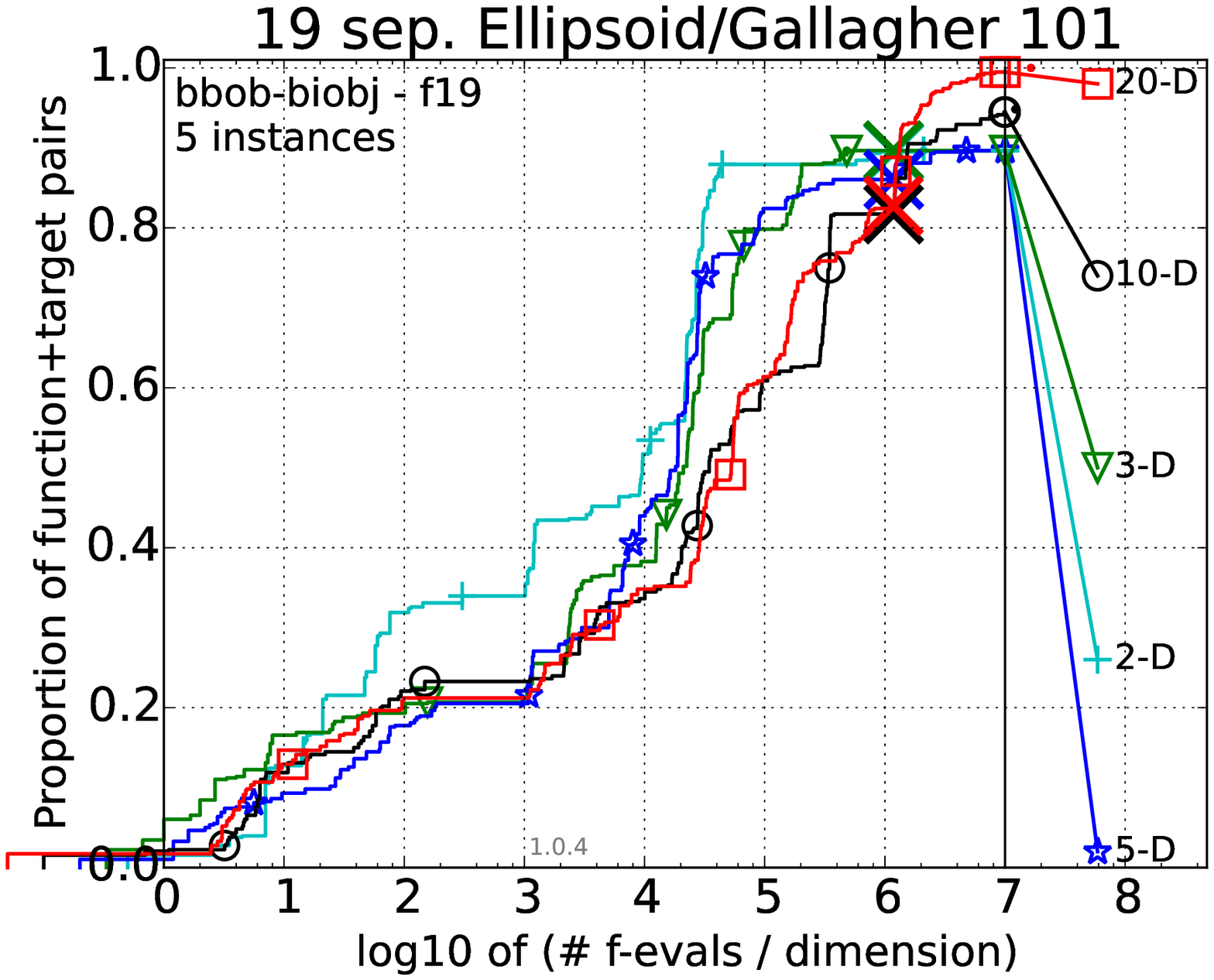}&
\includegraphics[width=0.25\textwidth]{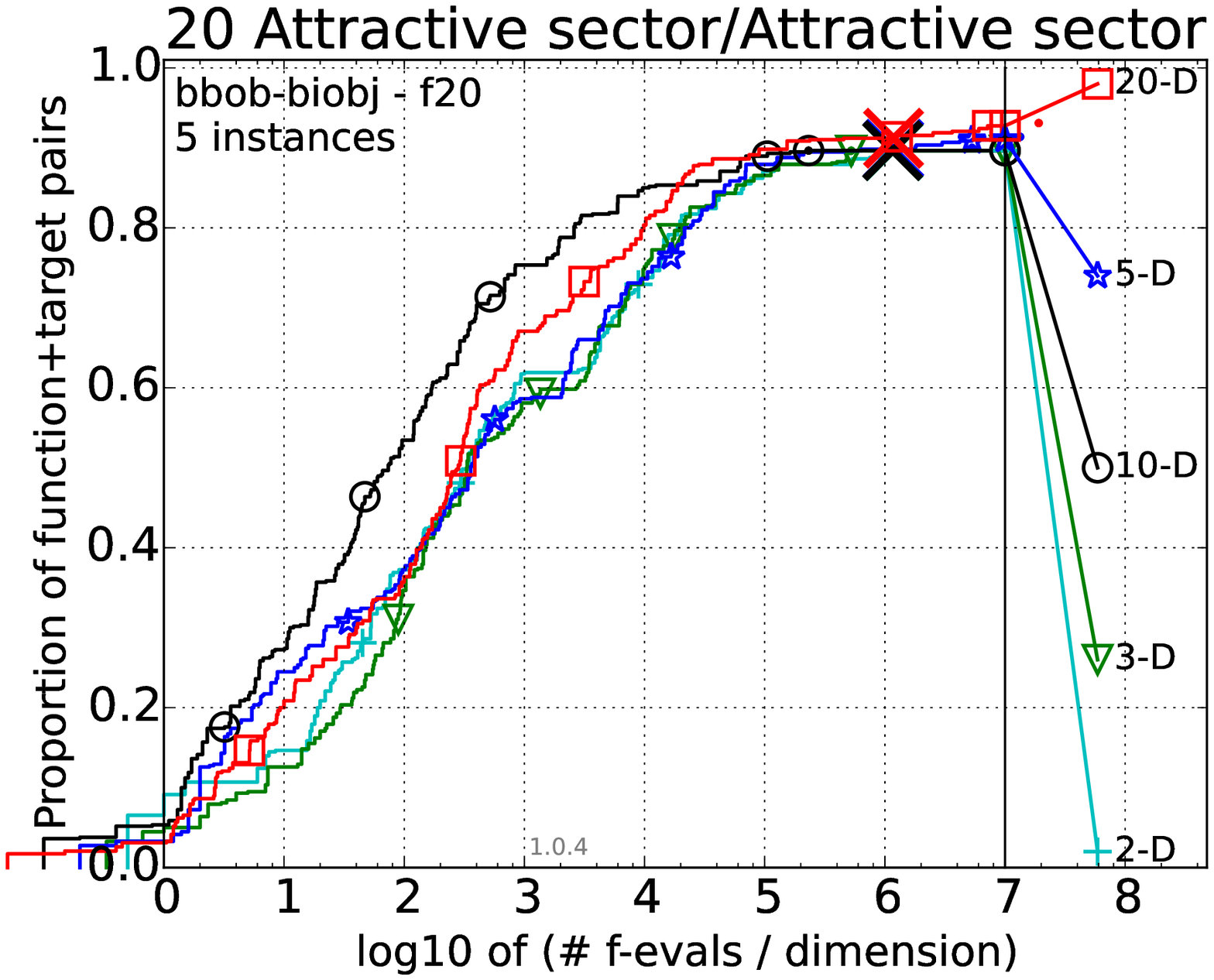}\\[-1.8ex]
\includegraphics[width=0.25\textwidth]{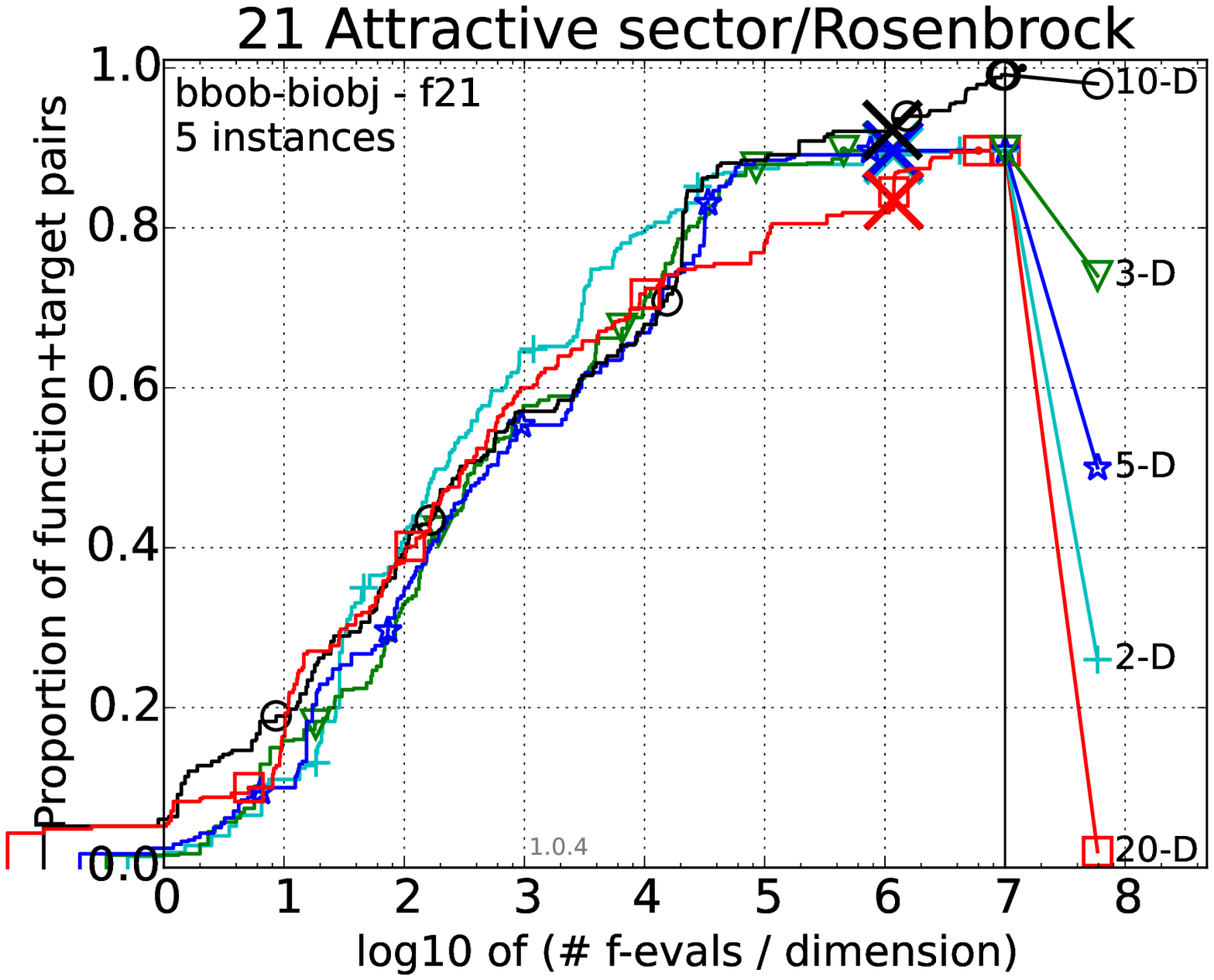}&
\includegraphics[width=0.25\textwidth]{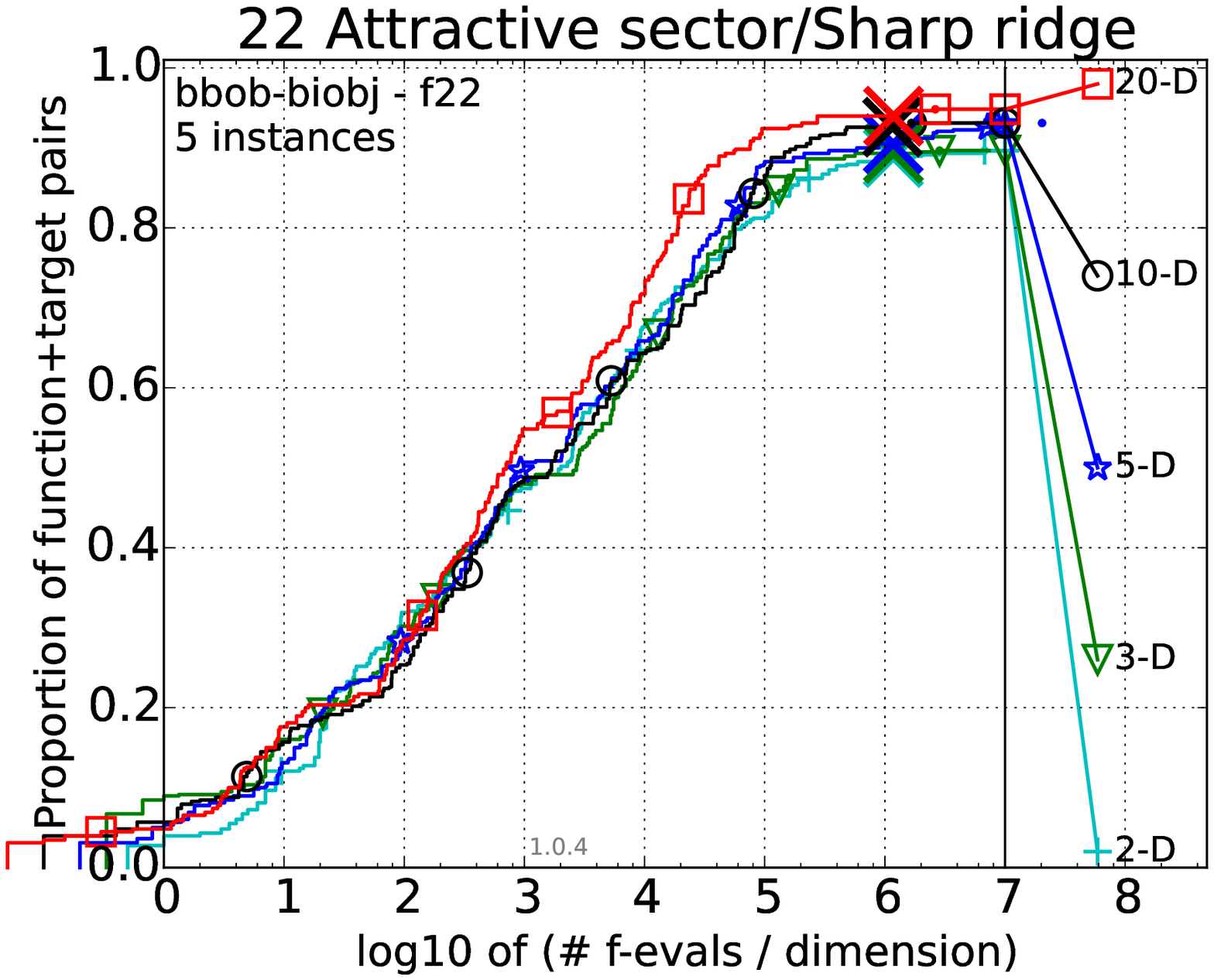}&
\includegraphics[width=0.25\textwidth]{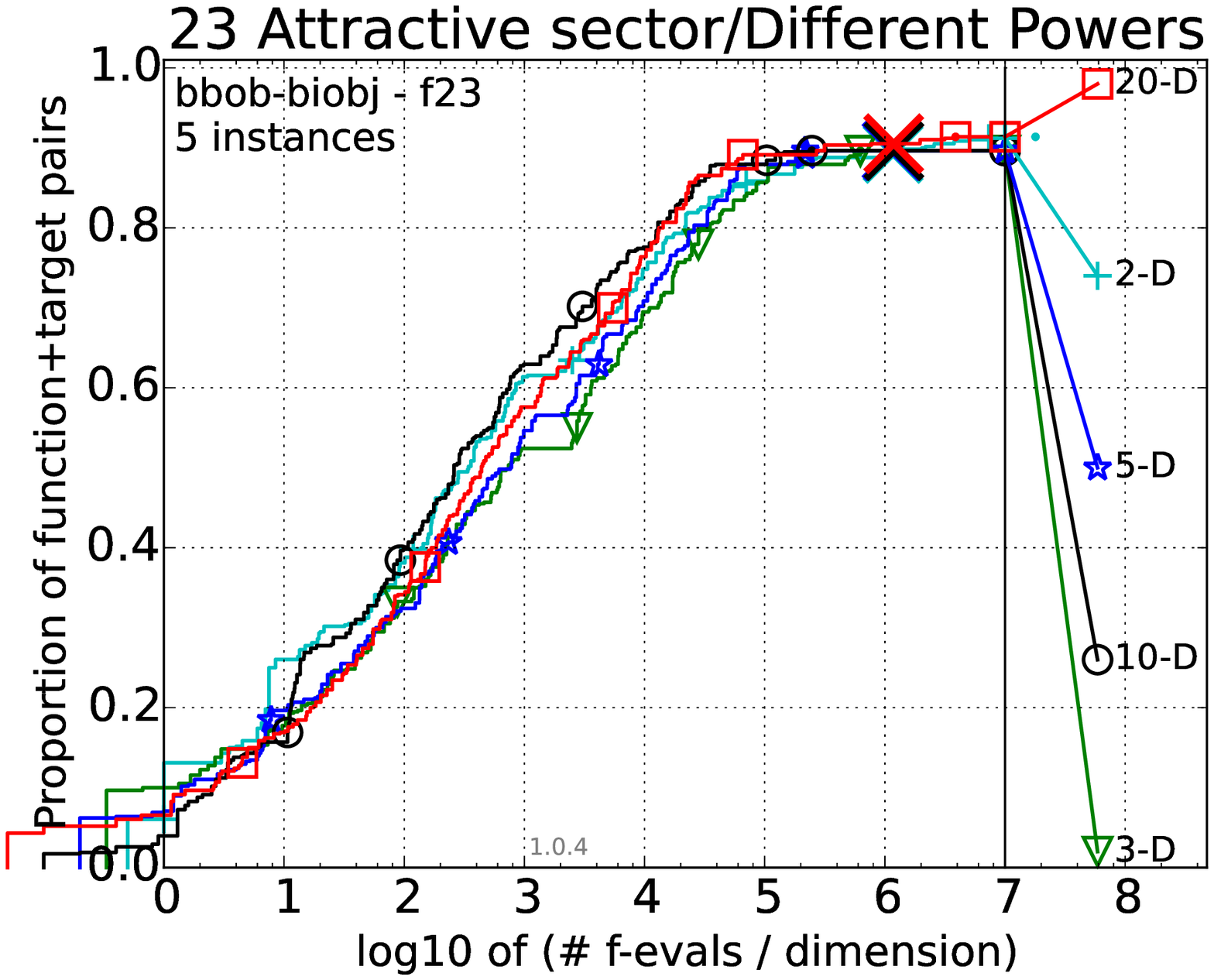}&
\includegraphics[width=0.25\textwidth]{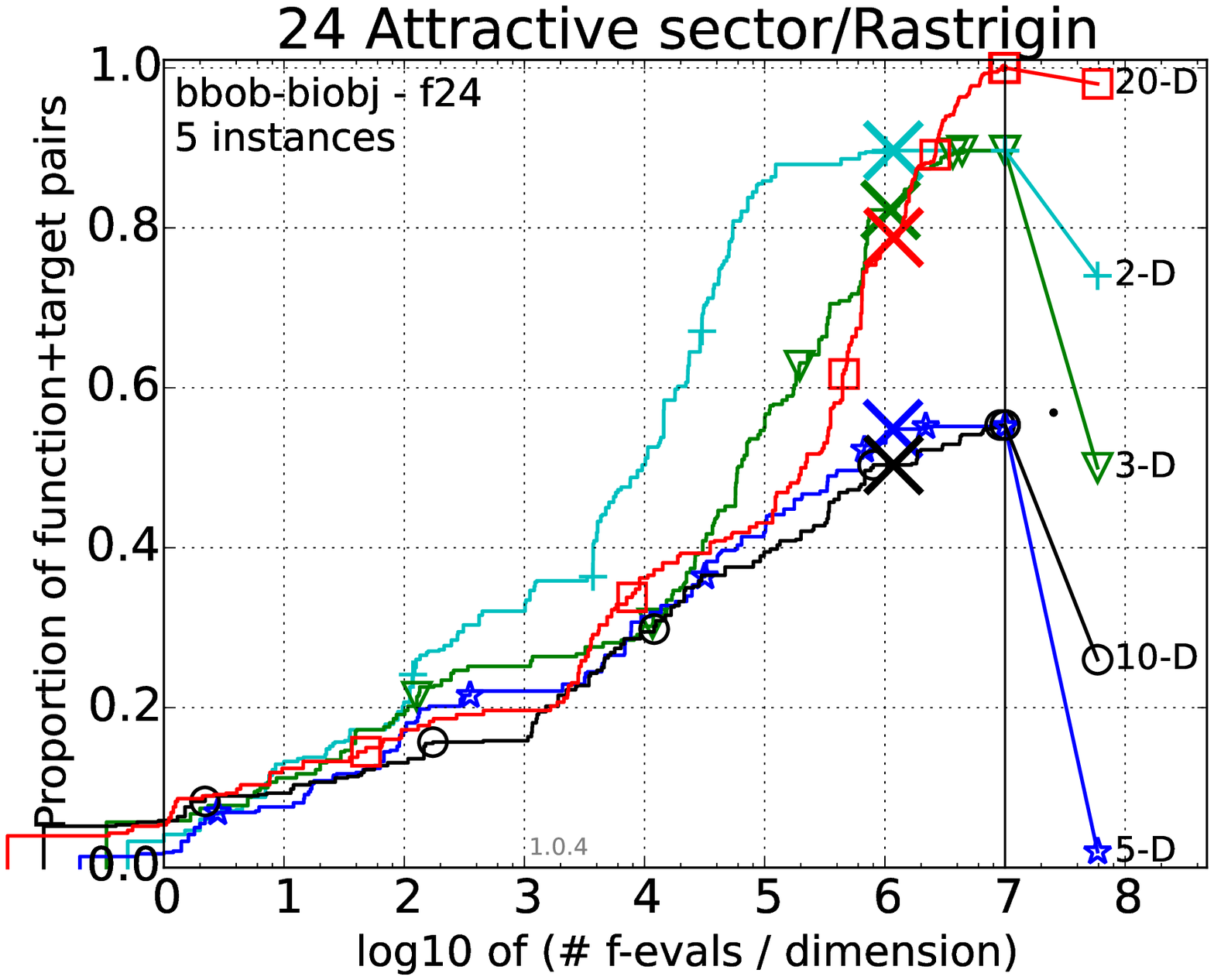}\\[-1.8ex]
\includegraphics[width=0.25\textwidth]{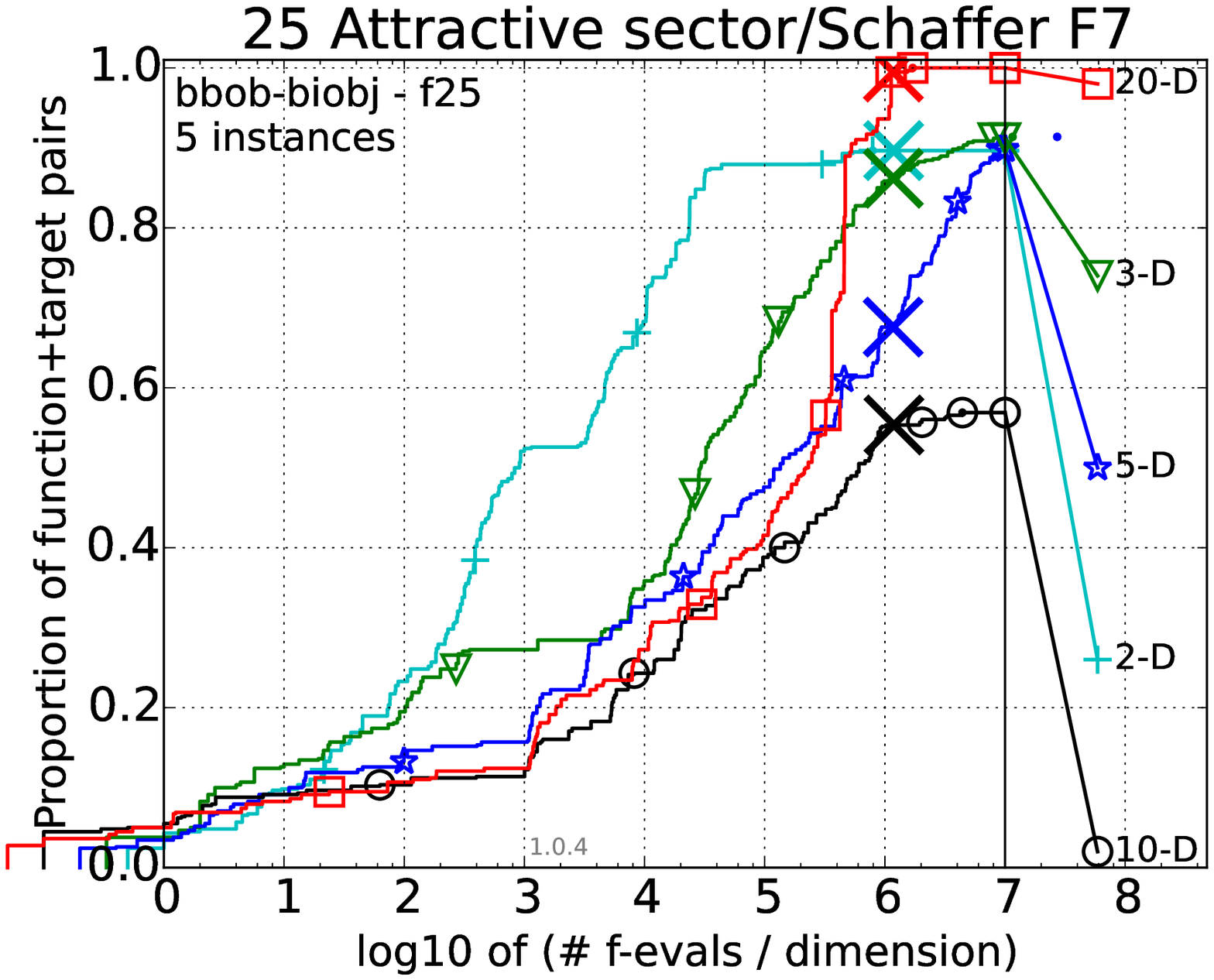}&
\includegraphics[width=0.25\textwidth]{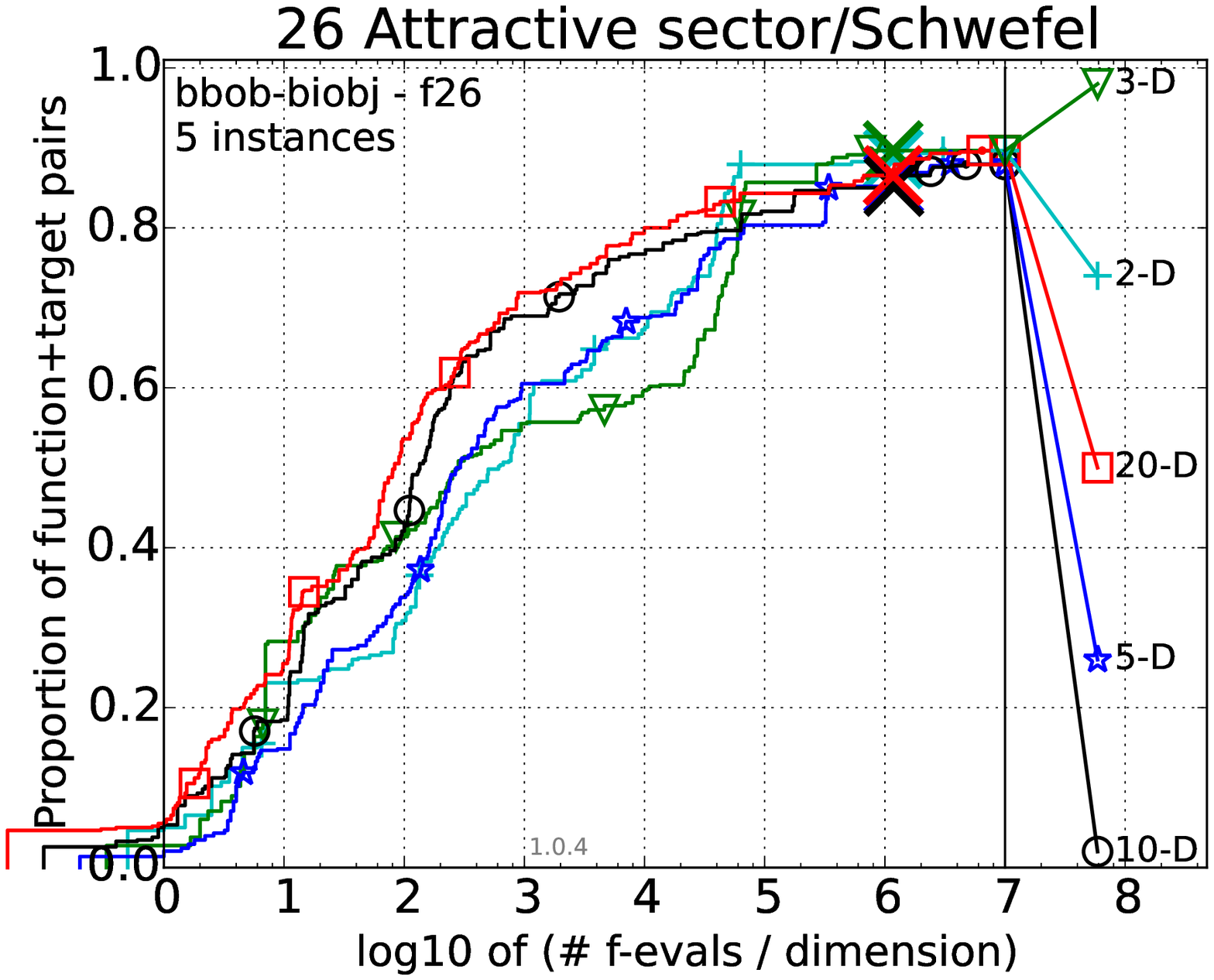}&
\includegraphics[width=0.25\textwidth]{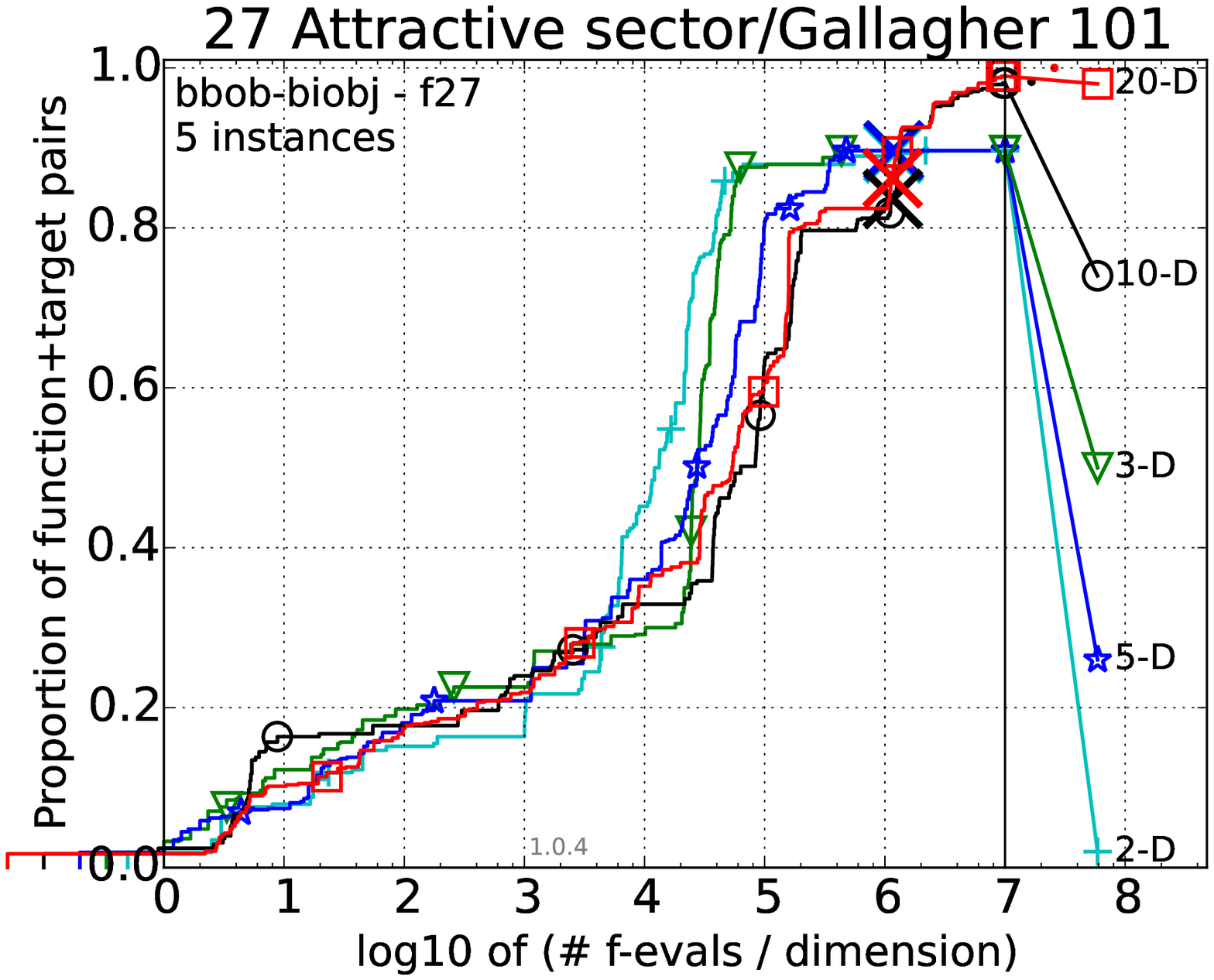}&
\includegraphics[width=0.25\textwidth]{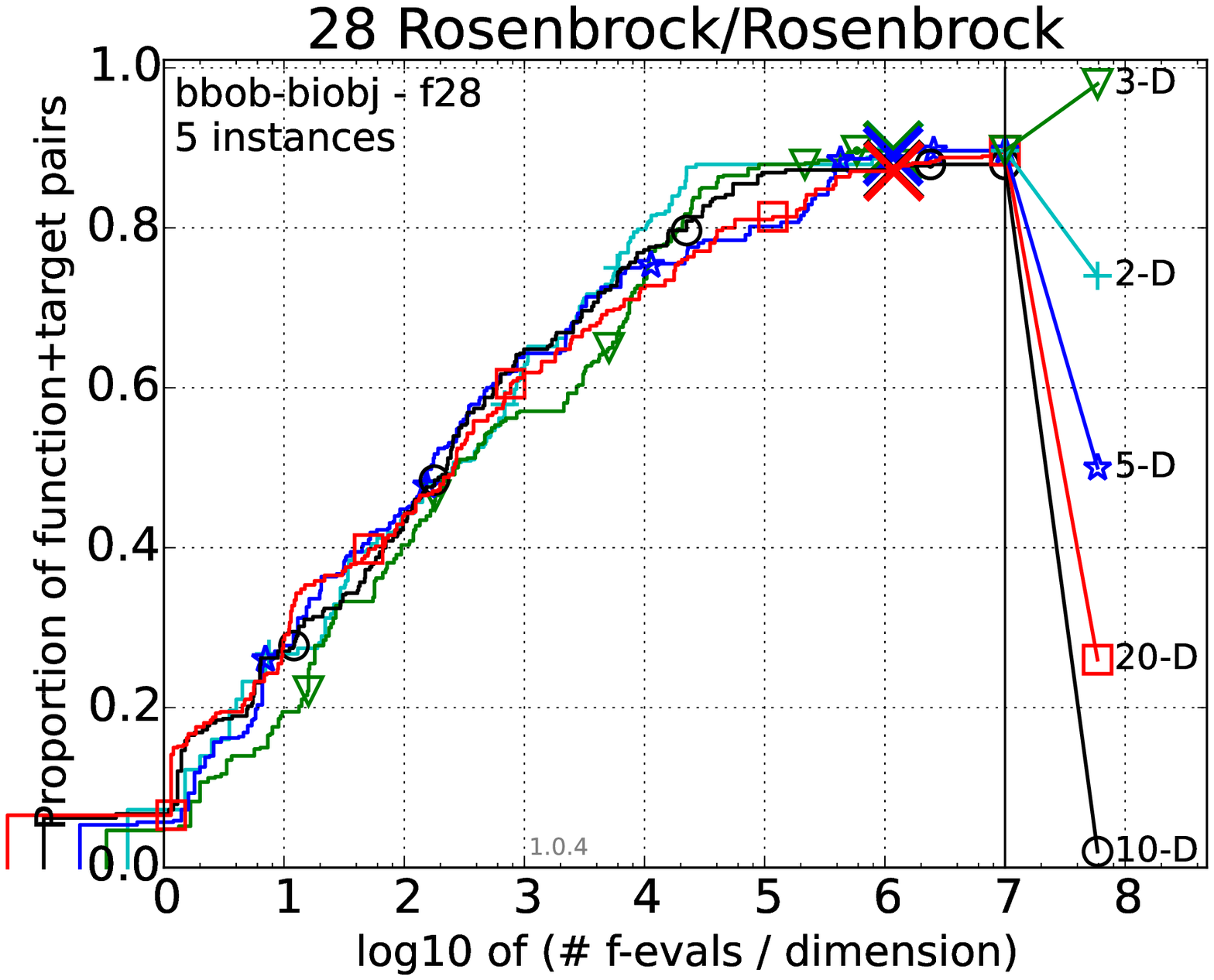}\\[-1.8ex]
\includegraphics[width=0.25\textwidth]{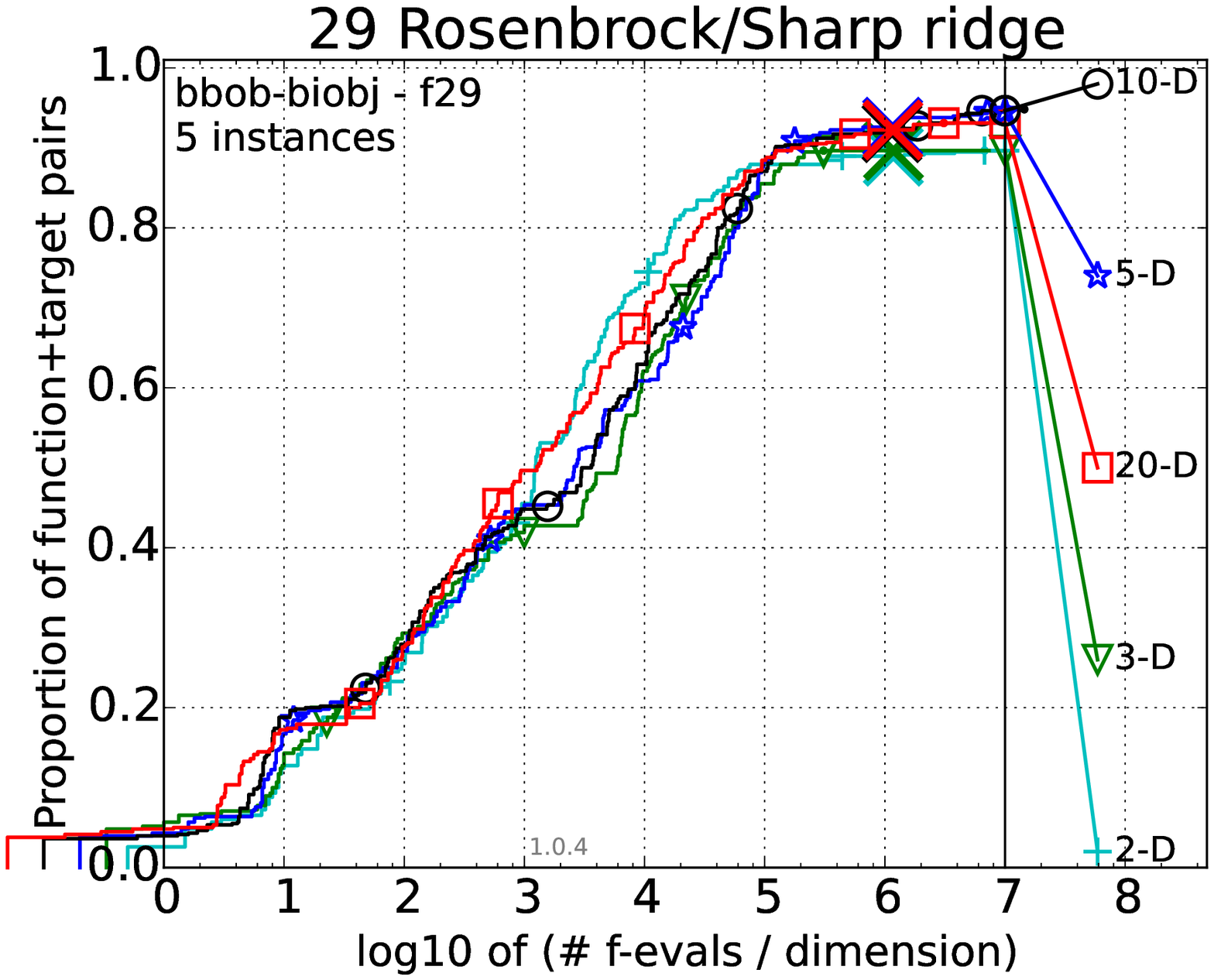}&
\includegraphics[width=0.25\textwidth]{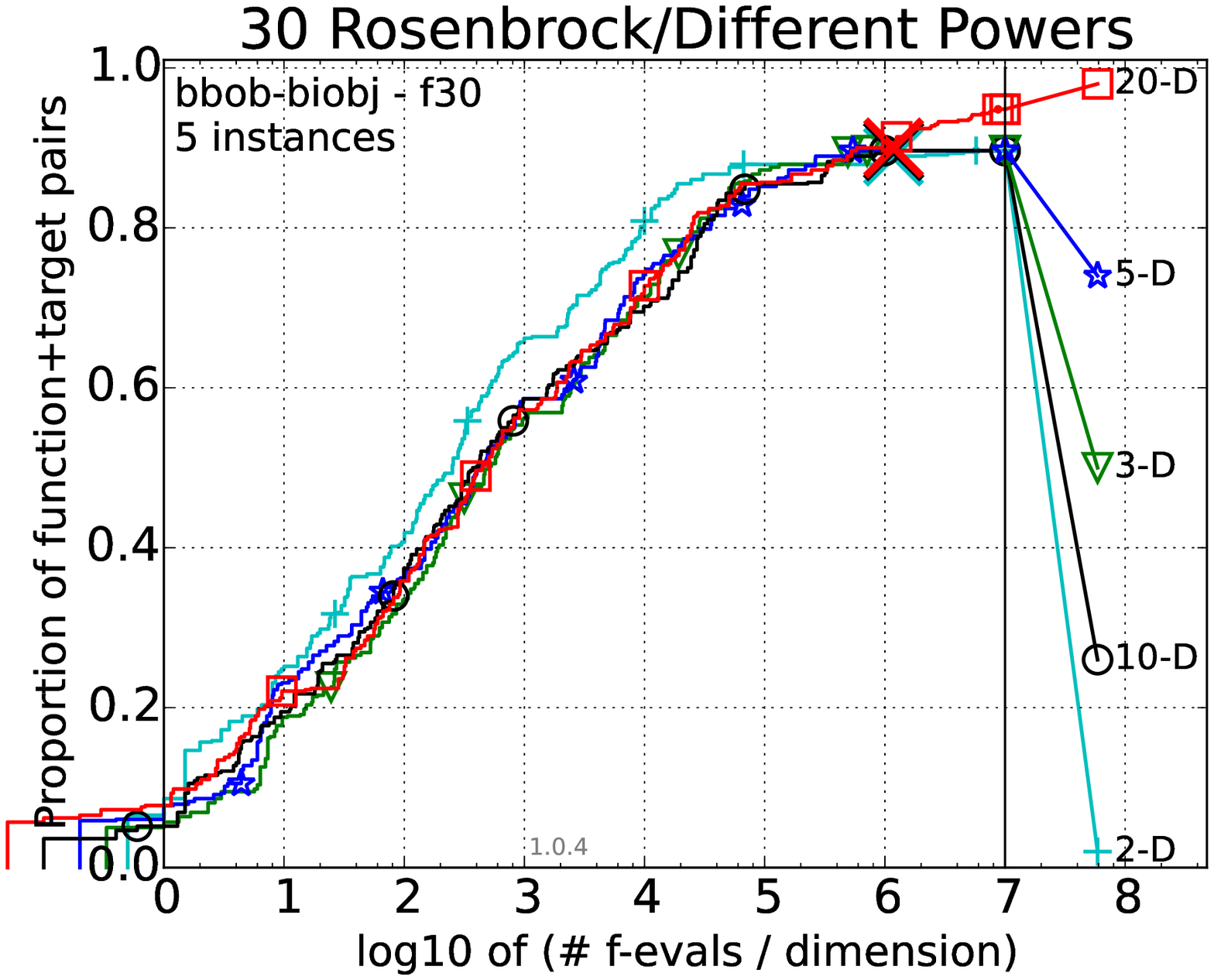}&
\includegraphics[width=0.25\textwidth]{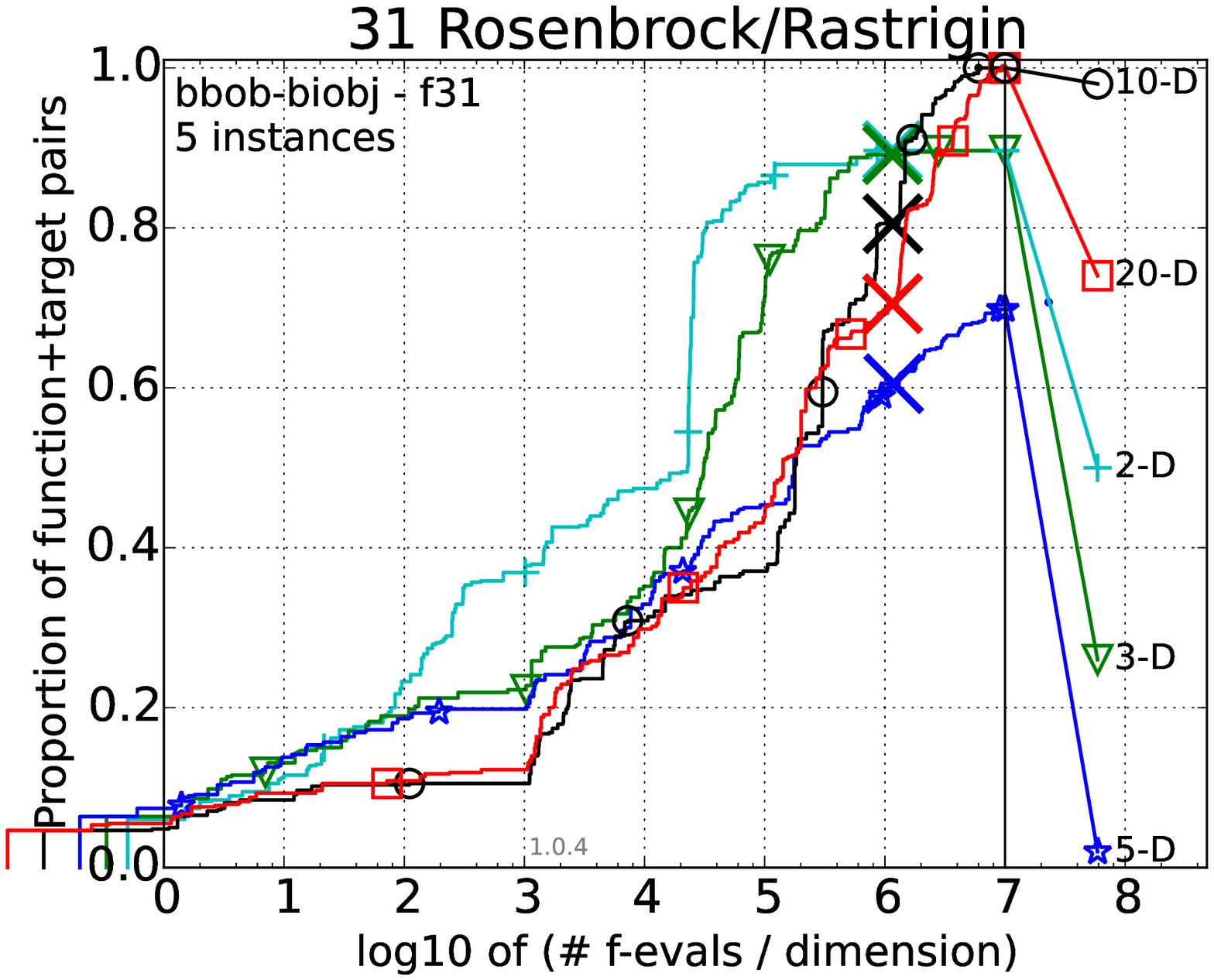}&
\includegraphics[width=0.25\textwidth]{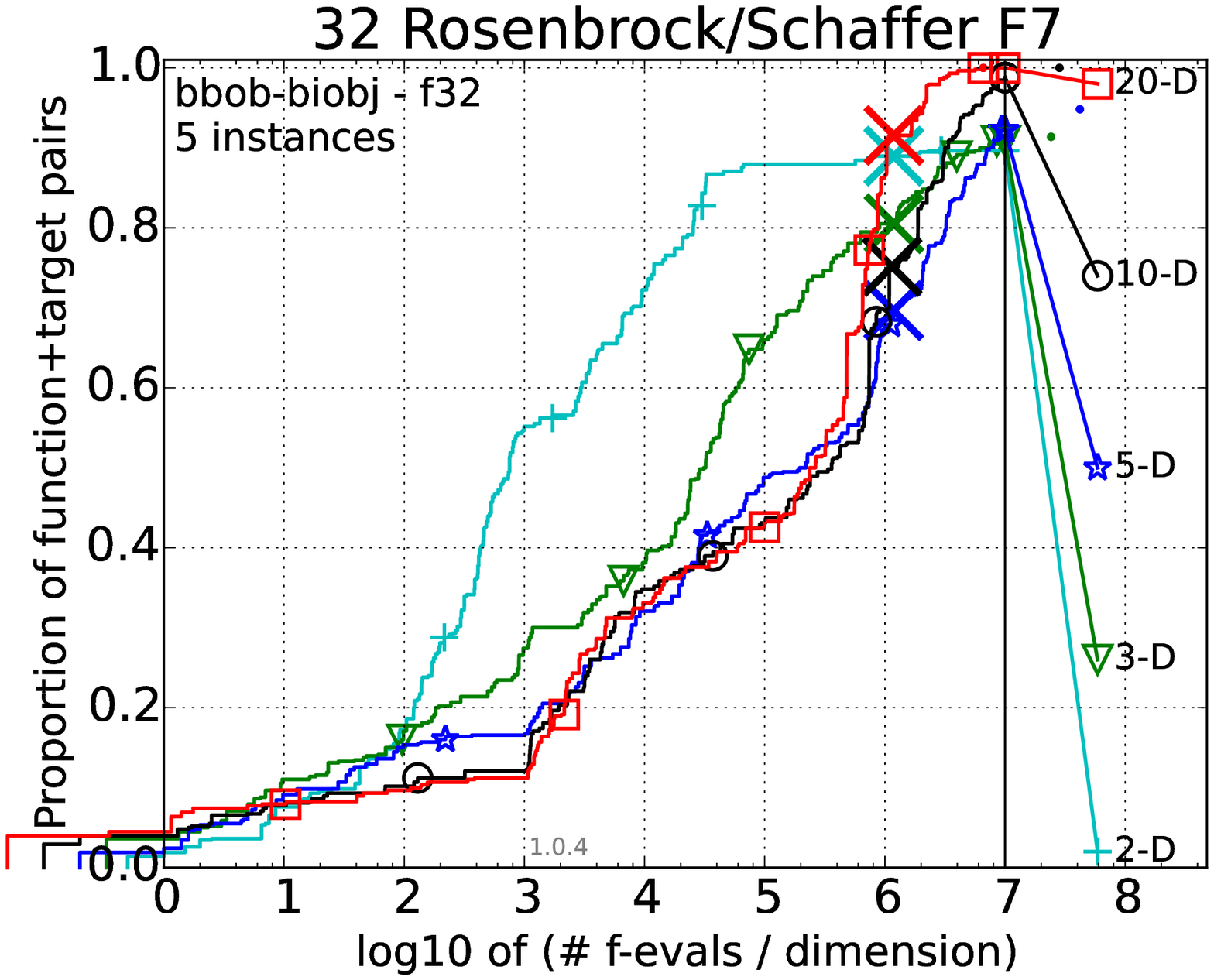}\\[-1.8ex]
\includegraphics[width=0.25\textwidth]{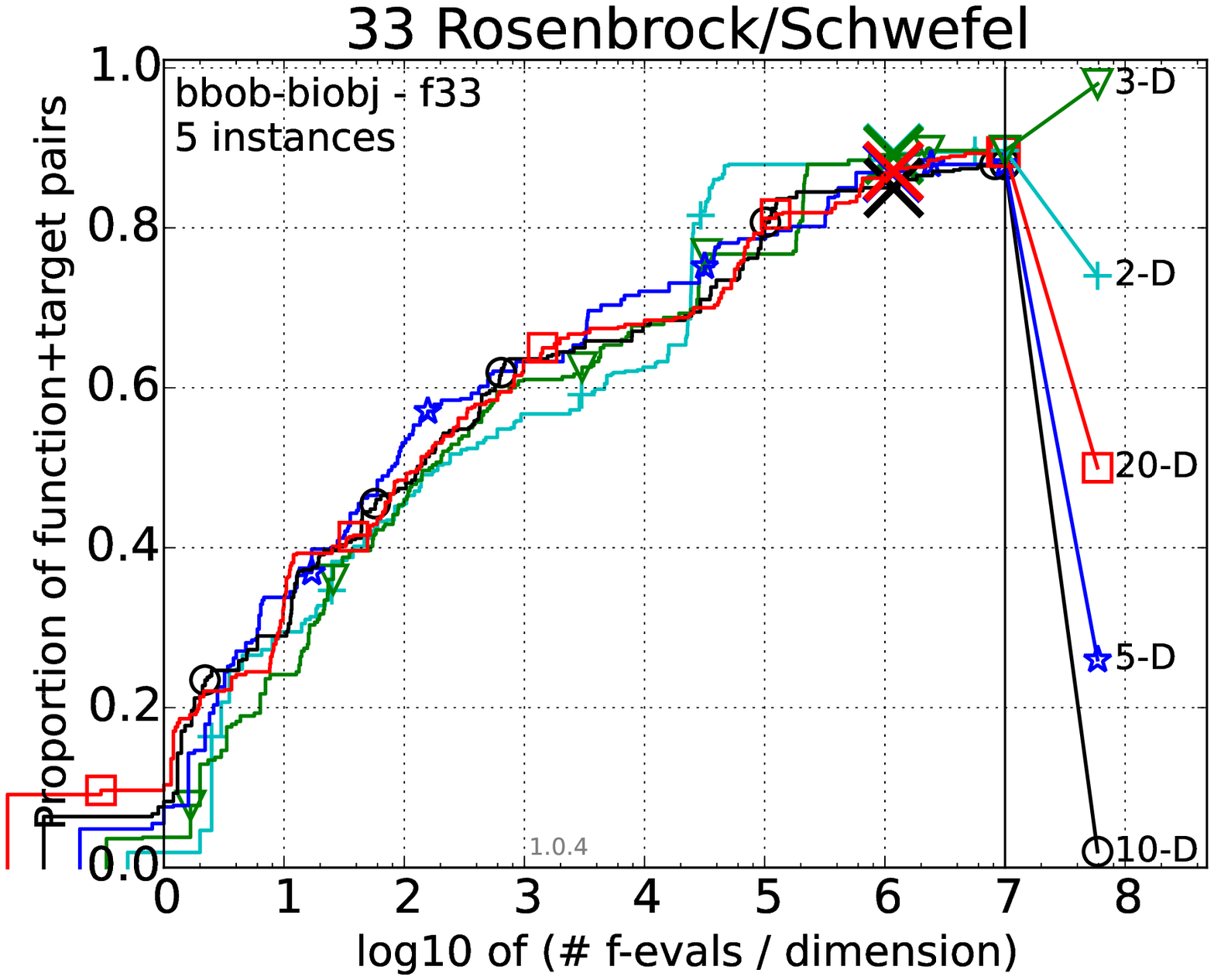}&
\includegraphics[width=0.25\textwidth]{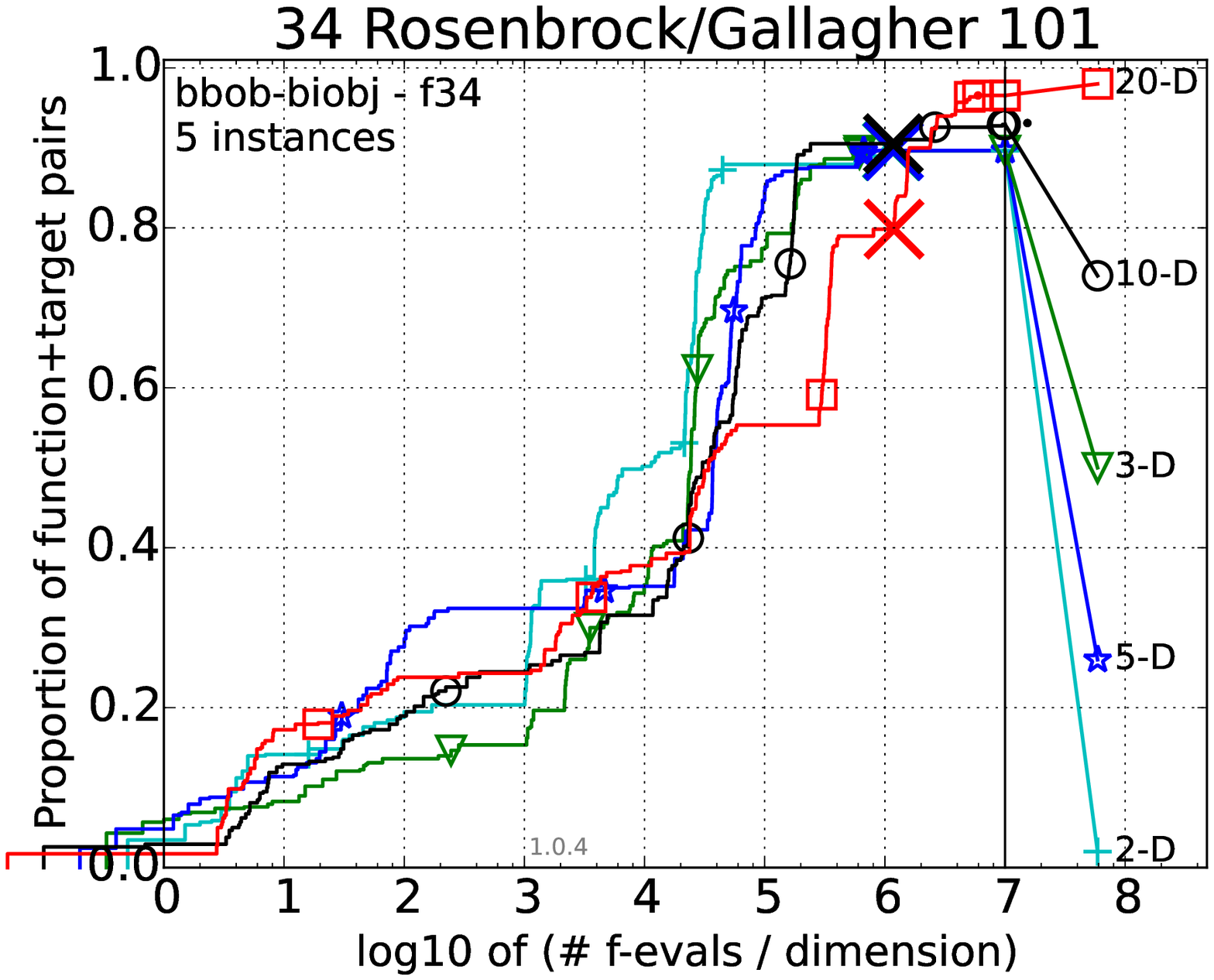}&
\includegraphics[width=0.25\textwidth]{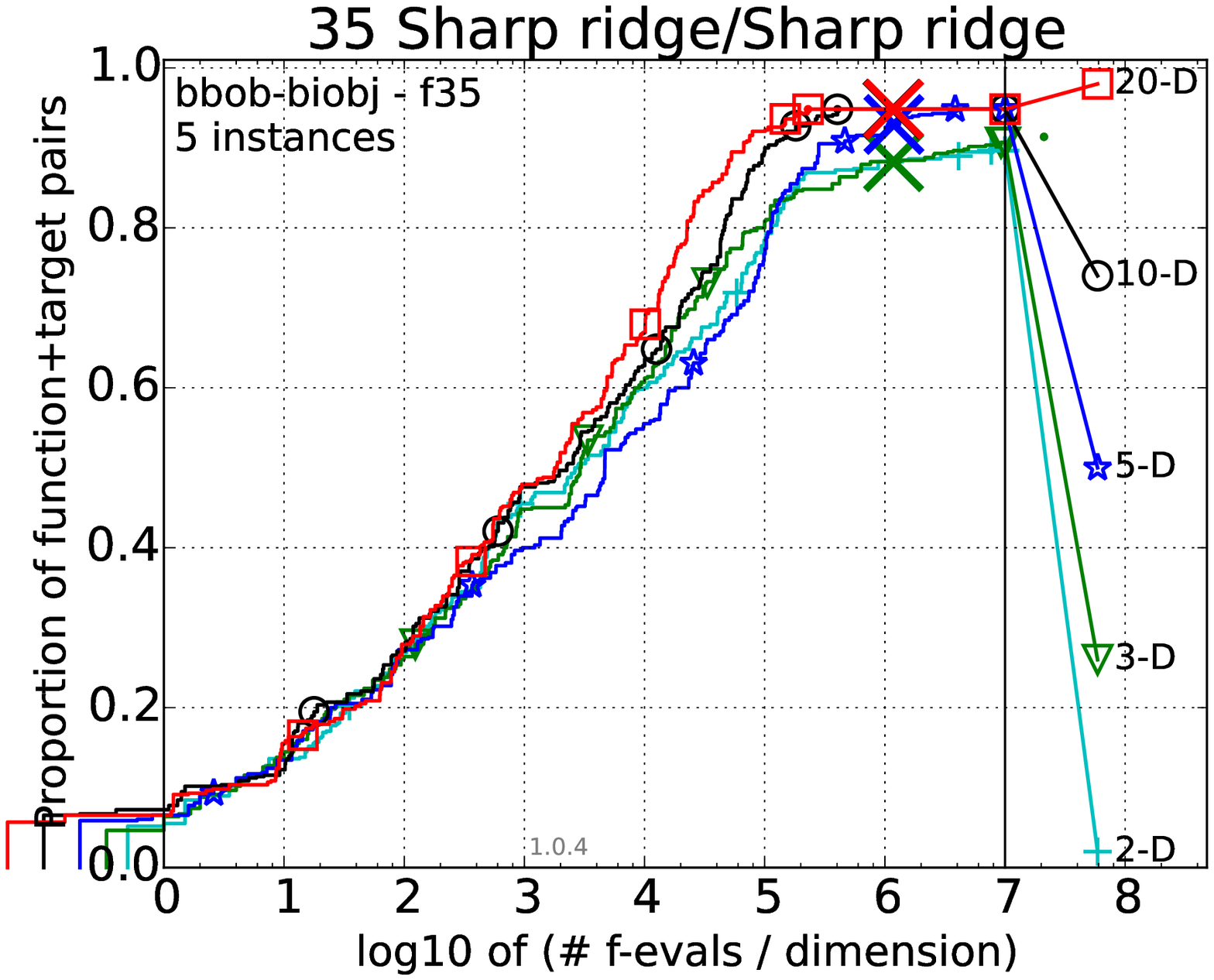}&
\includegraphics[width=0.25\textwidth]{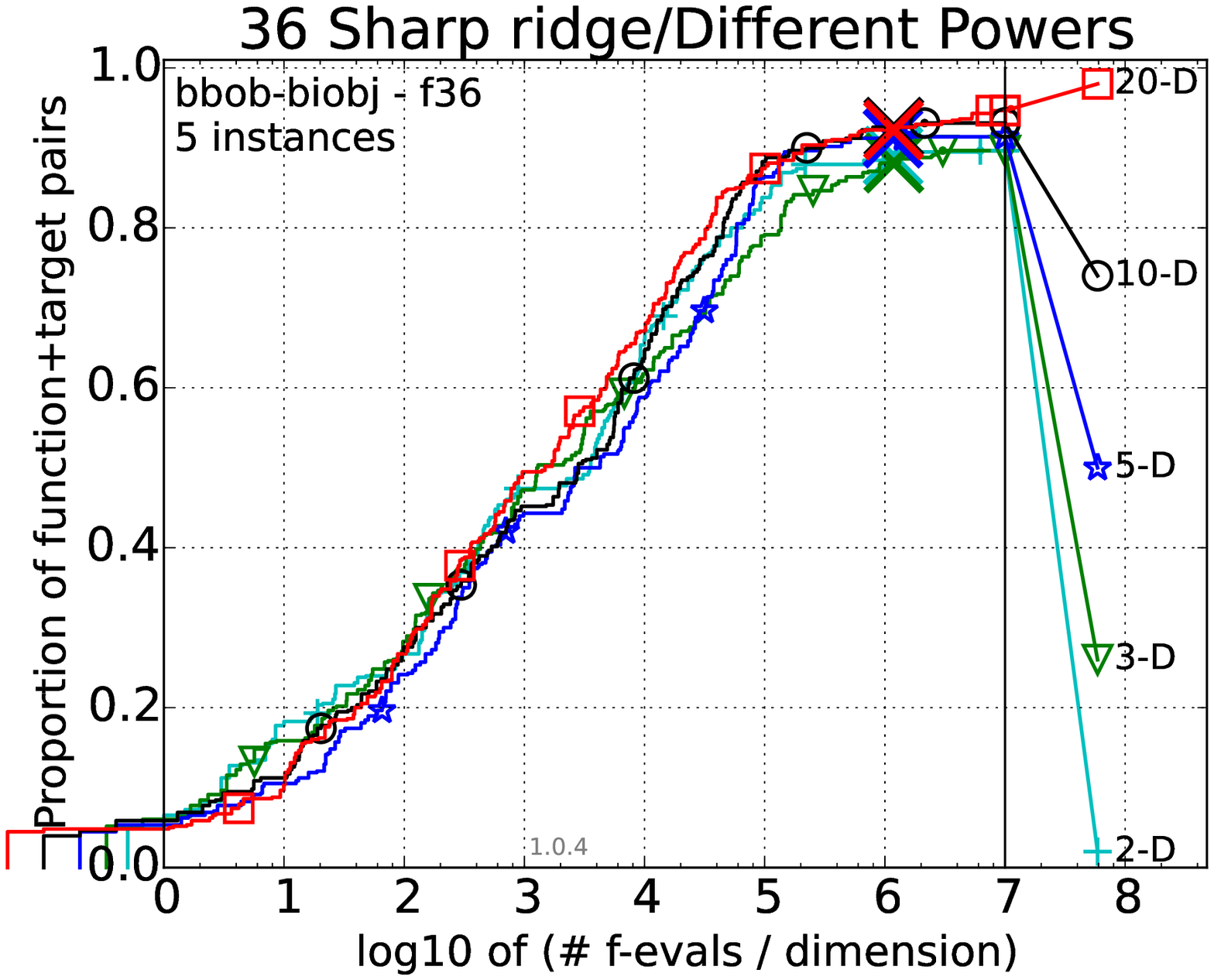}\\[-1.8ex]
\end{tabular}
 \caption{\label{fig:ECDFsingleTwo}
    Empirical cumulative distribution of simulated (bootstrapped) runtimes, 
    measured in number of objective function evaluations, divided by dimension 
    (FEvals/DIM) for the targets as given in Fig.~\ref{fig:ECDFsingleOne} 
    for functions 
    $f_{17}$ to $f_{36}$
    and all dimensions.
%
% Empirical cumulative distribution function (ECDF) per dimension for all 
% targets of each function as in Fig.~\ref{fig:ECDFsingleOne} but for $f_{17}$ till $f_{36}$.
 }
\end{figure*}

\begin{figure*}
\centering
\begin{tabular}{@{\hspace*{-0.018\textwidth}}l@{\hspace*{-0.02\textwidth}}l@{\hspace*{-0.02\textwidth}}l@{\hspace*{-0.02\textwidth}}l@{\hspace*{-0.02\textwidth}}l@{\hspace*{-0.02\textwidth}}}
\includegraphics[width=0.25\textwidth]{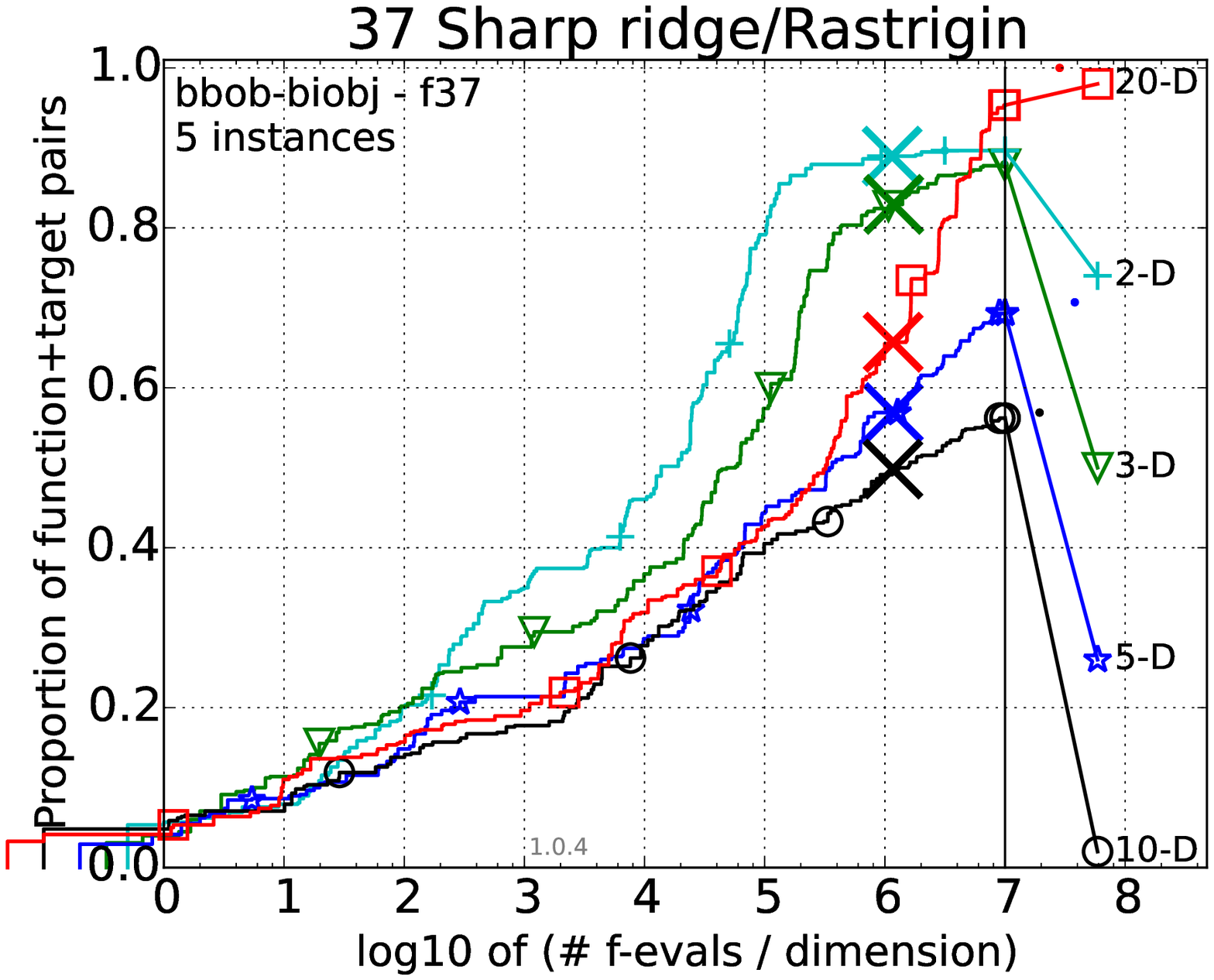}&
\includegraphics[width=0.25\textwidth]{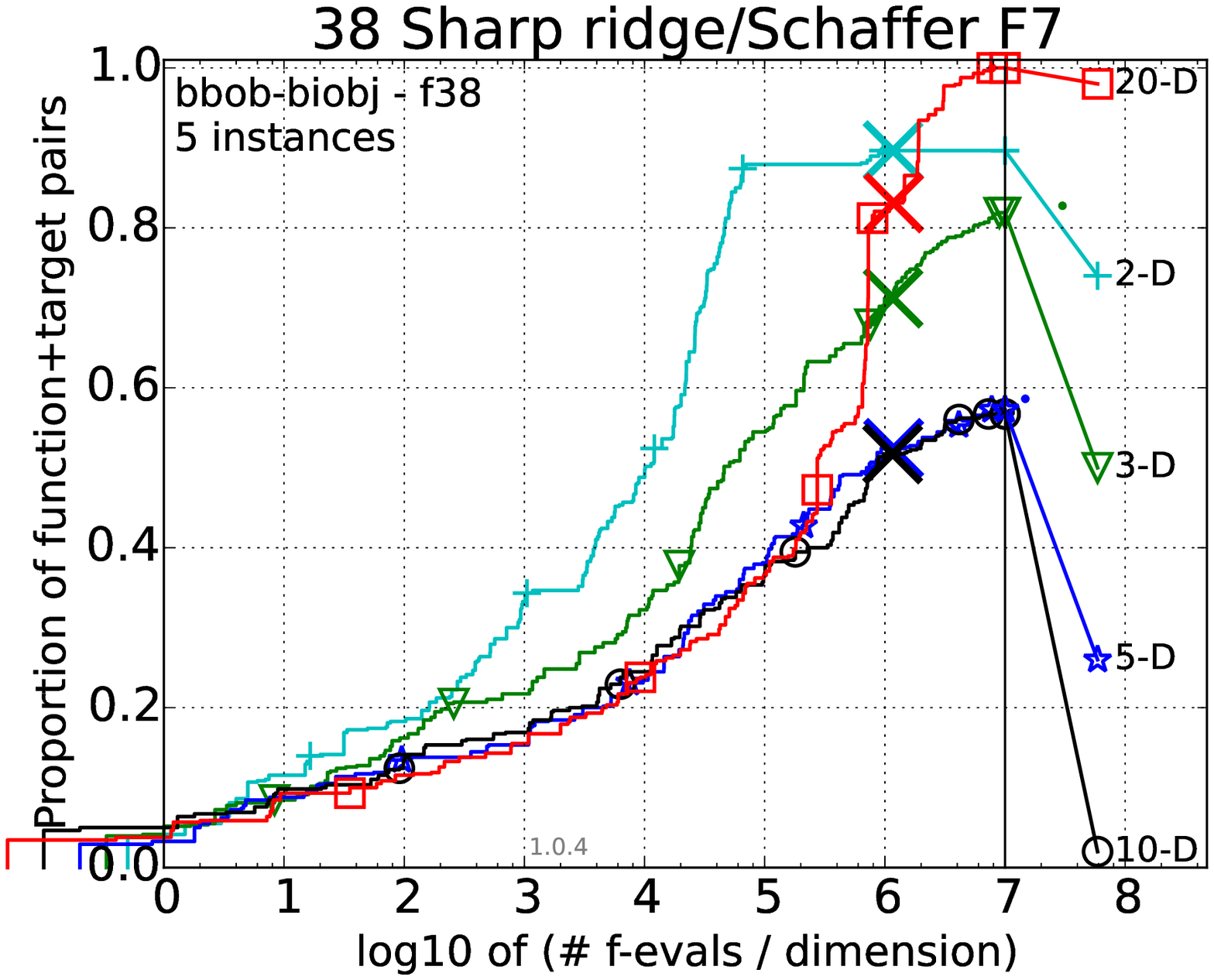}&
\includegraphics[width=0.25\textwidth]{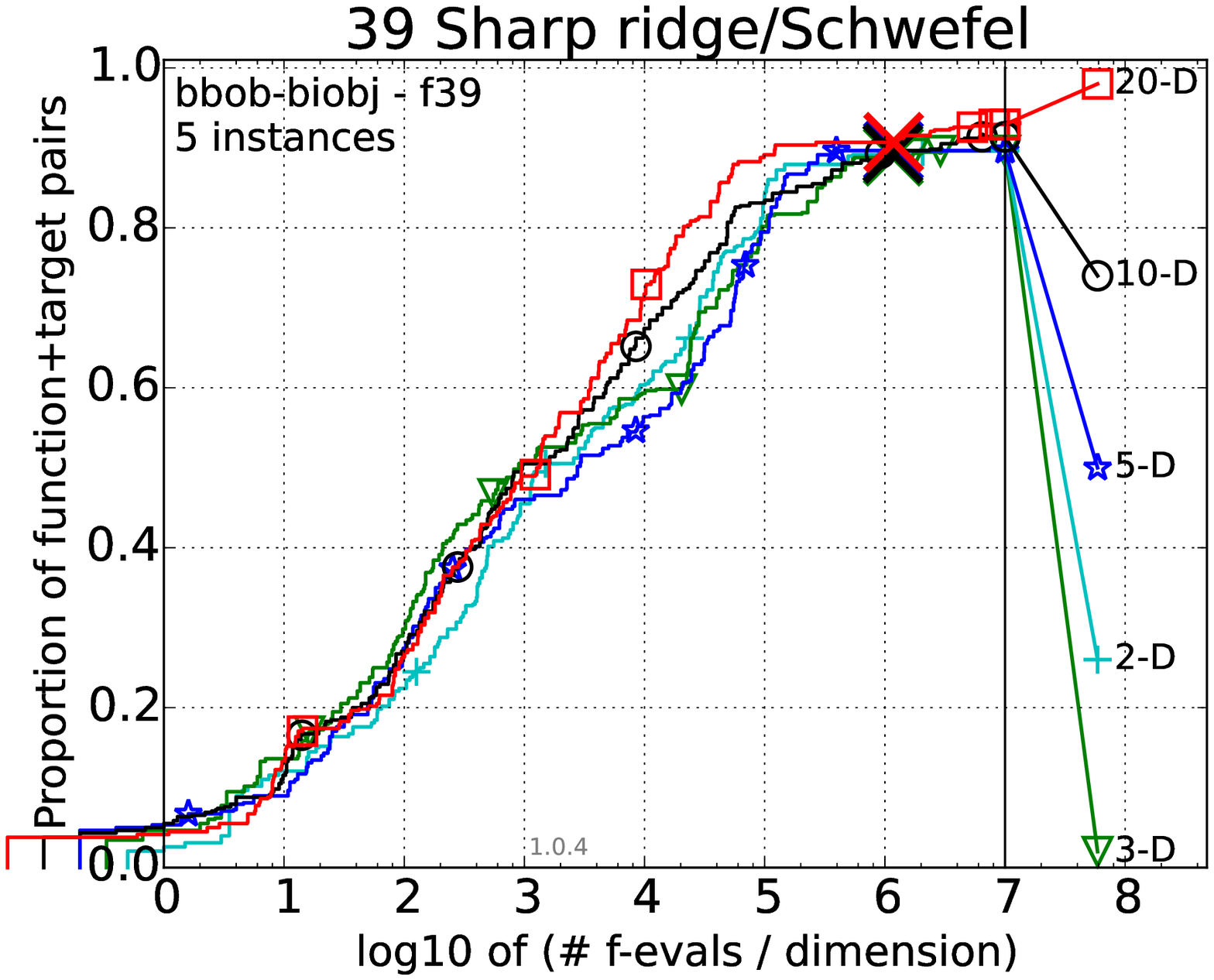}&
\includegraphics[width=0.25\textwidth]{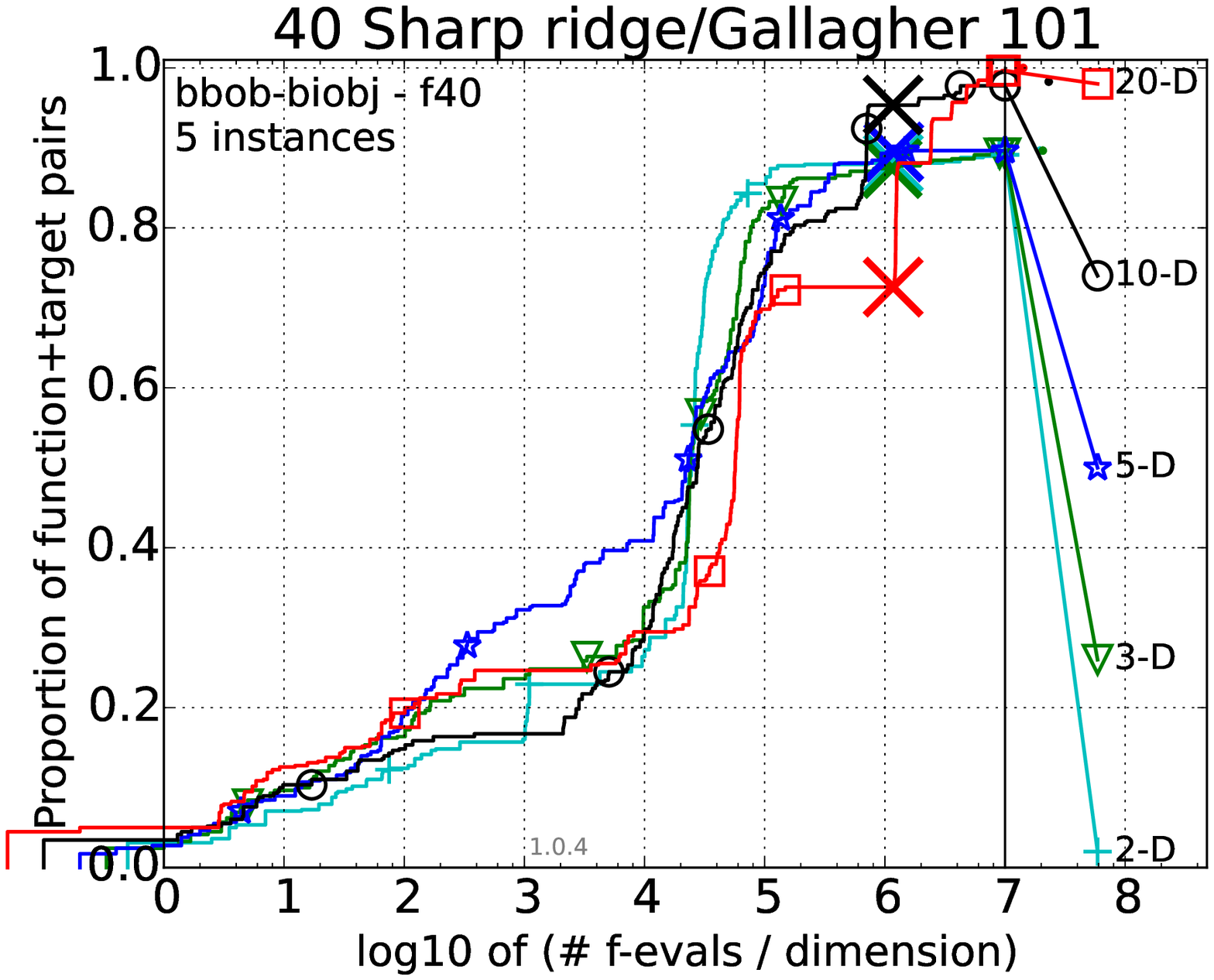}\\[-1.8ex]
\includegraphics[width=0.25\textwidth]{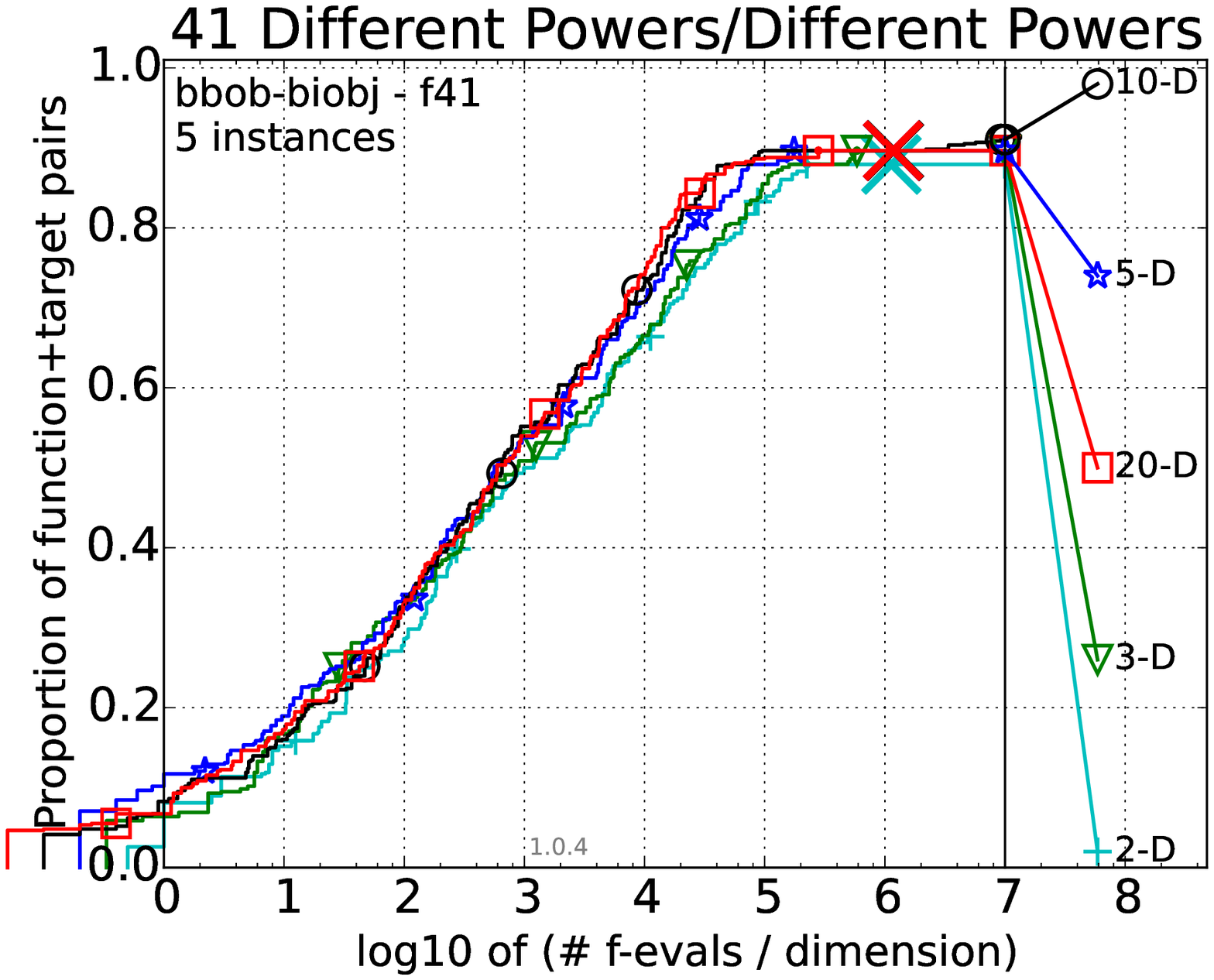}&
\includegraphics[width=0.25\textwidth]{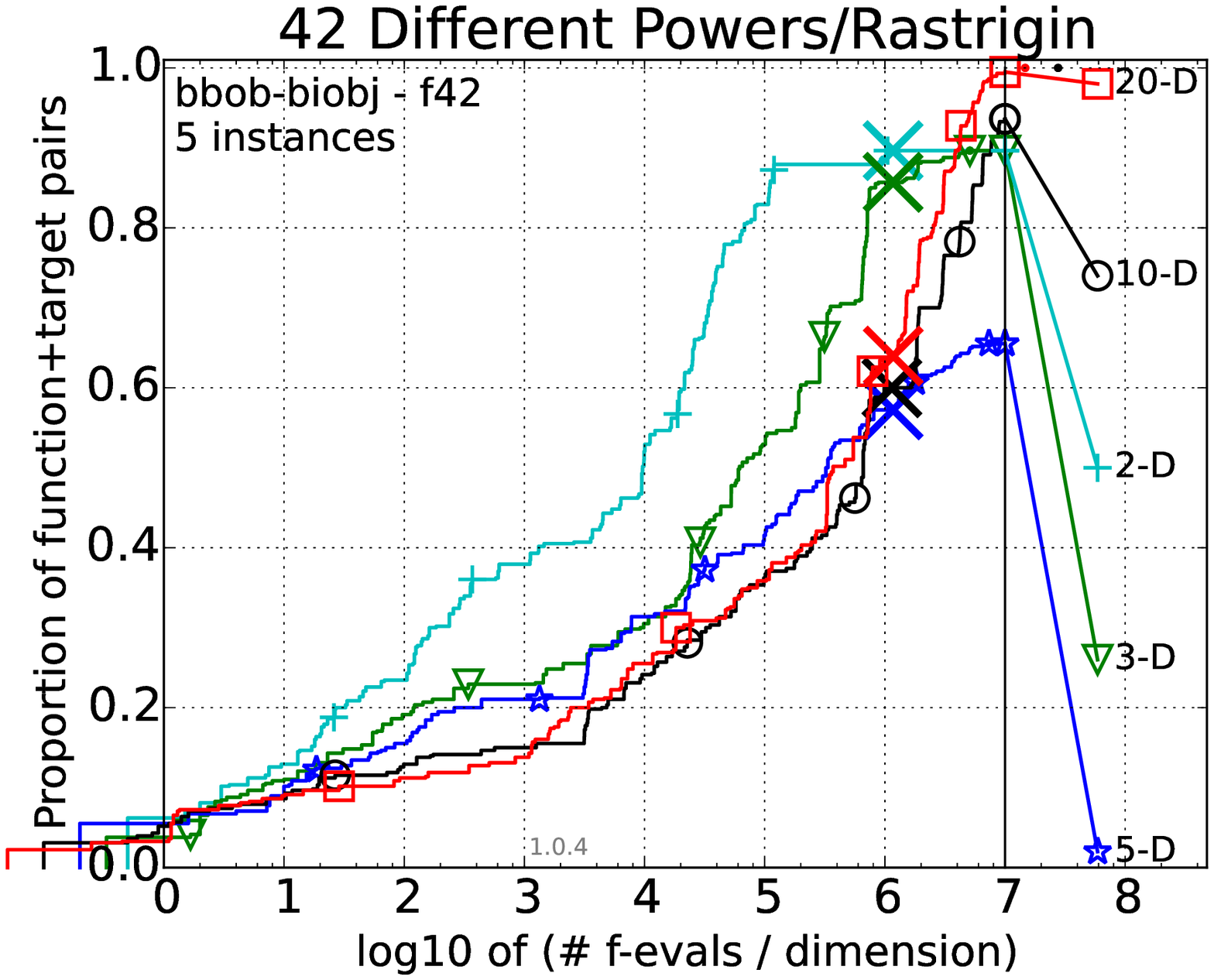}&
\includegraphics[width=0.25\textwidth]{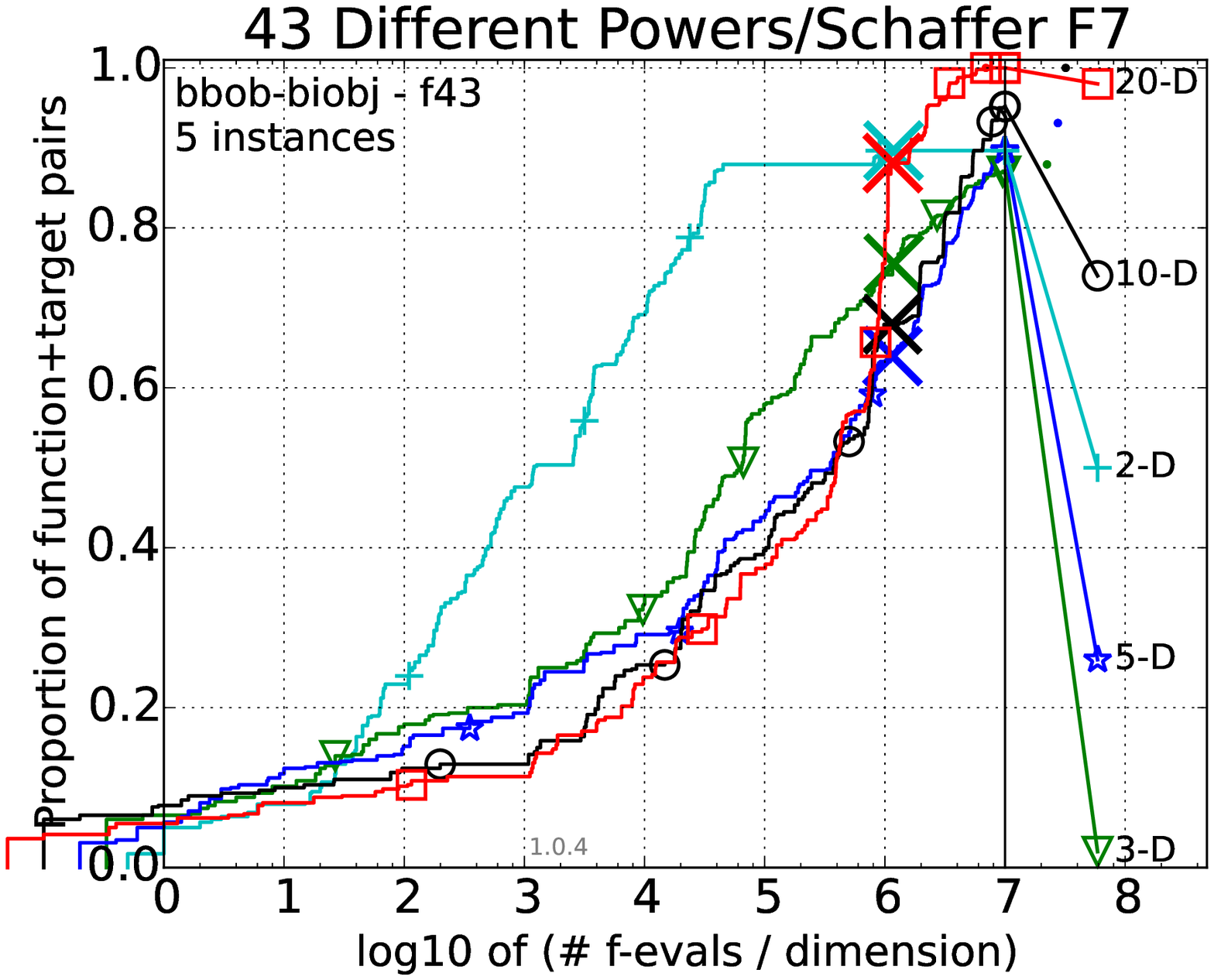}&
\includegraphics[width=0.25\textwidth]{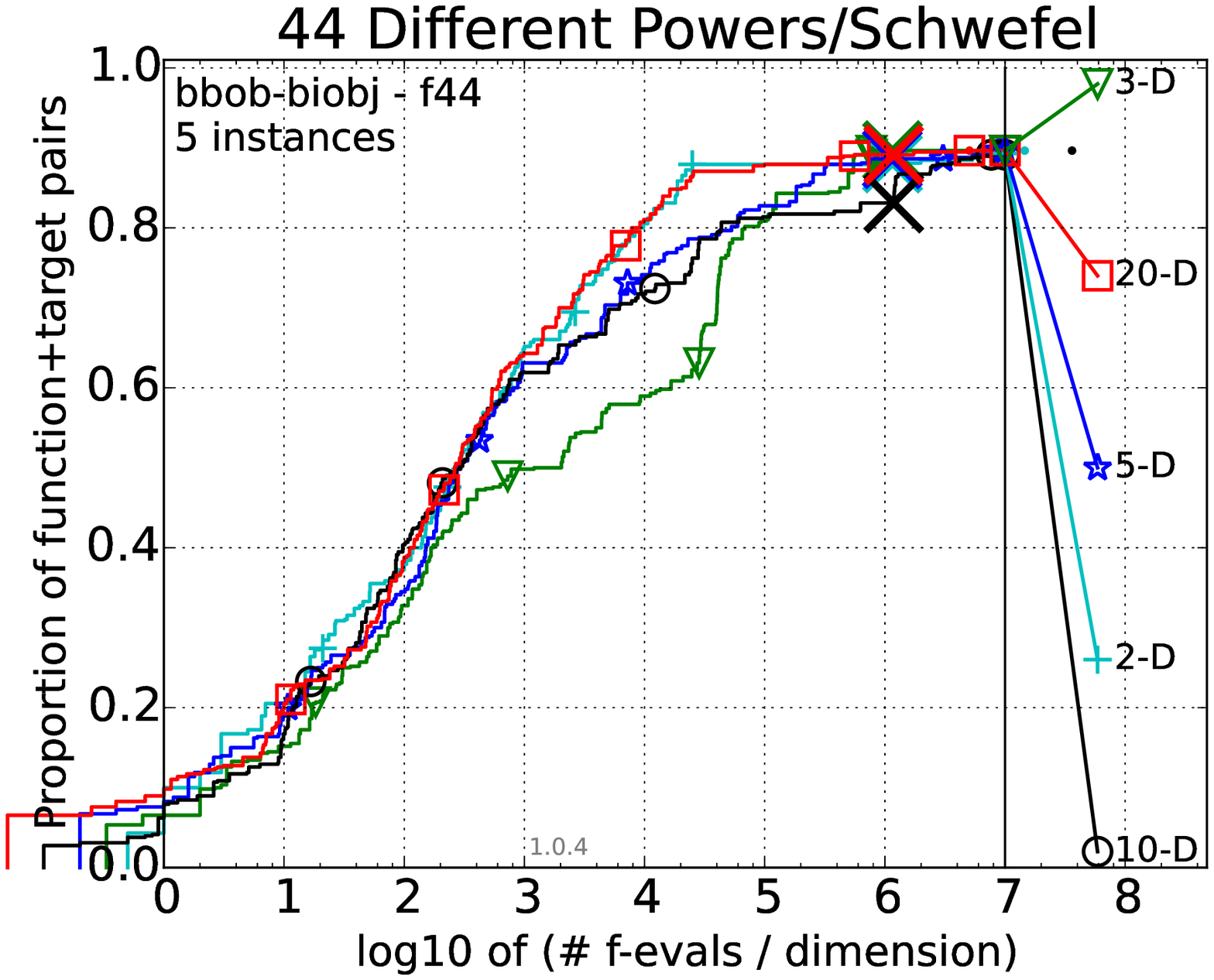}\\[-1.8ex]
\includegraphics[width=0.25\textwidth]{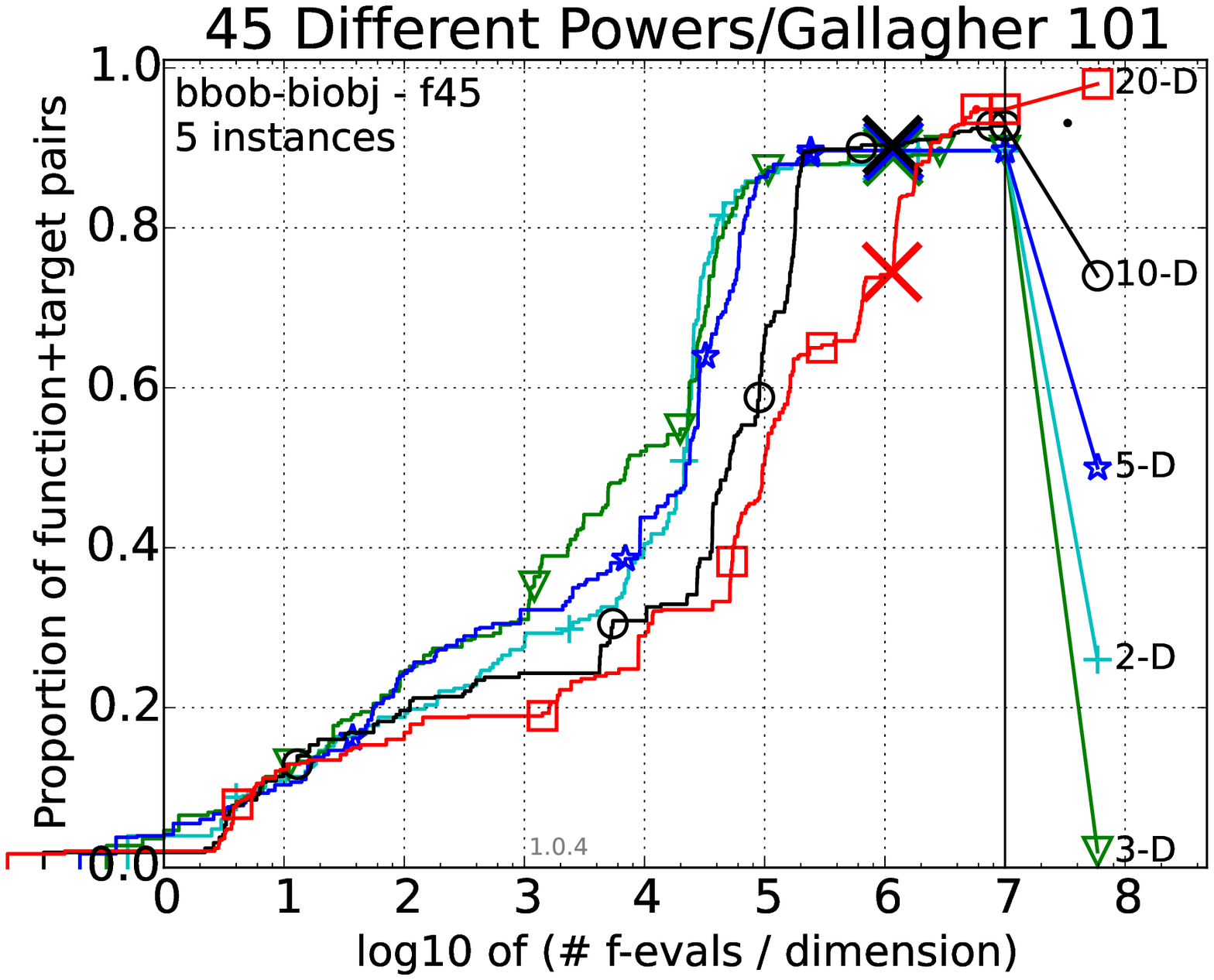}&
\includegraphics[width=0.25\textwidth]{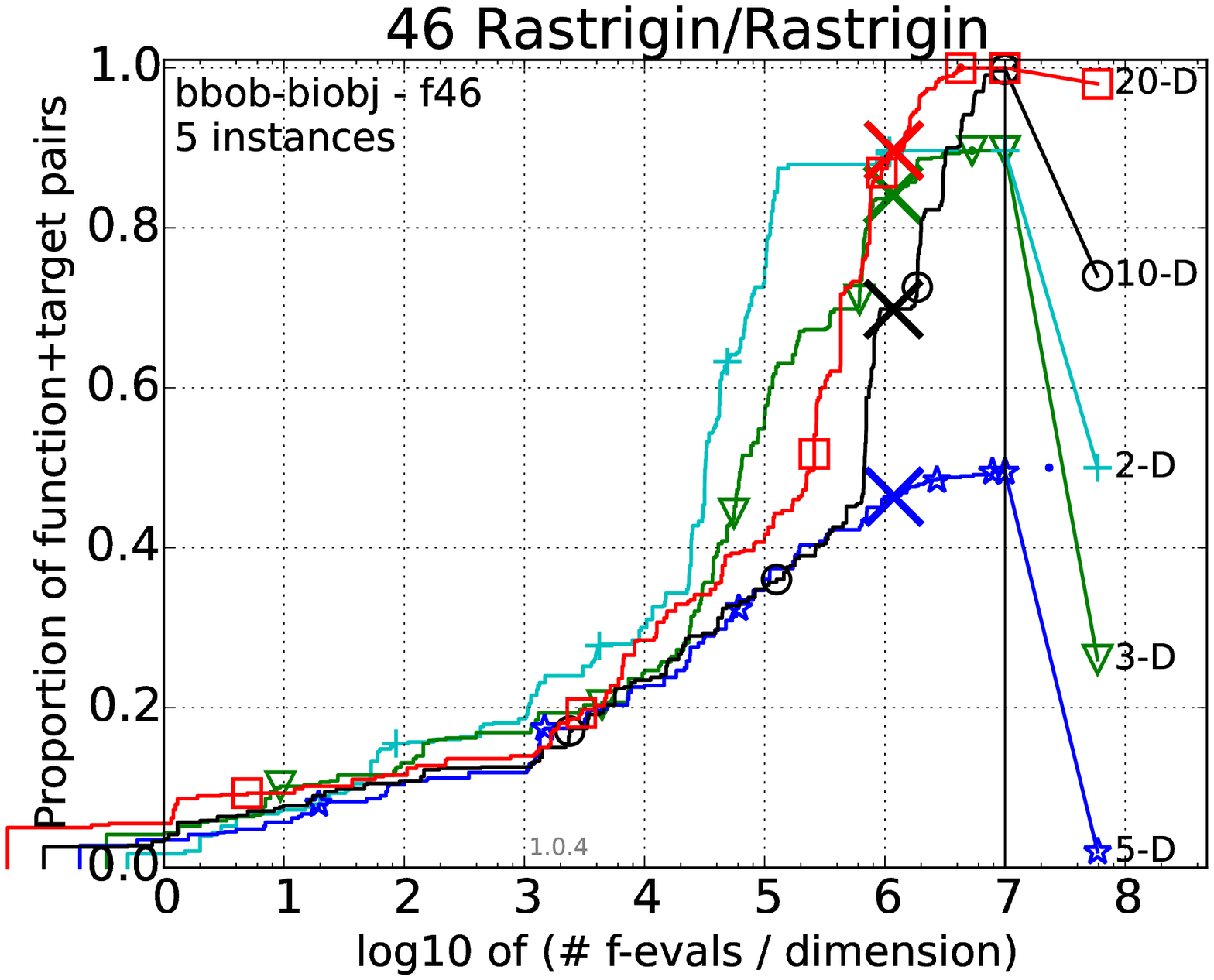}&
\includegraphics[width=0.25\textwidth]{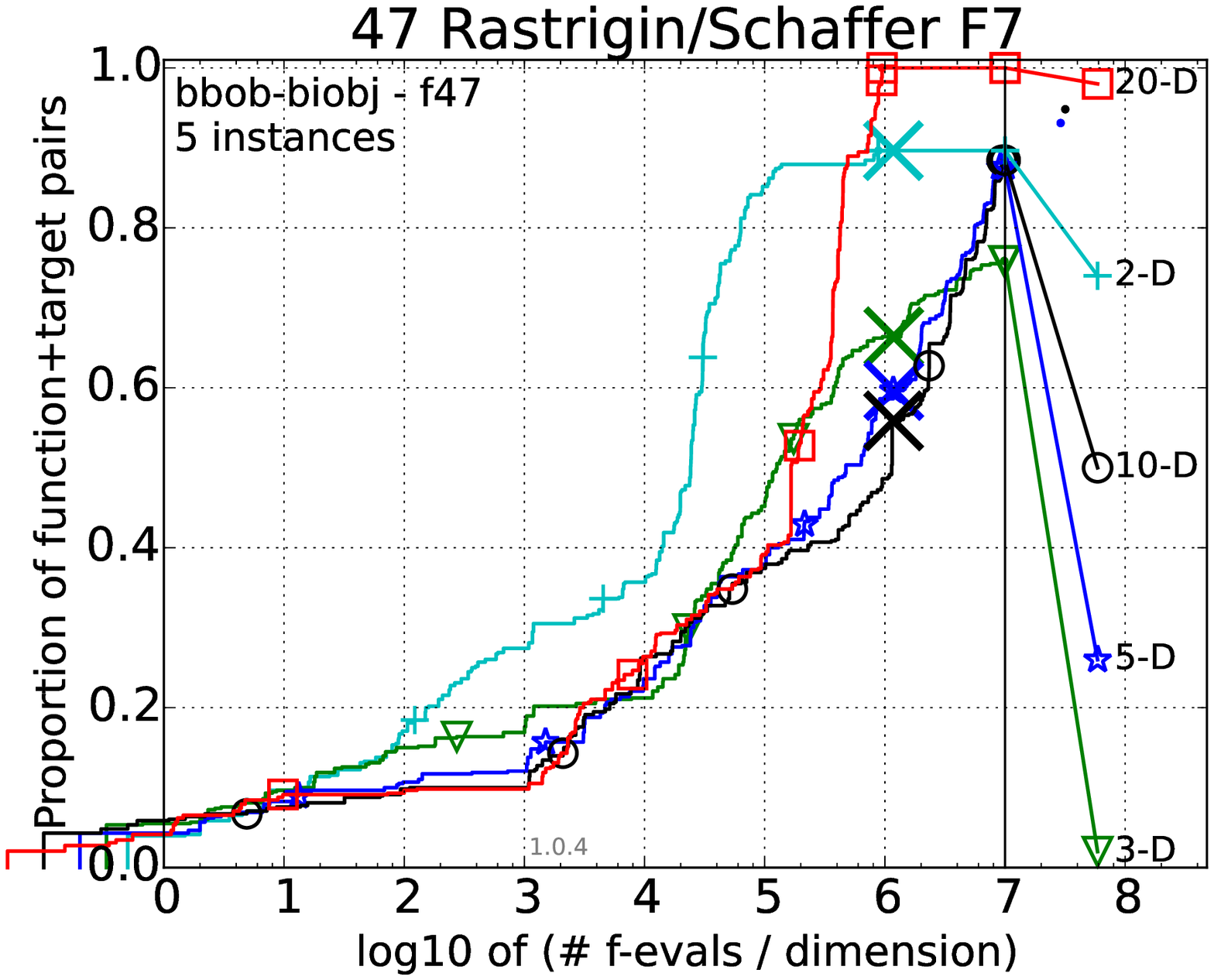}&
\includegraphics[width=0.25\textwidth]{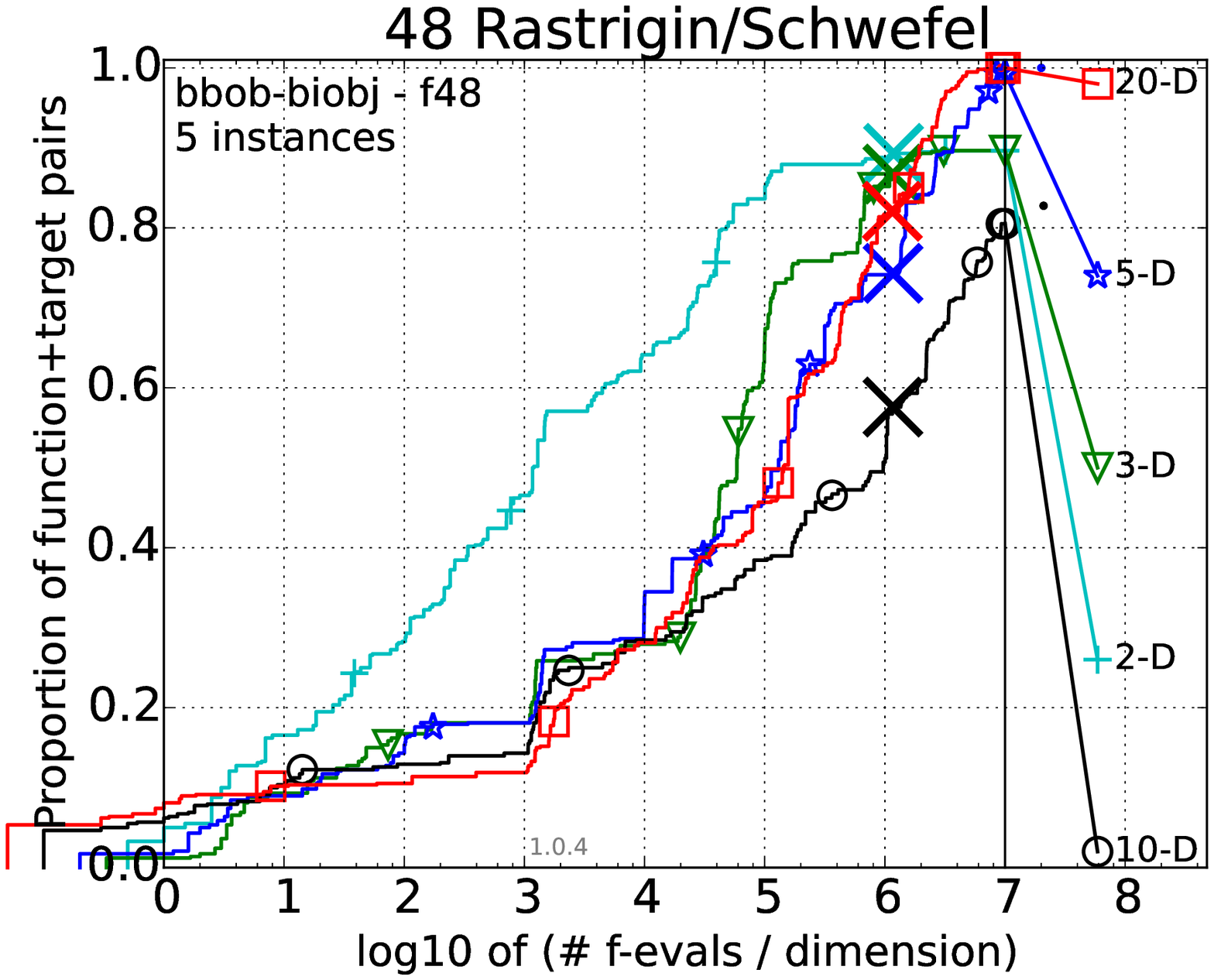}\\[-1.8ex]
\includegraphics[width=0.25\textwidth]{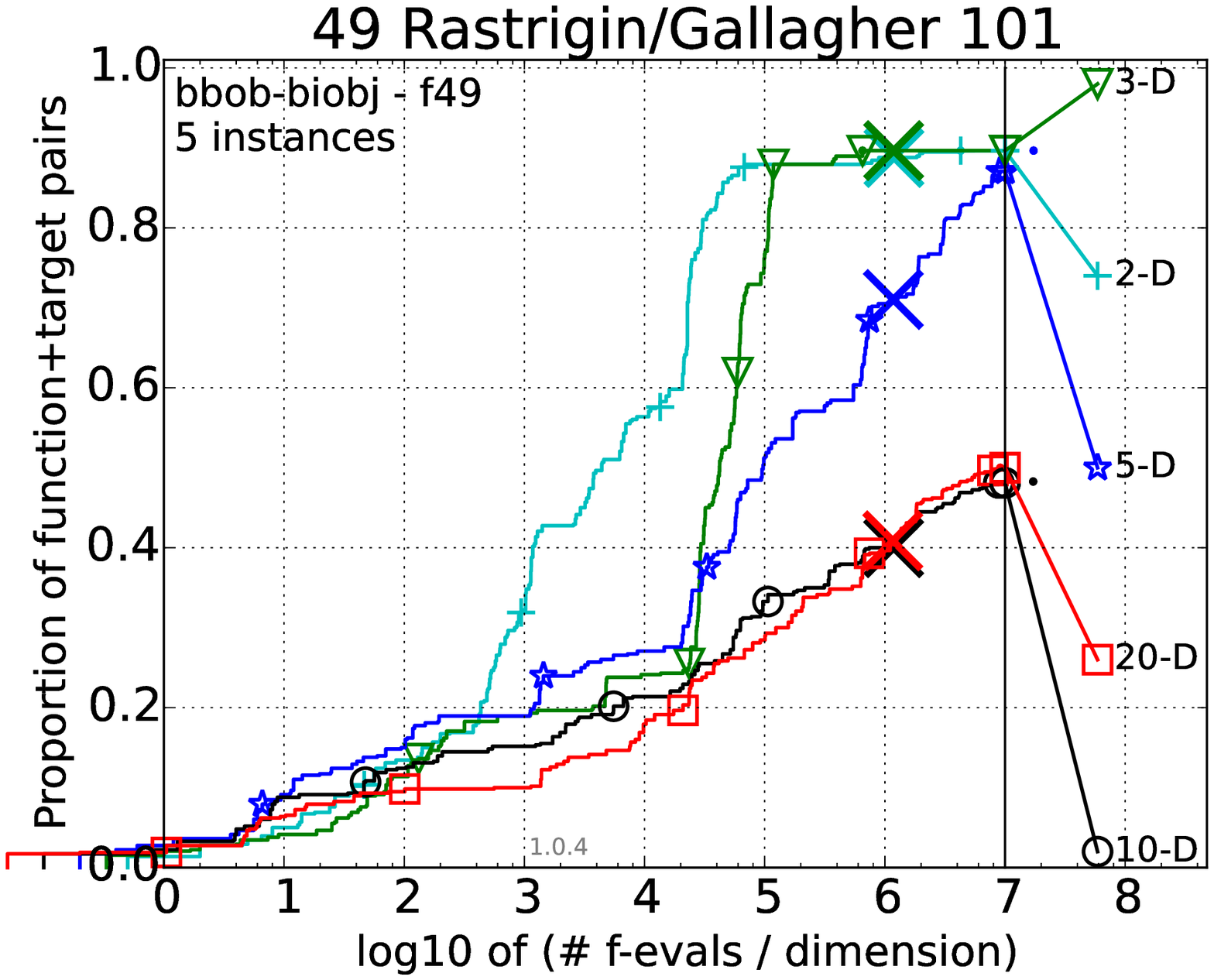}&
\includegraphics[width=0.25\textwidth]{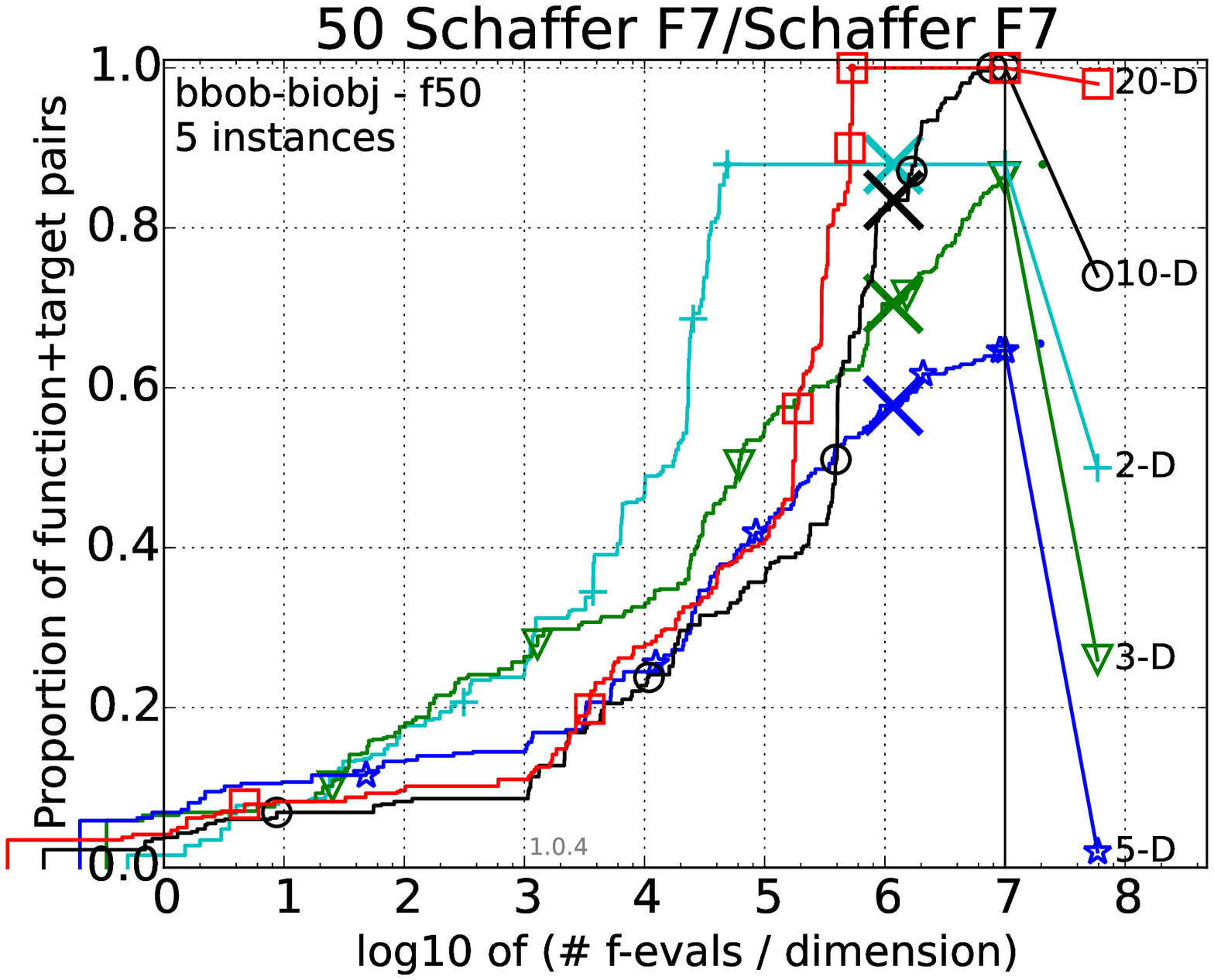}&
\includegraphics[width=0.25\textwidth]{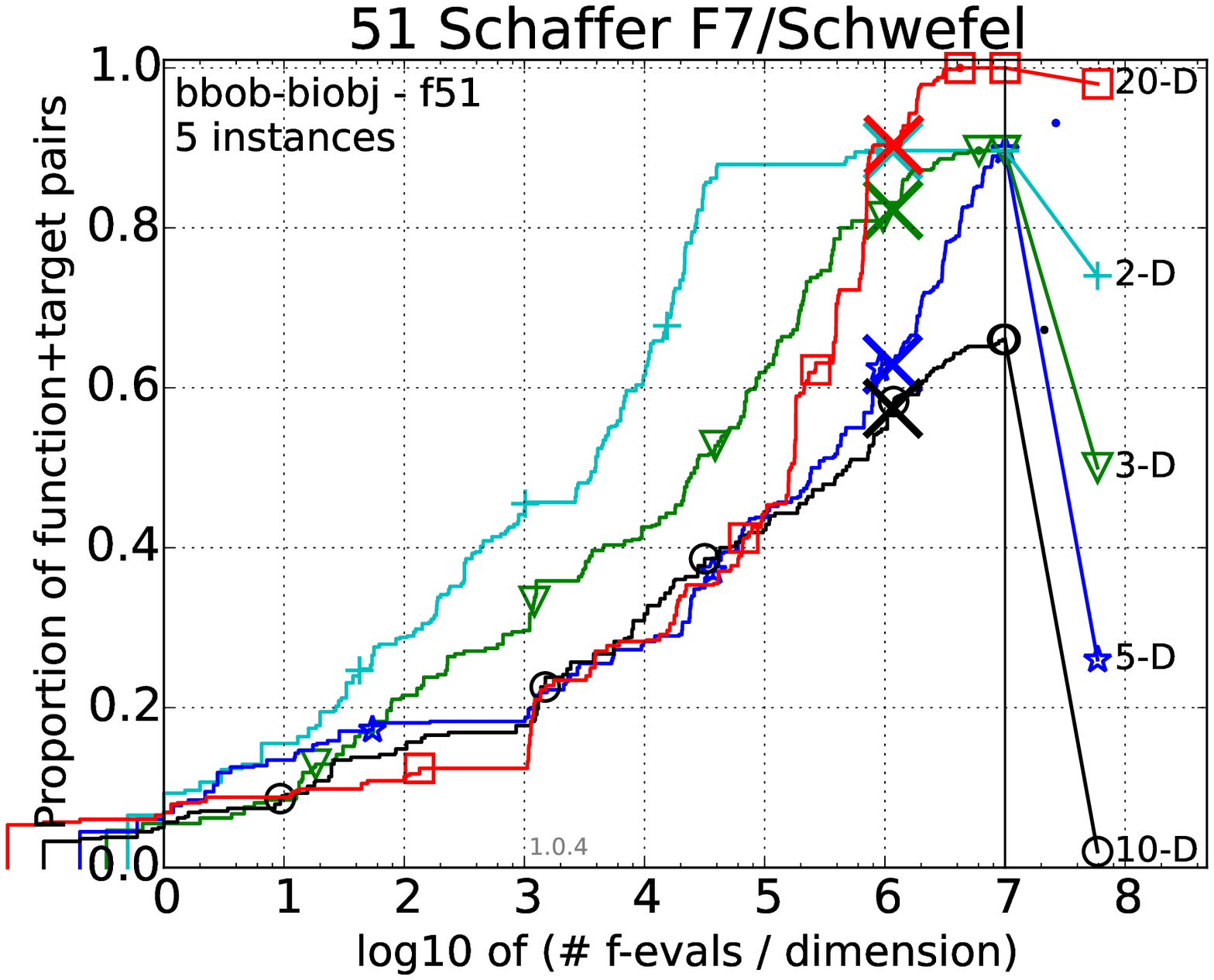}&
\includegraphics[width=0.25\textwidth]{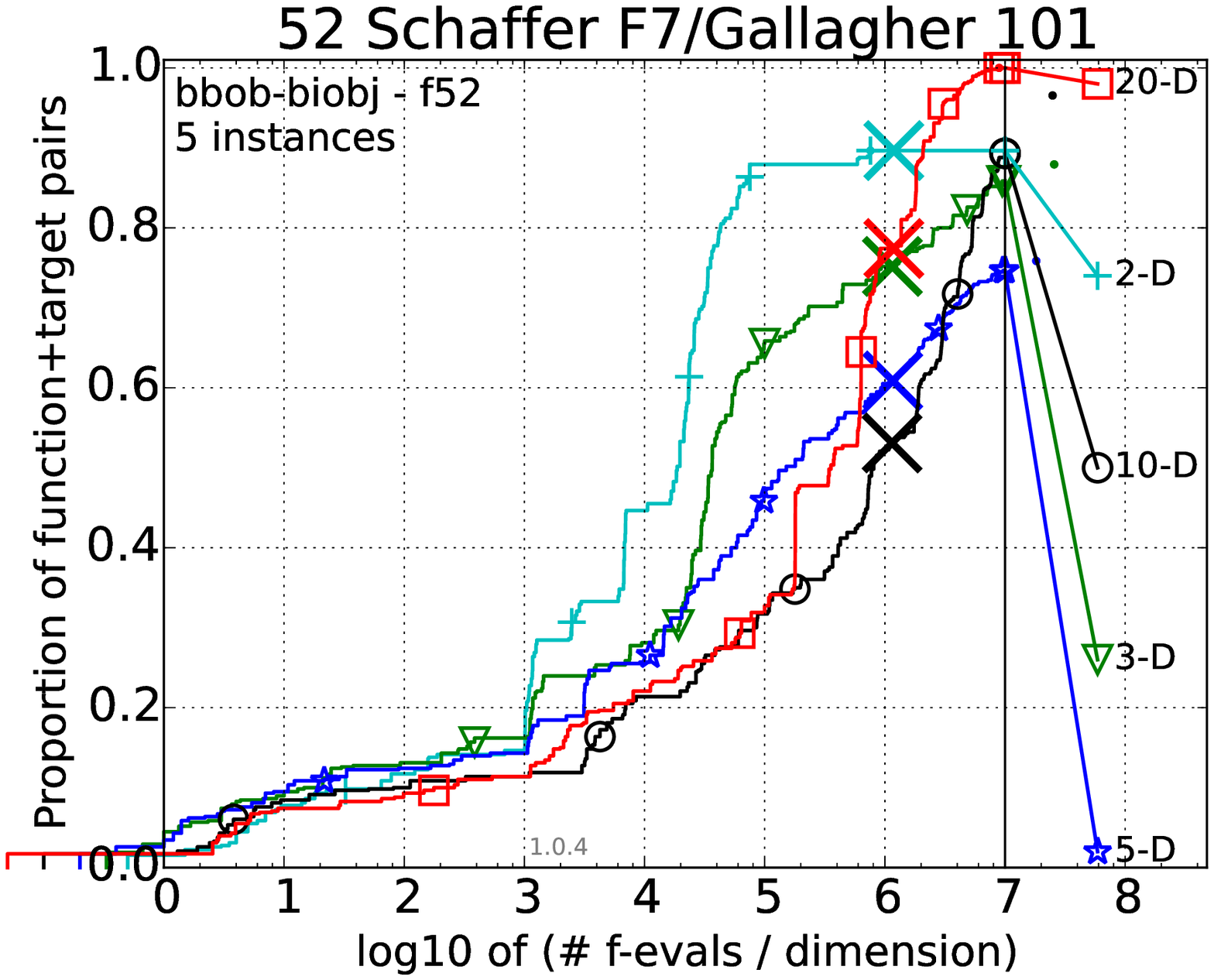}
\end{tabular}
\begin{tabular}{@{\hspace*{-0.018\textwidth}}l@{\hspace*{-0.02\textwidth}}l@{\hspace*{-0.02\textwidth}}l@{\hspace*{-0.02\textwidth}}l@{\hspace*{-0.02\textwidth}}}
\includegraphics[width=0.25\textwidth]{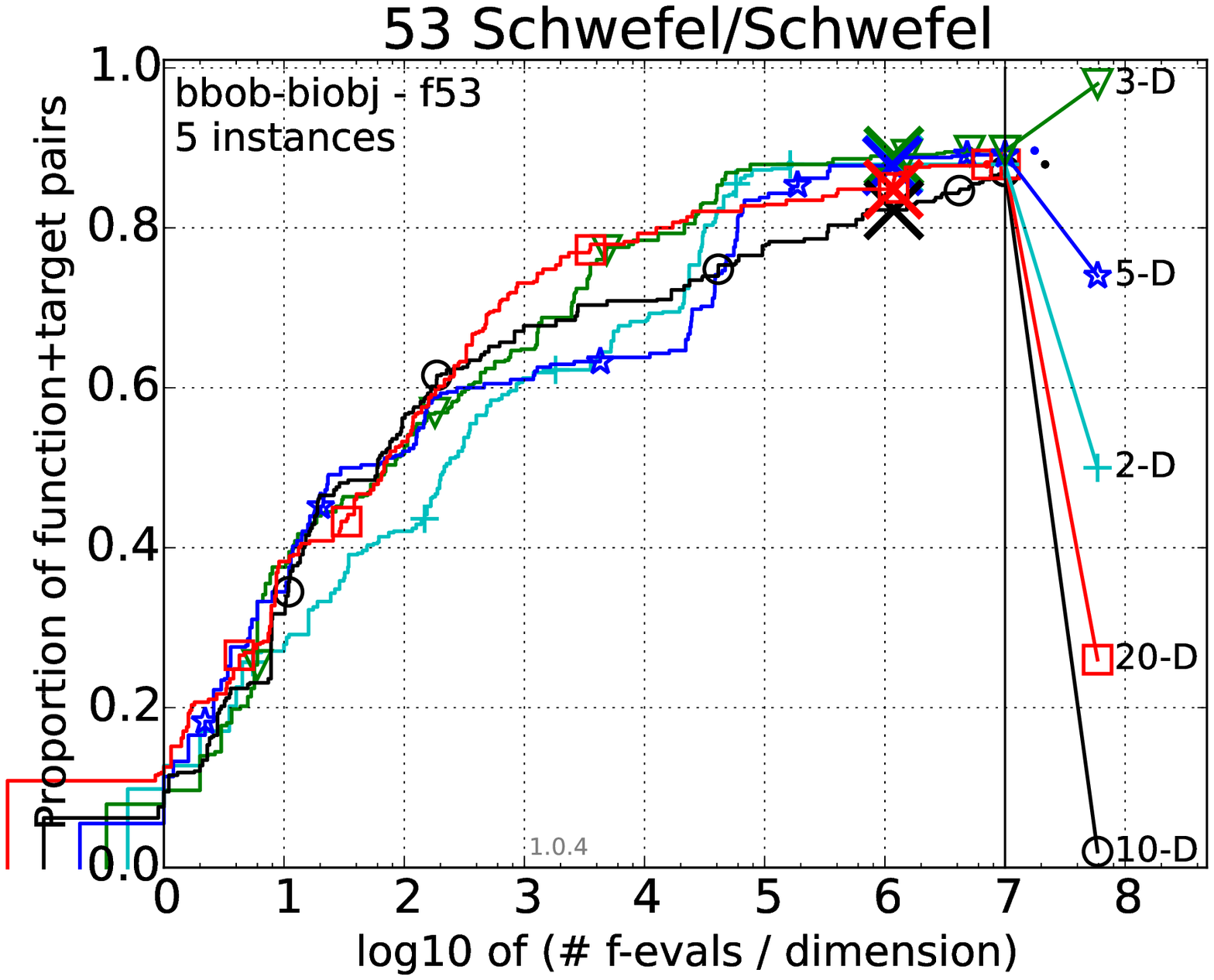}&
\includegraphics[width=0.25\textwidth]{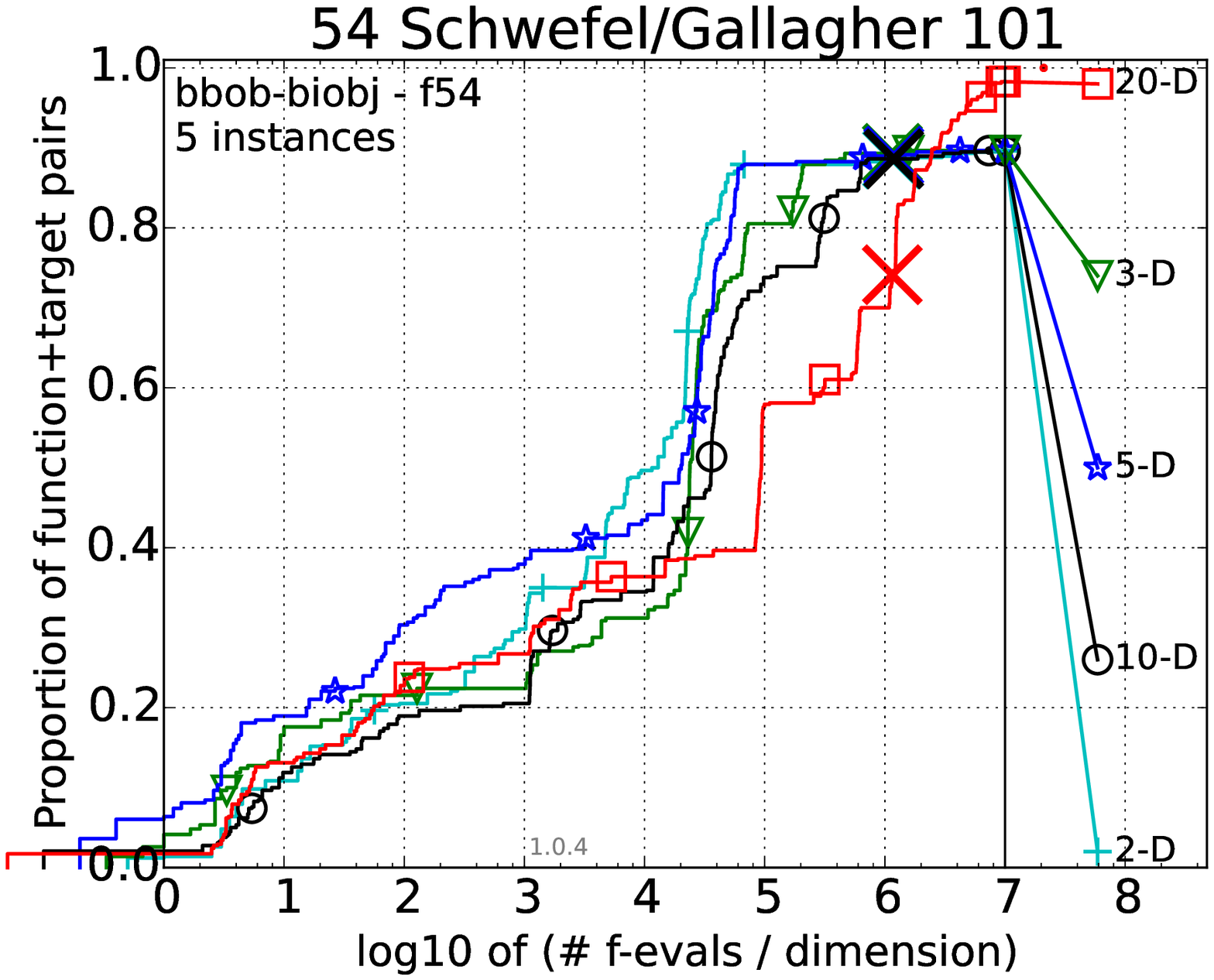}&
\includegraphics[width=0.25\textwidth]{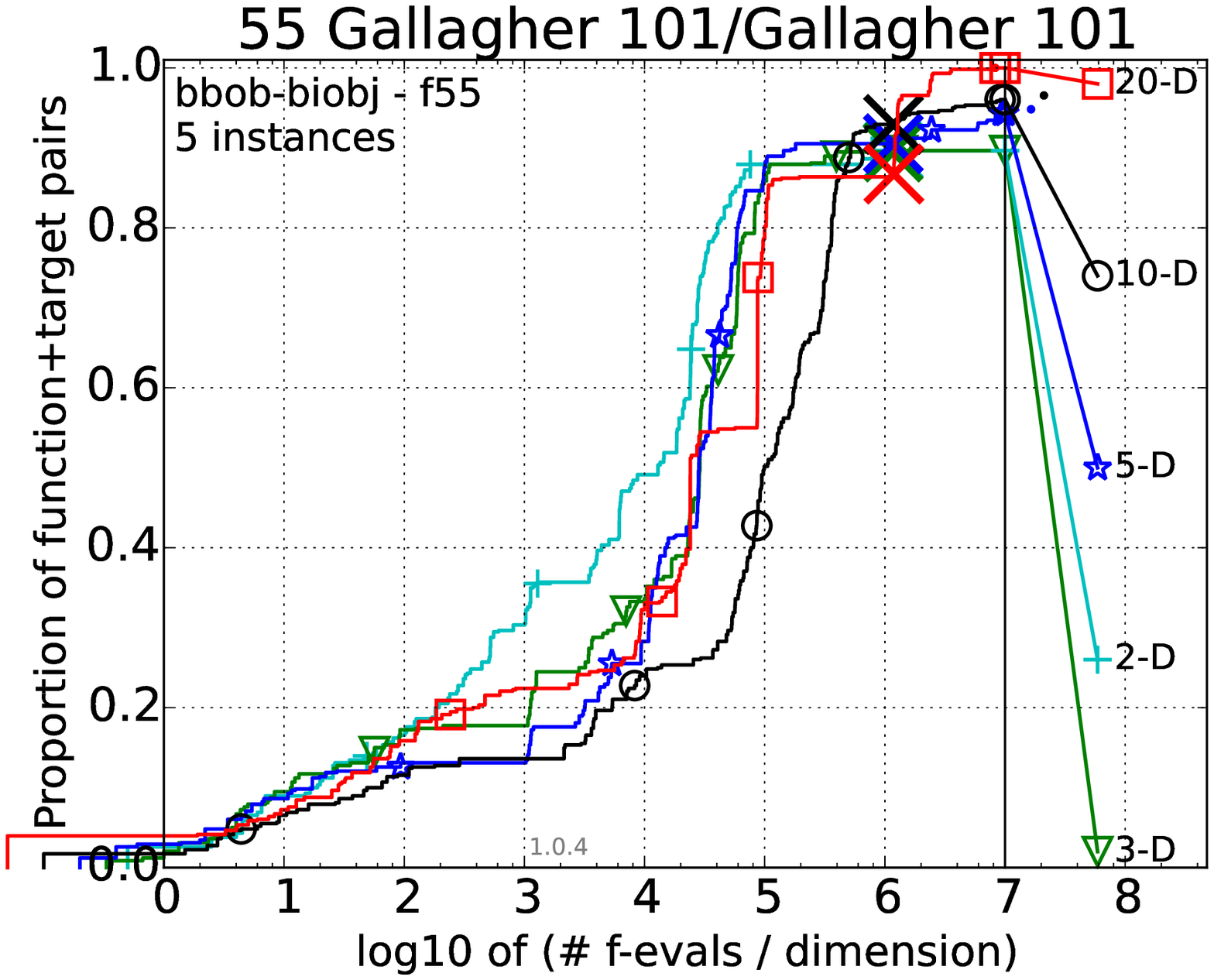}\\[-1.8ex]
\end{tabular}
 \caption{\label{fig:ECDFsingleThree}
    Empirical cumulative distribution of simulated (bootstrapped) runtimes, 
    measured in number of objective function evaluations, divided by dimension 
    (FEvals/DIM) for the targets as given in Fig.~\ref{fig:ECDFsingleOne} 
    for functions 
    $f_{37}$ to $f_{55}$
    and all dimensions. 
% Empirical cumulative distribution function (ECDF) per dimension for all targets of each function as in Fig.~\ref{fig:ECDFsingleOne} but for $f_{37}$ till $f_{55}$.
 }
\end{figure*}

\begin{figure*}
\begin{tabular}{c@{\hspace*{-0.02\textwidth}}c@{\hspace*{-0.02\textwidth}}c@{\hspace*{-0.02\textwidth}}c}
separable-separable & separable-moderate & separable-ill-cond. & separable-multimodal\\
\includegraphics[width=0.268\textwidth,trim=0 0 0 13mm, clip]{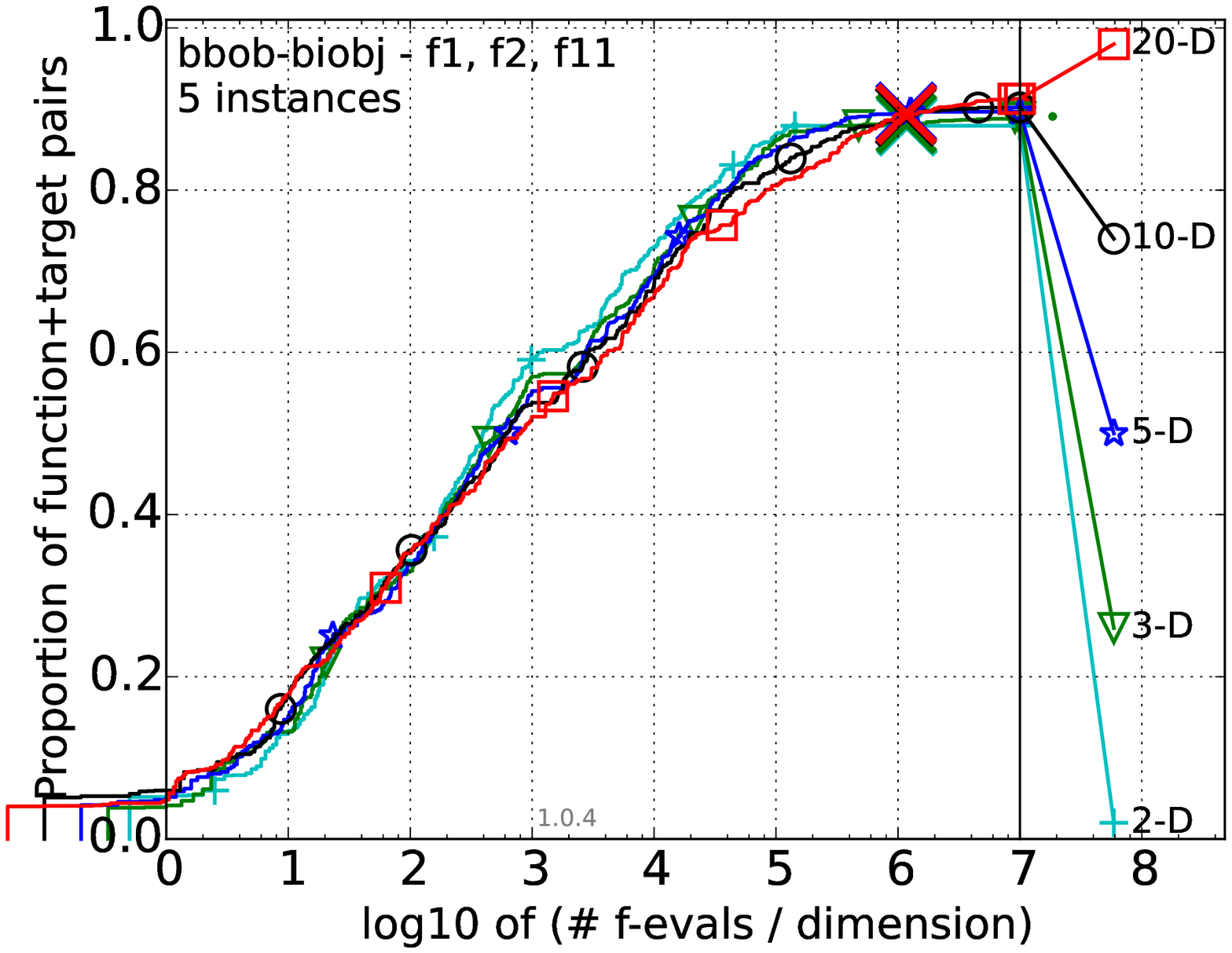} &
\includegraphics[width=0.268\textwidth,trim=0 0 0 13mm, clip]{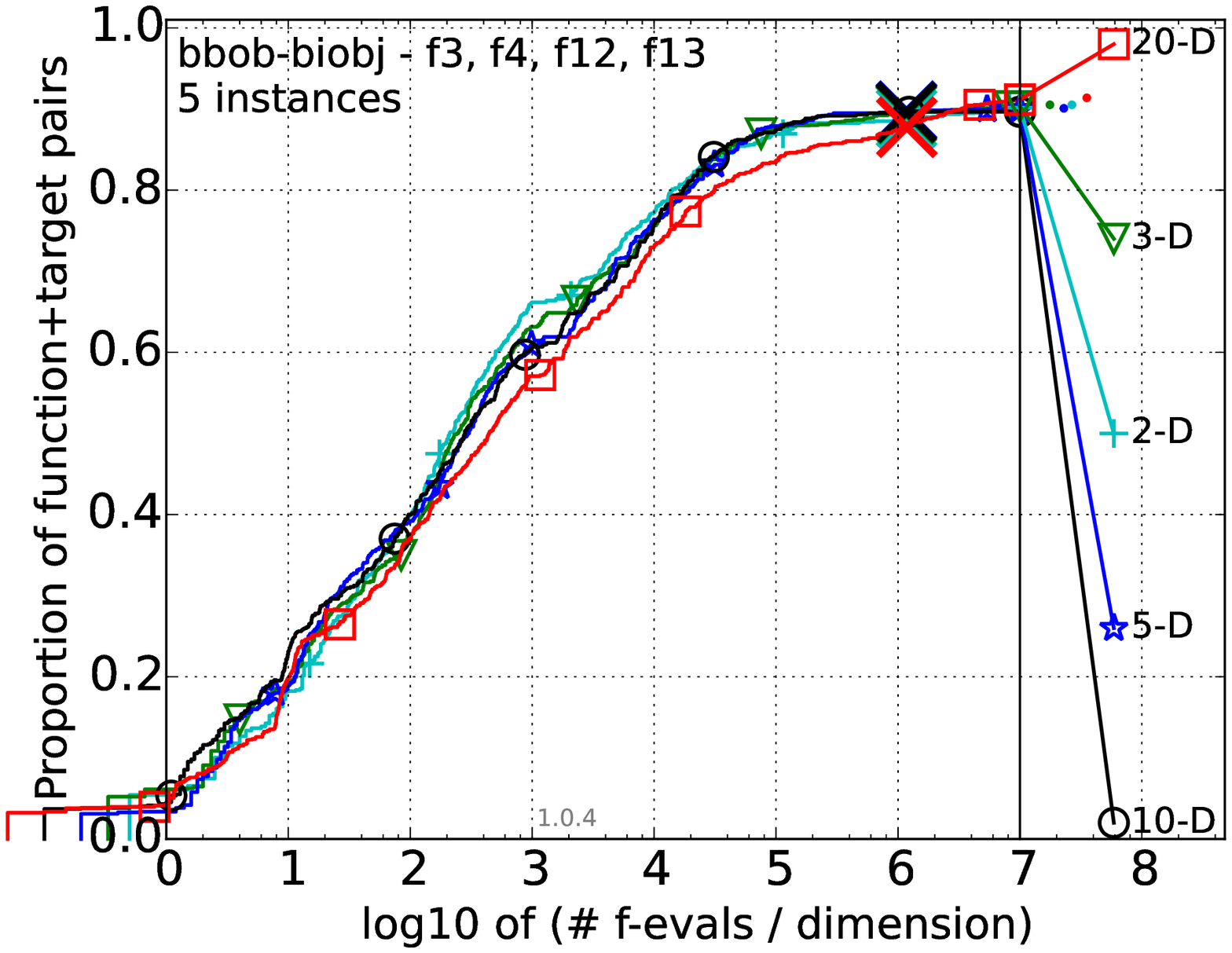} &
\includegraphics[width=0.268\textwidth,trim=0 0 0 13mm, clip]{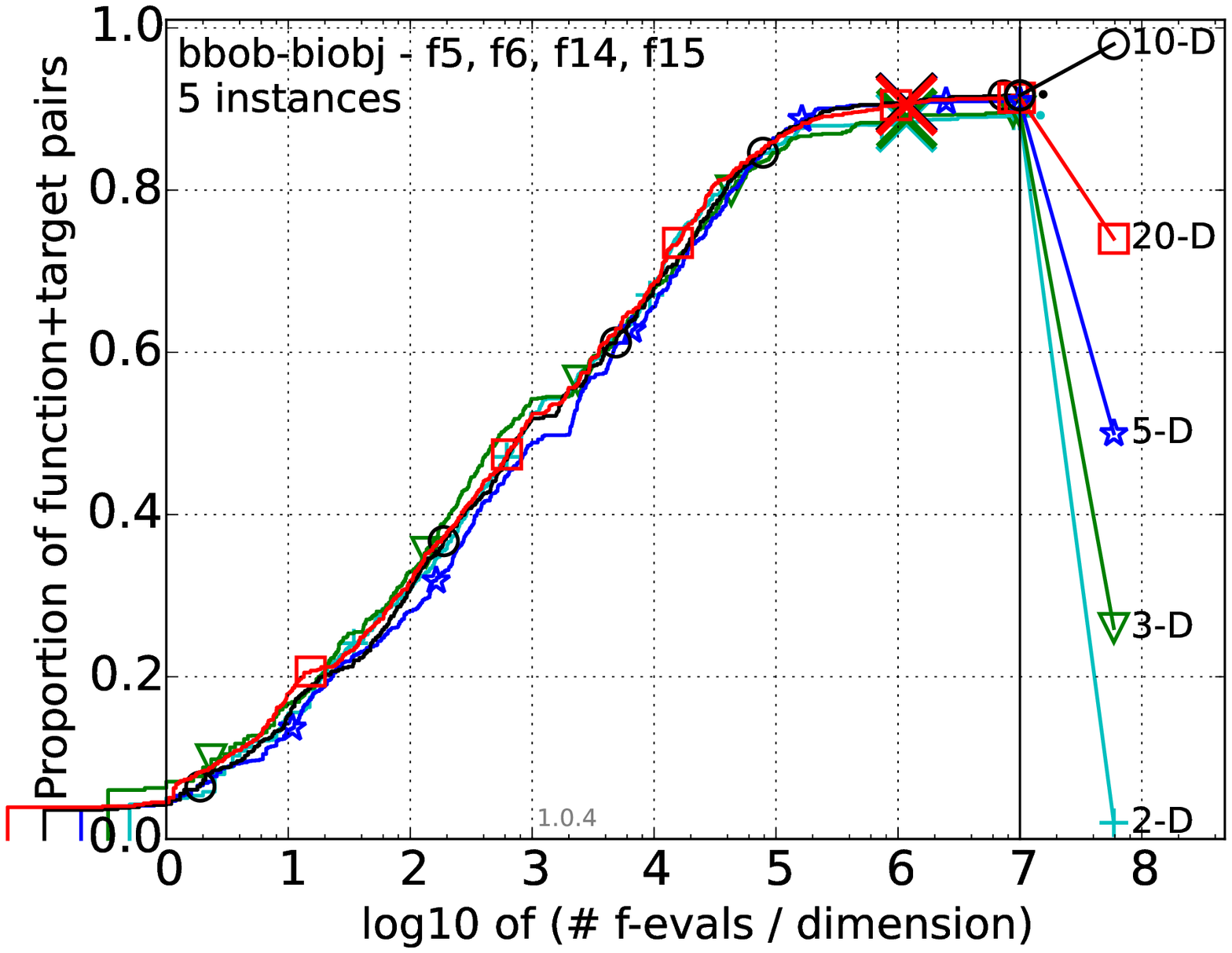} &
\includegraphics[width=0.268\textwidth,trim=0 0 0 13mm, clip]{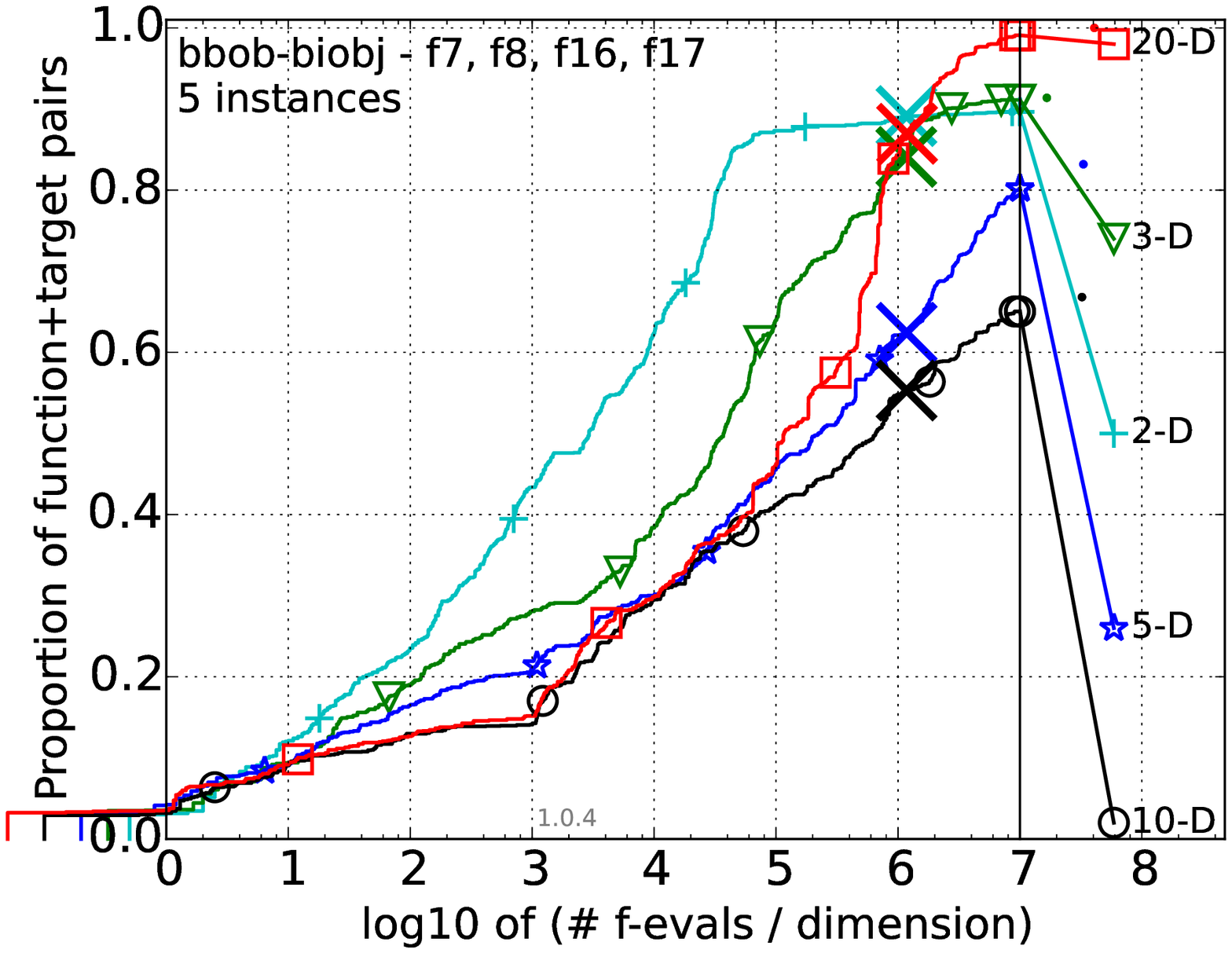}\\
separable-weakstructure & moderate-moderate & moderate-ill-cond. & moderate-multimodal\\
\includegraphics[width=0.268\textwidth,trim=0 0 0 13mm, clip]{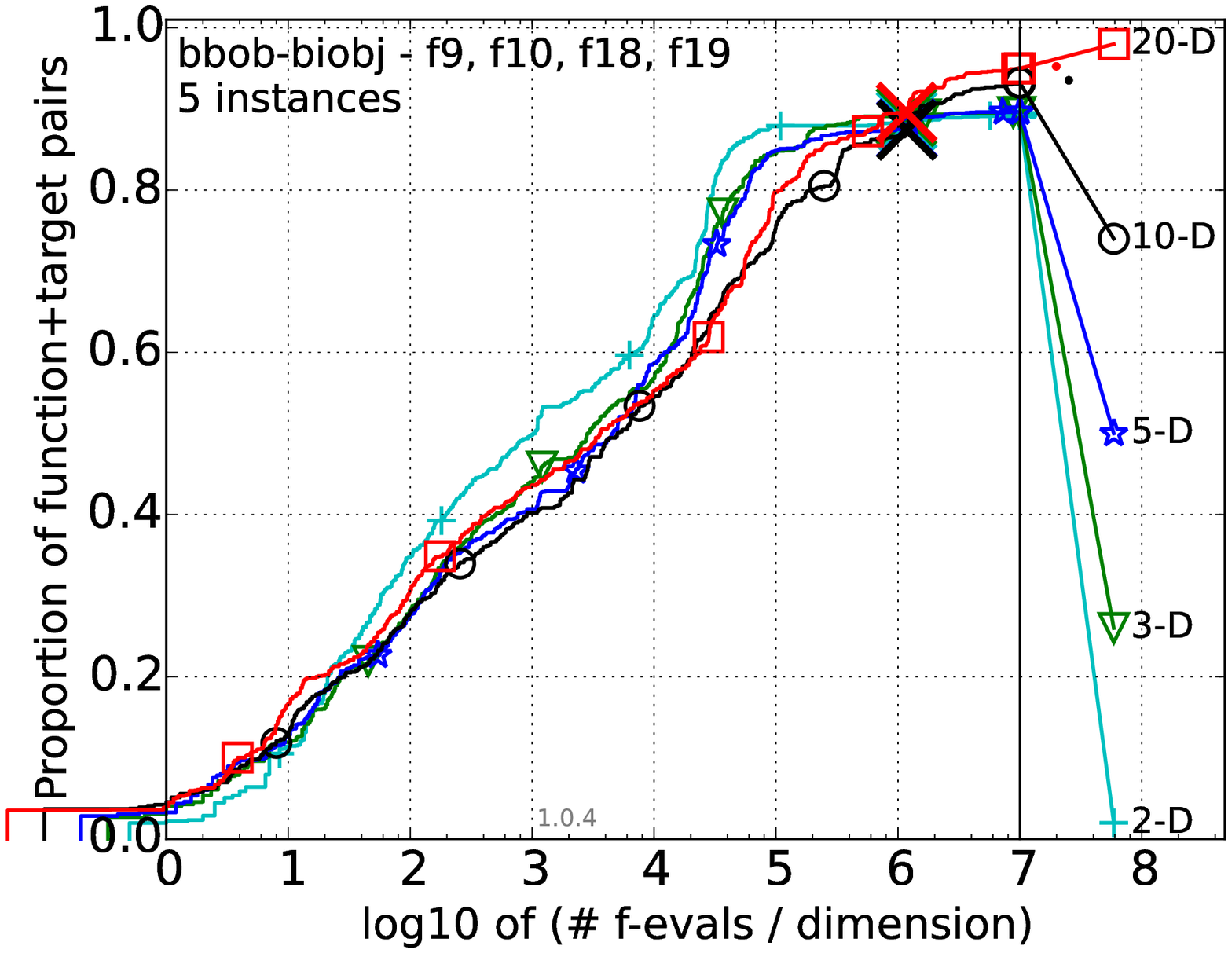} &
\includegraphics[width=0.268\textwidth,trim=0 0 0 13mm, clip]{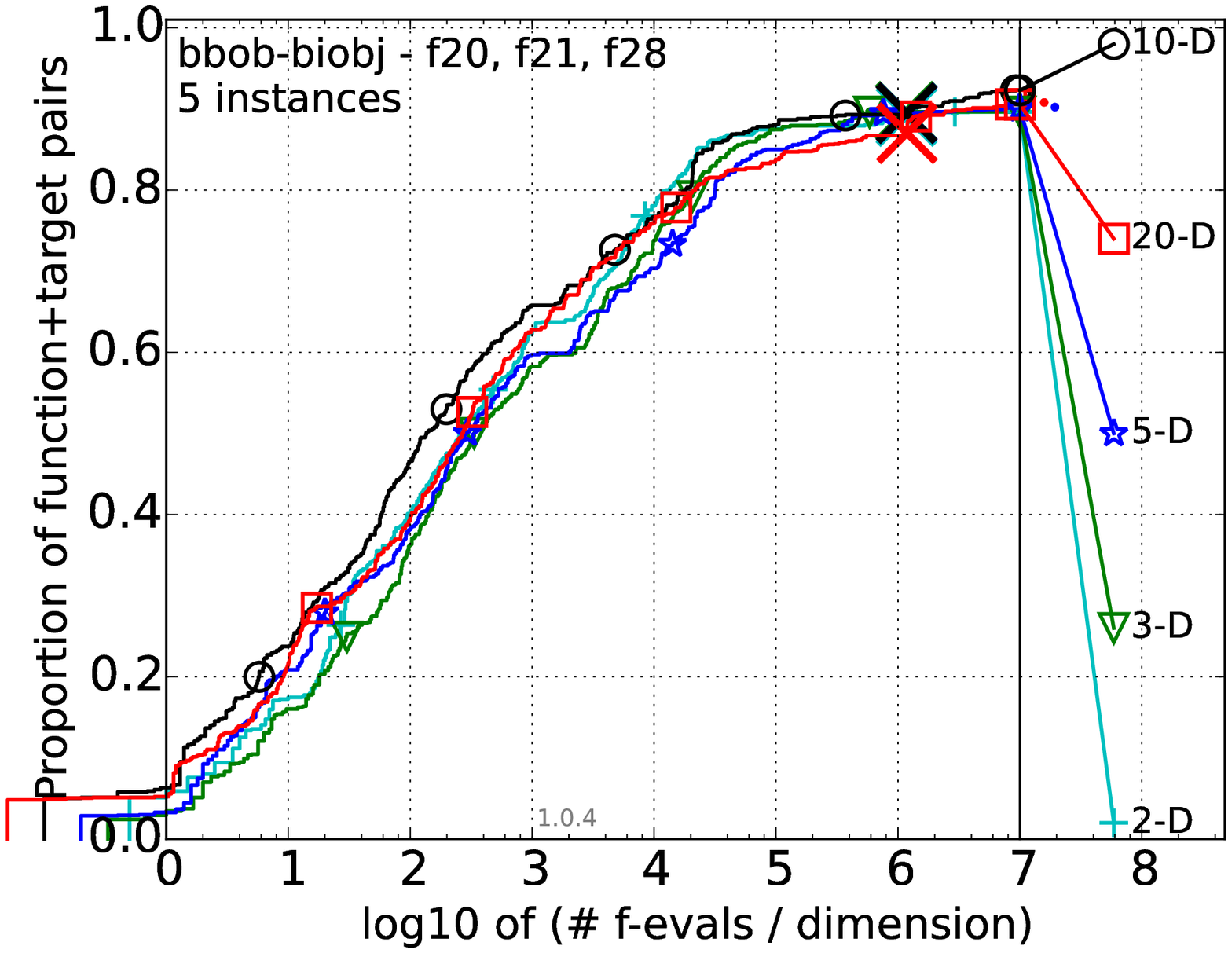} &
\includegraphics[width=0.268\textwidth,trim=0 0 0 13mm, clip]{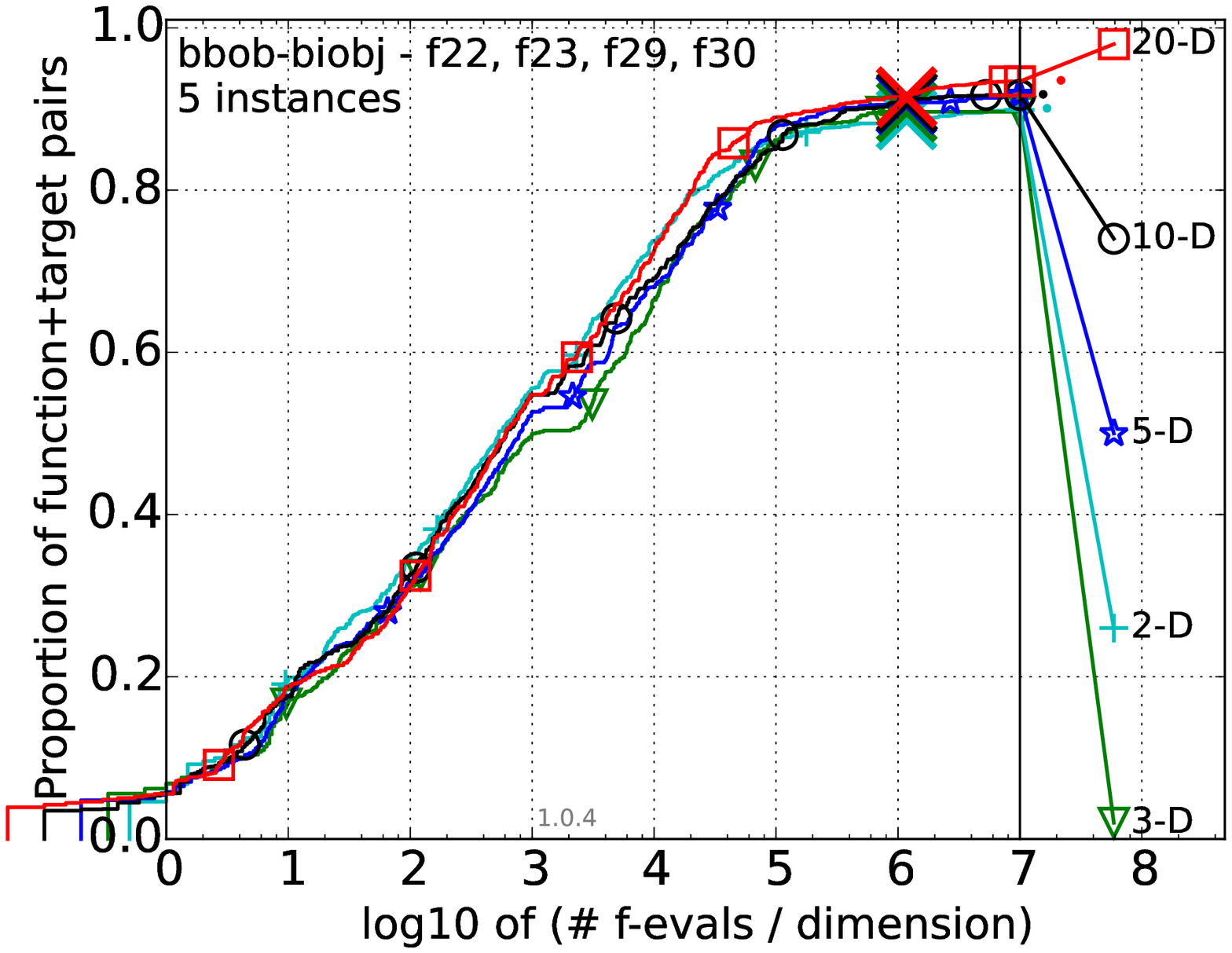} &
\includegraphics[width=0.268\textwidth,trim=0 0 0 13mm, clip]{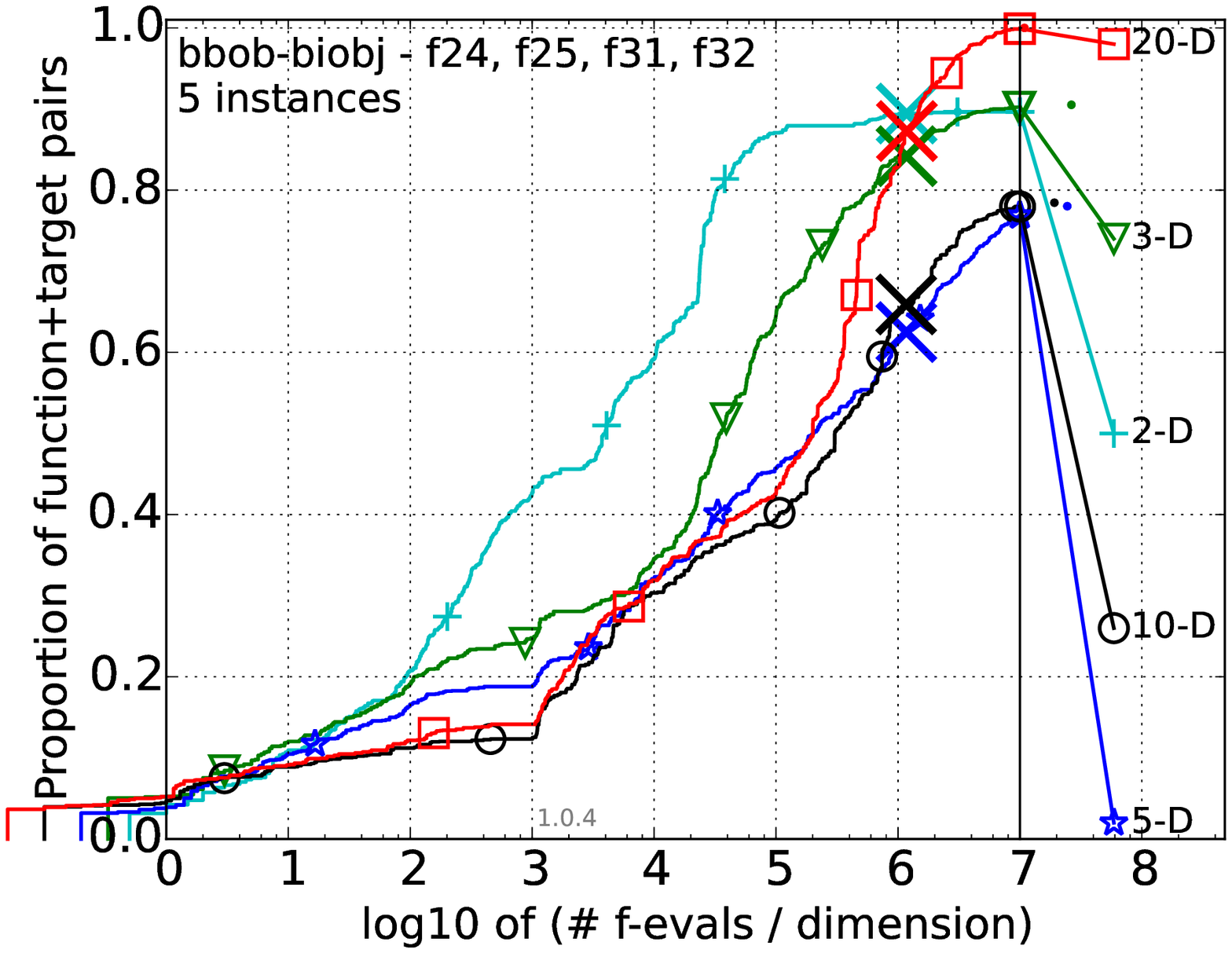}\\
moderate-weakstructure & ill-cond.-ill-cond. & ill-cond.-multimodal & ill-cond.-weakstructure\\
\includegraphics[width=0.268\textwidth,trim=0 0 0 13mm, clip]{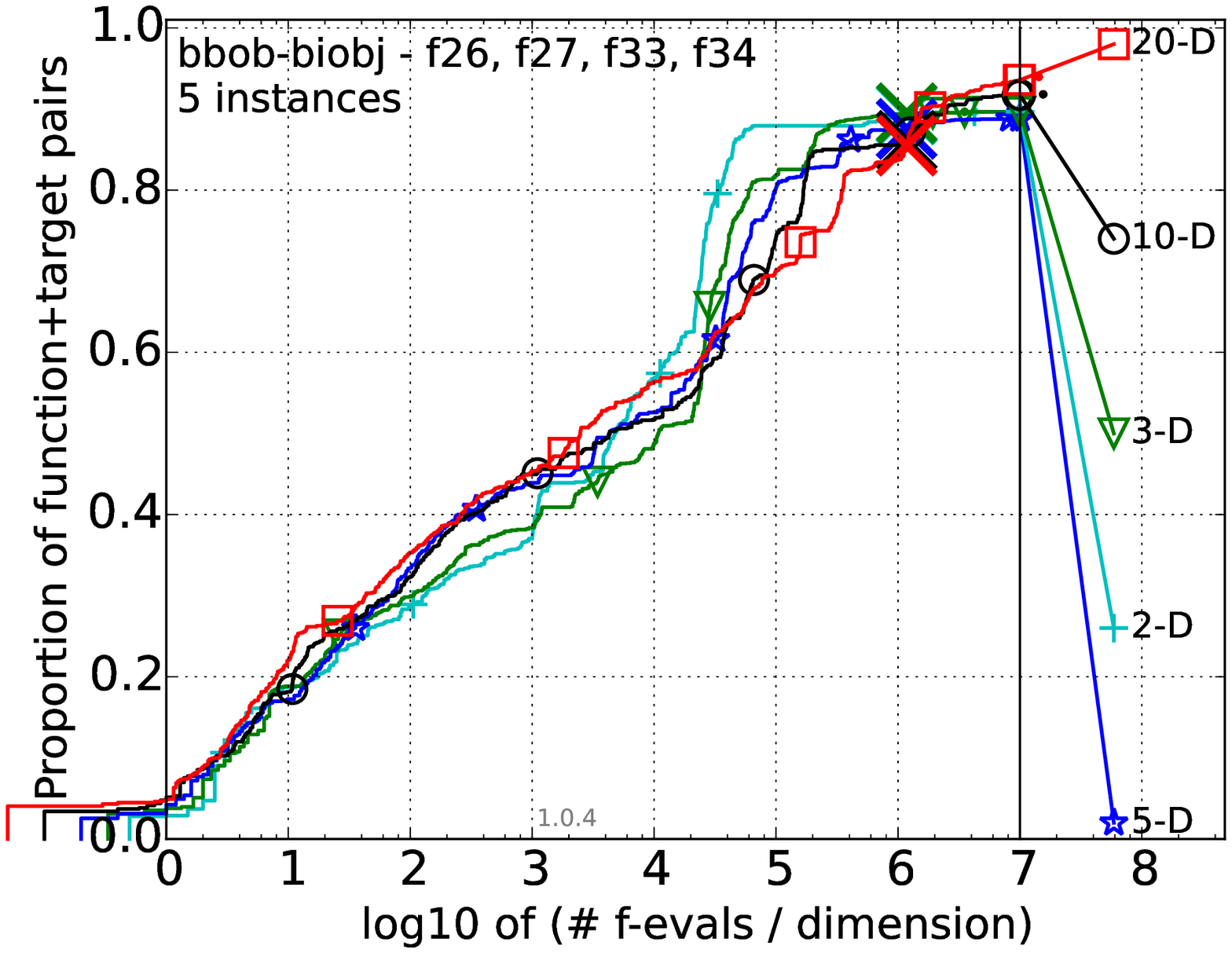} &
\includegraphics[width=0.268\textwidth,trim=0 0 0 13mm, clip]{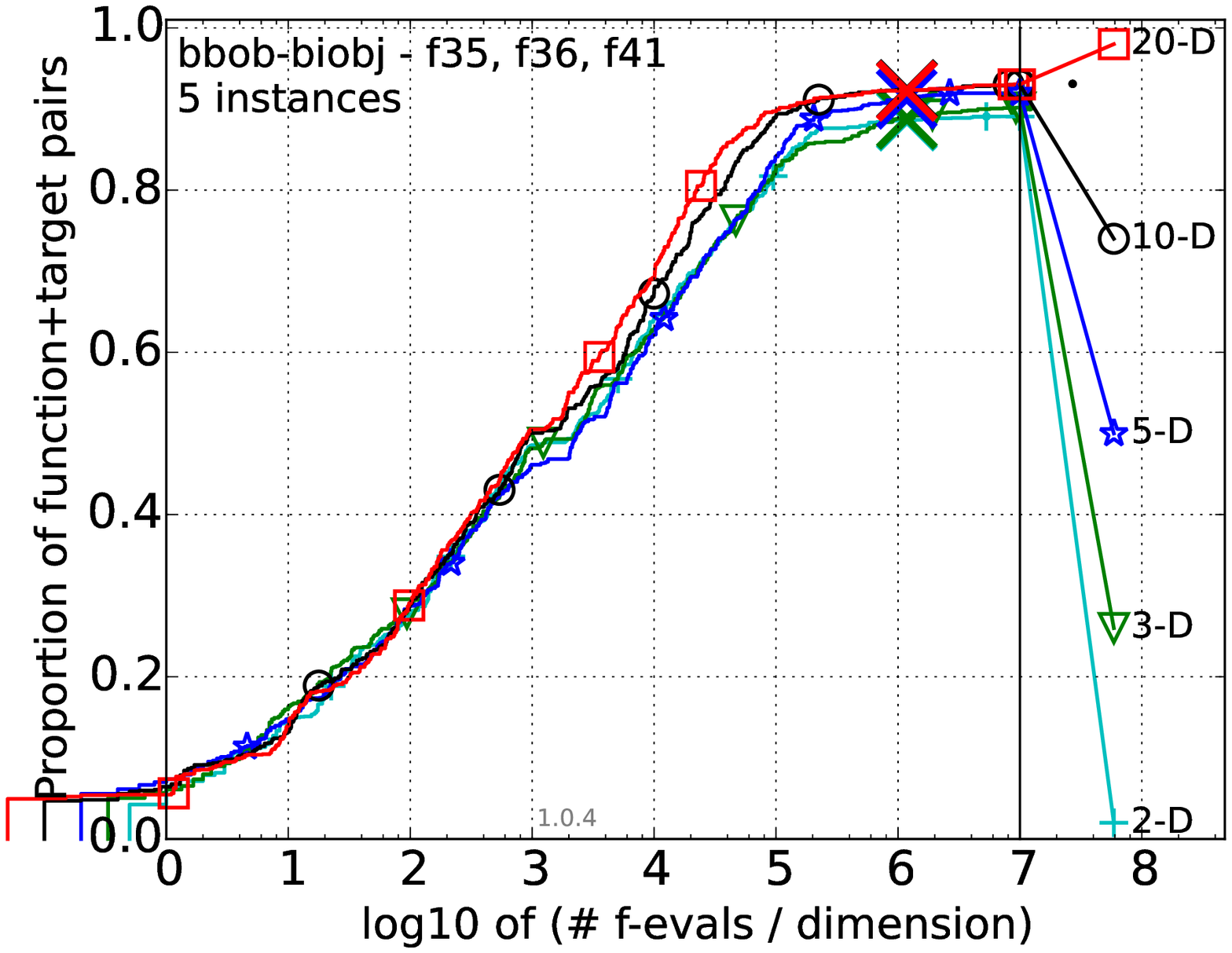} &
\includegraphics[width=0.268\textwidth,trim=0 0 0 13mm, clip]{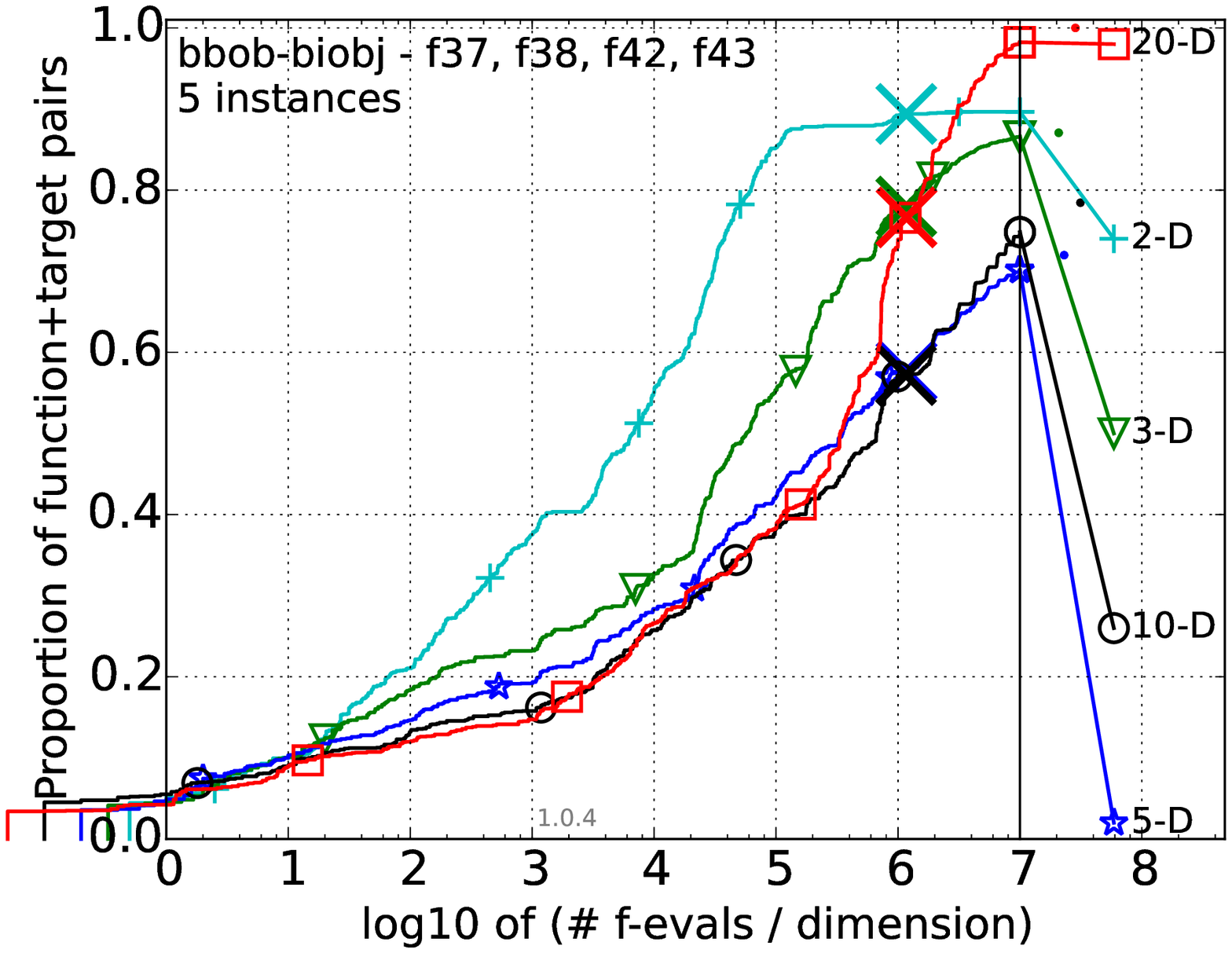} &
\includegraphics[width=0.268\textwidth,trim=0 0 0 13mm, clip]{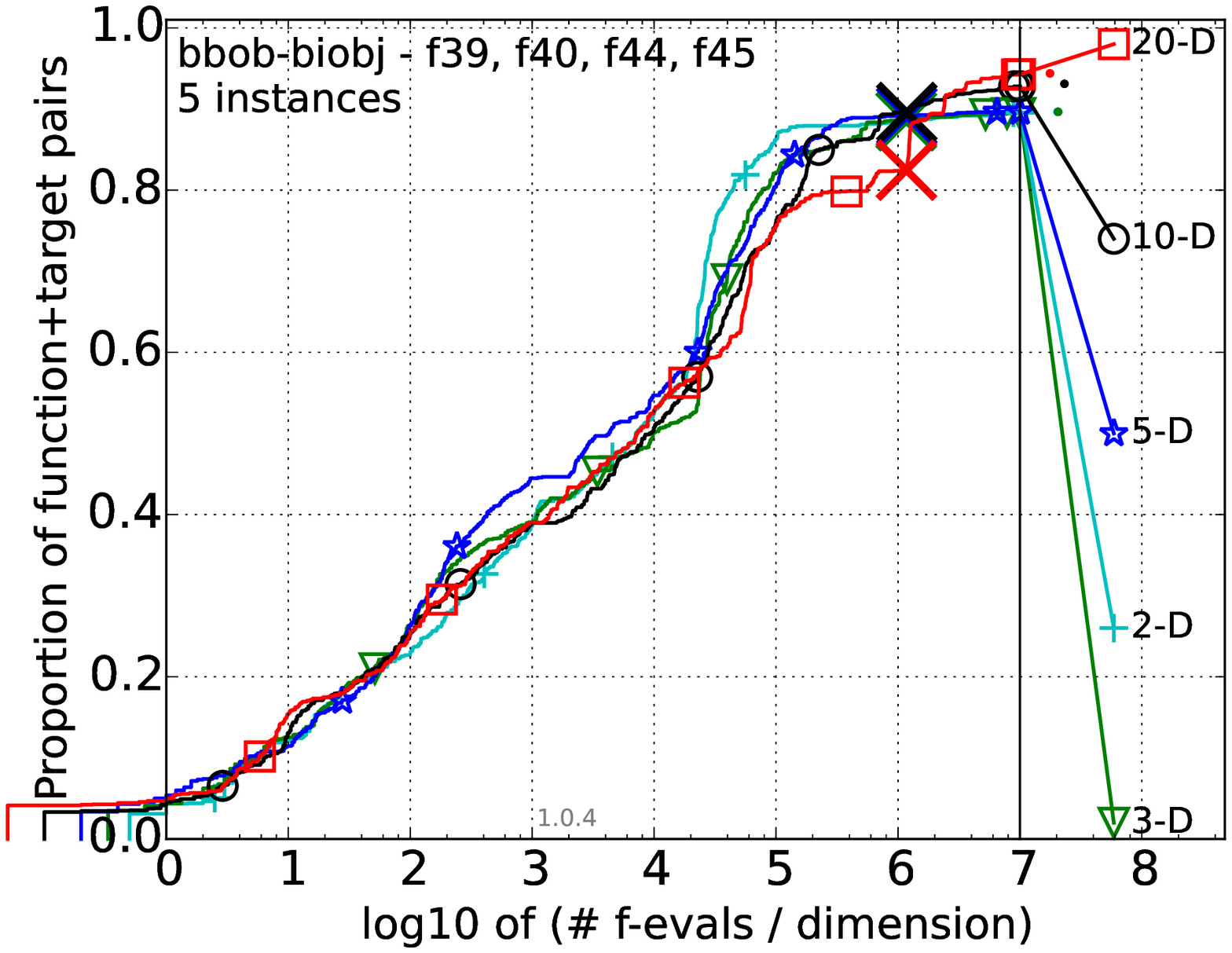} \\
multimodal-multimodal & multimodal-weakstructure & weakstructure-weakstructure & all 55 functions\\
\includegraphics[width=0.268\textwidth,trim=0 0 0 13mm, clip]{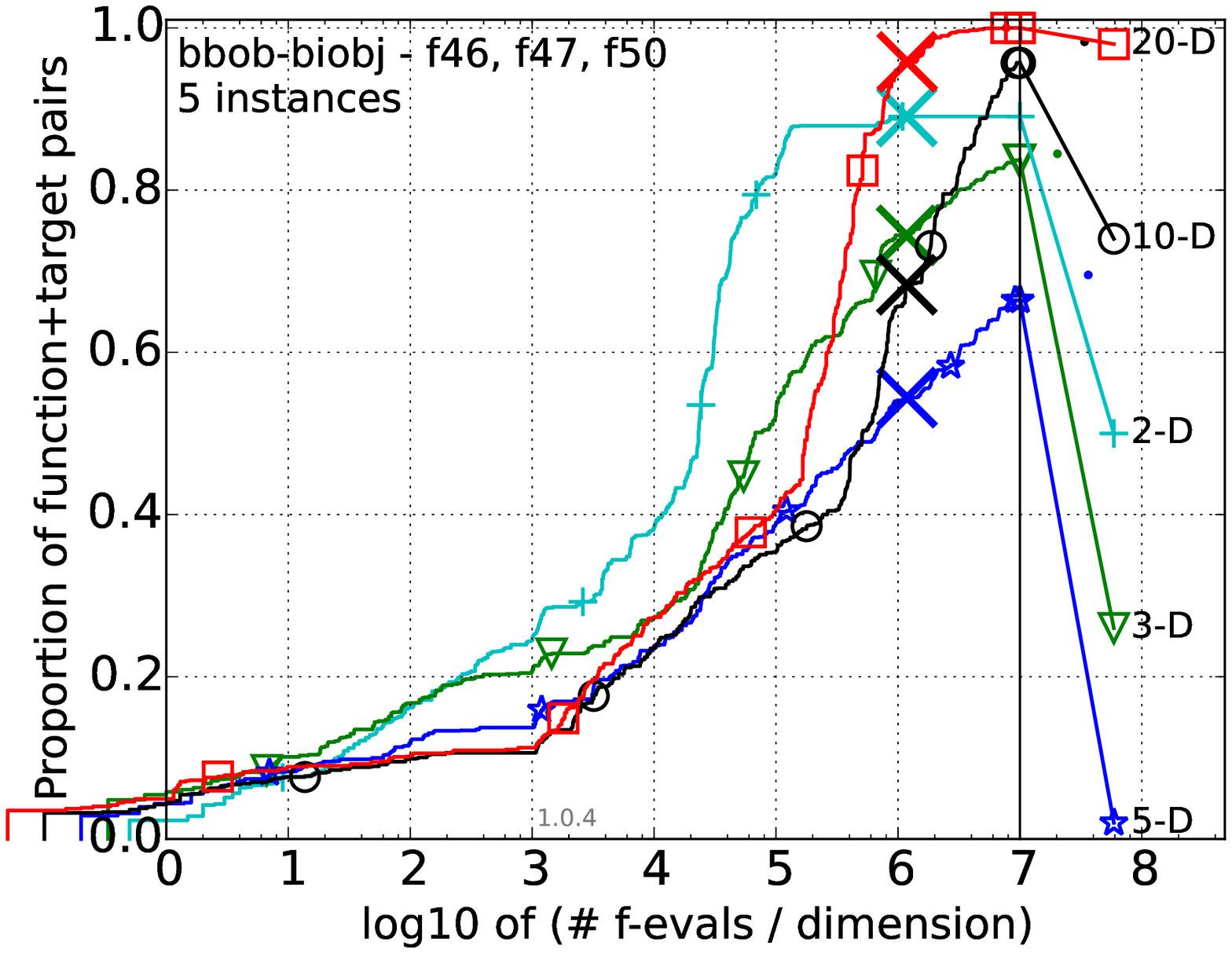} &
\includegraphics[width=0.268\textwidth,trim=0 0 0 13mm, clip]{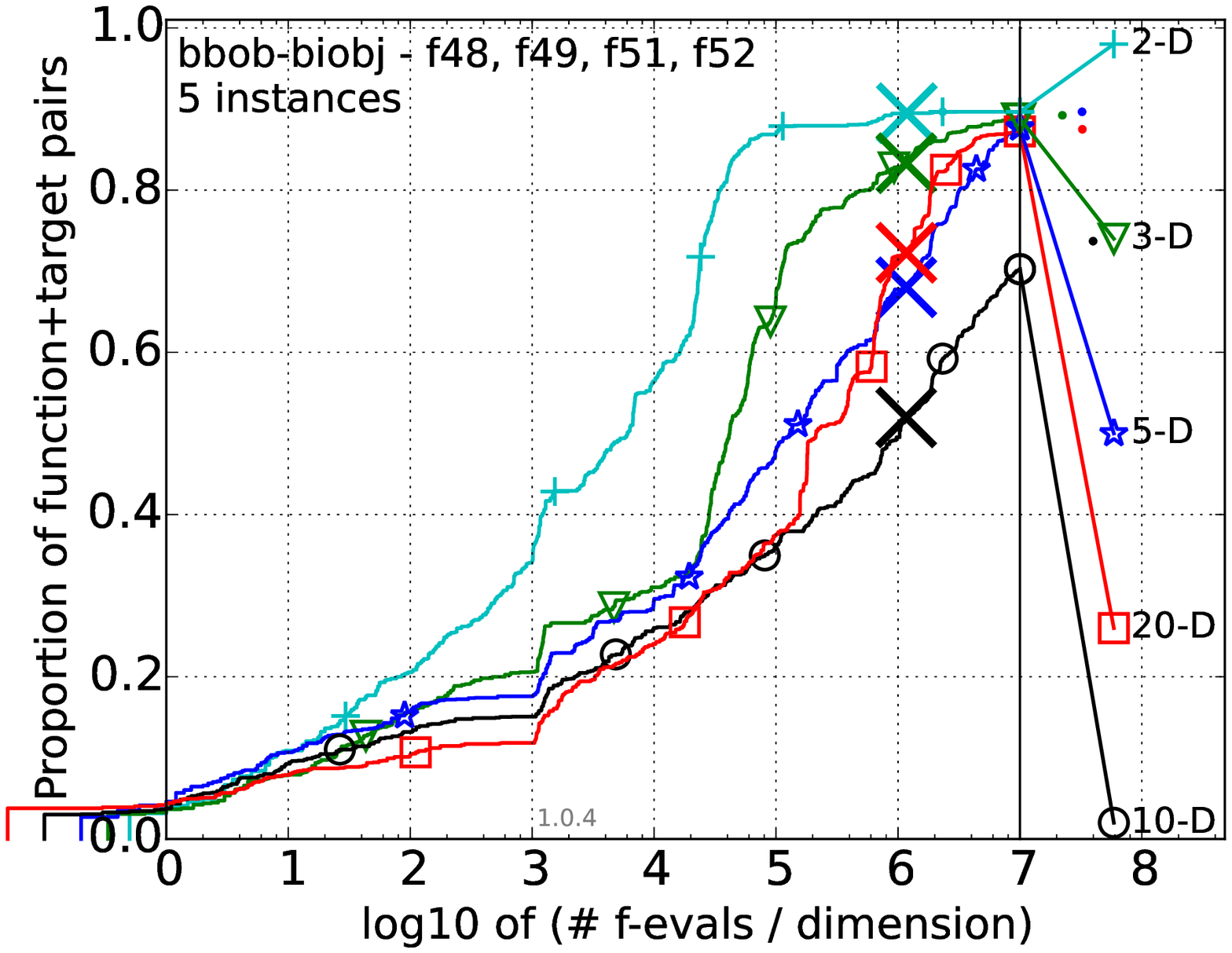} &
\includegraphics[width=0.268\textwidth,trim=0 0 0 13mm, clip]{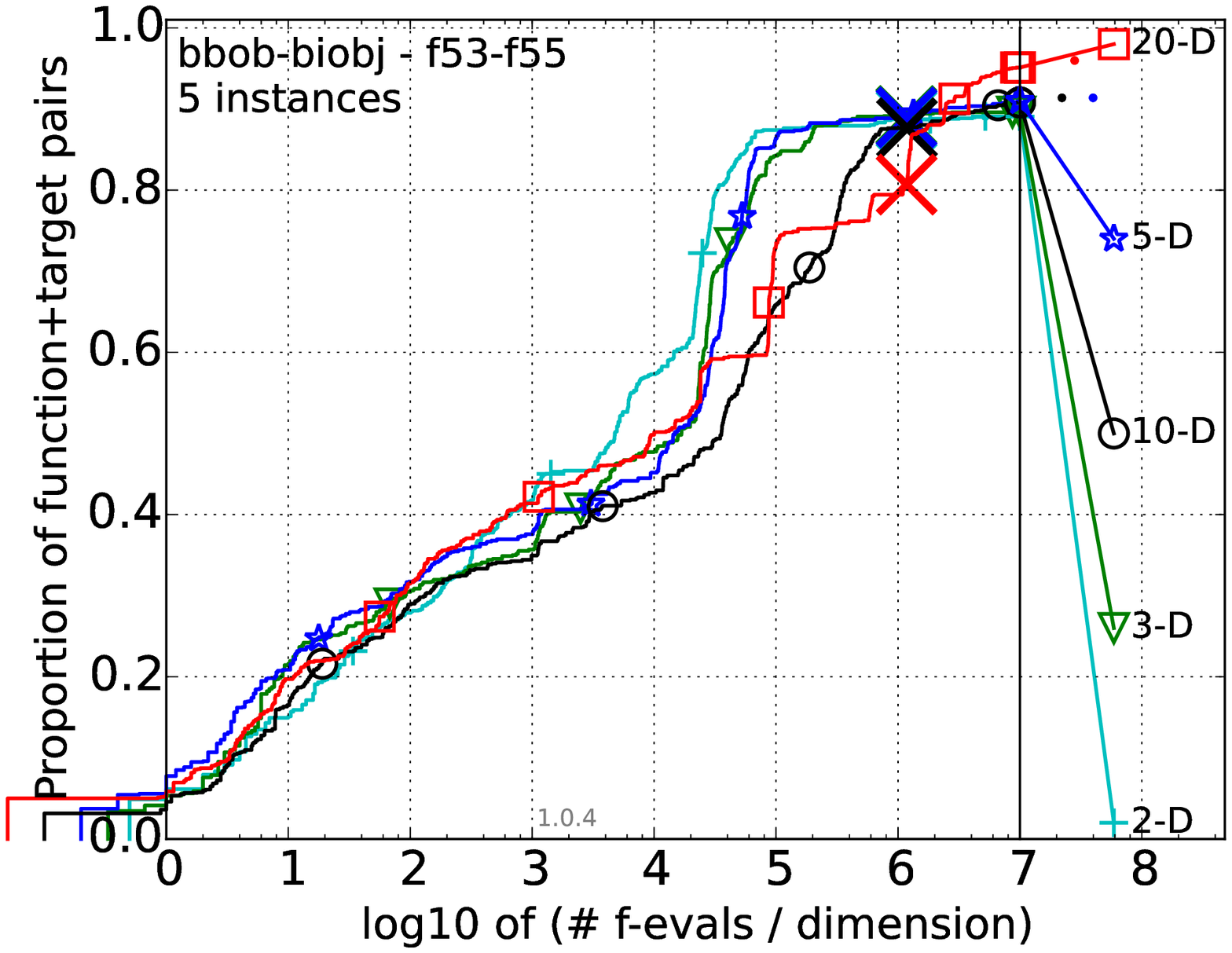} &
\includegraphics[width=0.268\textwidth,trim=0 0 0 13mm, clip]{pprldmany-single-functions/pprldmany}
\vspace*{-0.5ex}
\end{tabular}
 \caption{\label{fig:ECDFsGroups}
 \bbobecdfcaptionallgroups{}
 }
\end{figure*}

Results of HMO-CMA-ES from experiments according to \cite{biobj2016exp}
on the benchmark functions given in \cite{biobj2016func} are presented in
Figures~\ref{fig:all}, \ref{fig:ECDFsingleOne}, \ref{fig:ECDFsingleTwo},
\ref{fig:ECDFsingleThree}, and \ref{fig:ECDFsGroups}, and in
Table~\ref{tab:aRTs}.

For each problem instance, the performance of HMO-CMA-ES is assessed in
terms of the hypervolume metric~\cite{wagner2007pareto} (to be maximized),
the Lebesgue measure of the points that are a) dominated by at least one
objective vector found by the algorithm, and b) dominate a given
reference point. This hypervolume is assessed relative to a reference
value, which is the dominated hypervolume of a reference Pareto front
consisting of the best known set of objective vectors for this problem.
The task of maximizing the hypervolume is equivalent to minimizing the
difference between reference hypervolume and achieved hypervolume (to be
minimized). The reference hypervolume defines $58$ target values for this
difference, which are multiples of the reference hypervolume with the
factors
$\{-10^{-4}, -10^{-4.2}, -10^{-4.4}, -10^{-4.6}, -10^{-4.8},$\\
$-10^{-5}, 0, 10^{-5}, 10^{-4.9},\dots, 10^0\}$.
All results are reported in terms of the fraction of reached target
values. This normalization makes the results roughly comparable across
different problem types.

Hence, if an algorithm finds a non-dominated front of exactly the same
quality as the best known, then it reaches 52 out of 58 targets (including $0$).
This corresponds to roughly $0.9 \approx \frac{52}{58}$ on the vertical
axis of the plots used in this paper, i.e., a curve stopping at around
$0.9$ suggest that the best known approximation was reached.
In most cases, the best known approximation provided by the biobj-BBOB
2016 is very close to the true best value of the Pareto front and thus
$0.9$ is about the maximum possible value one can reach. This is often
the case on uni-modal functions. However, on some multi-modal functions
(i.e., when at least one of the objectives is multi-modal) the current
best approximation can be further improved and thus an algorithm can reach
targets with negative factors corresponding to better hypervolume values
than the reference of biobj-BBOB 2016. Indeed, the introduction of
negative targets was motivated by the fact that the current known
approximations are not the best possible ones in some cases.

Figure \ref{fig:all} shows the aggregated results over all 55 functions
for search space dimension $n = 2,3,5,10,20$. The saturation at a value
of $0.9$ can be well observed on 2-dimensional problems, where the best
known hypervolume values are indeed very close to the optimal ones. In
this case, HMO-CMA-ES solves most of the problems after about $10^5n$
function evaluations, from where on the curves stagnate.
The problems become harder with $n$, hence more function evaluations are
typically needed to reach a similar average performance.

Table~\ref{tab:aRTs} shows the average runtime to reach given targets on
5- and 20-dimensional problems. Figures~\ref{fig:ECDFsingleOne},
\ref{fig:ECDFsingleTwo}, \ref{fig:ECDFsingleThree} show the empirical
cumulative distribution of simulated (bootstrapped) runtimes
\cite{biobj2016exp} for all 55 functions and all considered problem
dimensions. Some functions can be associated with a much higher variance
in the results and with a faster than linear growth of the complexity
w.r.t.~$n$. This is often the case for multi-modal functions, but
sometimes also appears for ill-conditioned problems. On some (often
20-dimensional) problems the best known hypervolume value can be
improved, this happens when the curve crosses the value of about $0.9$.

\section{Conclusion}

We have presented a hybrid algorithm for multi-objective optimization,
combining of a number of well-performing components for single- and
multi-objective optimization.
We showed that the proposed algorithm can solve almost all biobj-BBOB
problems. When large computational budgets are considered it finds high
quality solution sets for nearly all combinations of problem type and
search space dimension. We attribute this robustness to the combination
of different optimizer components into a hybrid algorithm.
The performance relative to other multi-objective algorithms will be known
as soon as the results of the biobj-BBOB 2016 edition are available.

The algorithm has a set of hyperparameters, mostly start and stop times
(iteration numbers) encoded as multiples of $n$. Better tuning of these
would most probably improve its performance. It may be possible to
replace some of the hard numbers with adaptive stopping criteria.

The algorithm can be outperformed on some functions even by its own
individual components (the price of hybridization) or when a particular
computational budget is considered (the price for its good anytime
performance). It should be possible to considerably reduce these effects
with online prioritization of individual components depending on their
relative performance. However, this step is left for future work.

\section{Acknowledgements}	

We thank Frank Hutter and Martin Pil{\'a}t for valuable discissions.

%%%%%%%%%%%%%%%%%%%%%%%%%%%%%%%%%%%%%%%%%%%%%%%%%%%%%%%%%%%%%%%%%%%%%%%%%%%%%%%
%%%%%%%%%%%%%%%%%%%%%%%%%%%%%%%%%%%%%%%%%%%%%%%%%%%%%%%%%%%%%%%%%%%%%%%%%%%%%%%

% Scaling of ECDFs with dimension

%%%%%%%%%%%%%%%%%%%%%%%%%%%%%%%%%%%%%%%%%%%%%%%%%%%%%%%%%%%%%%%%%%%%%%%%%%%%%%%

%%%%%%%%%%%%%%%%%%%%%%%%%%%%%%%%%%%%%%%%%%%%%%%%%%%%%%%%%%%%%%%%%%%%%%%%%%%%%%%
%%%%%%%%%%%%%%%%%%%%%%%%%%%%%%%%%%%%%%%%%%%%%%%%%%%%%%%%%%%%%%%%%%%%%%%%%%%%%%%

% Empirical cumulative distribution functions (ECDFs) per function group.

%%%%%%%%%%%%%%%%%%%%%%%%%%%%%%%%%%%%%%%%%%%%%%%%%%%%%%%%%%%%%%%%%%%%%%%%%%%%%%%

\newcommand{\rot}[2][2.5]{
  \hspace*{-3.5\baselineskip}%
  \begin{rotate}{90}\hspace{#1em}#2
  \end{rotate}}

%%%%%%%%%%%%%%%%%%%%%%%%%%%%%%%%%%%%%%%%%%%%%%%%%%%%%%%%%%%%%%%%%%%%%%%%%%%%%%%
%%%%%%%%%%%%%%%%%%%%%%%%%%%%%%%%%%%%%%%%%%%%%%%%%%%%%%%%%%%%%%%%%%%%%%%%%%%%%%%
 
% Table showing the average running time (aRT in number of function
% evaluations) to reach the given targets for functions $f_1$--$f_{55}$.

%%%%%%%%%%%%%%%%%%%%%%%%%%%%%%%%%%%%%%%%%%%%%%%%%%%%%%%%%%%%%%%%%%%%%%%%%%%%%%%

\begin{sidewaystable*}
\centering {\tiny
\parbox{0.499\textwidth}{\centering
   {\small 5-D}\\
   \begin{tabular}{@{}c@{}|*{6}{@{}r@{}@{}l@{}}|@{}r@{}@{}l@{}}
$\Delta f$ & \multicolumn{2}{c}{1e+0} & \multicolumn{2}{c}{1e-1} & \multicolumn{2}{c}{1e-2} & \multicolumn{2}{c}{1e-3} & \multicolumn{2}{c}{1e-4} & \multicolumn{2}{c}{1e-5} & \multicolumn{2}{|@{}r@{}}{\#succ}\\\hline
${\bf f_{1}}$ & 1& & 76&${\scriptscriptstyle(34)}$ & 622&${\scriptscriptstyle(119)}$ & 3680&${\scriptscriptstyle(232)}$ & 38122&${\scriptscriptstyle(727)}$ & \multicolumn{2}{@{}c|@{}}{6.1e5${\scriptscriptstyle(9e5)}$} & 5 & /5\\\hline
${\bf f_{2}}$ & 3 & .0${\scriptscriptstyle(2)}$ & 137&${\scriptscriptstyle(202)}$ & 675&${\scriptscriptstyle(78)}$ & 3282&${\scriptscriptstyle(1175)}$ & 37044&${\scriptscriptstyle(10035)}$ & \multicolumn{2}{@{}c|@{}}{3.0e5${\scriptscriptstyle(71100)}$} & 5 & /5\\\hline
${\bf f_{3}}$ & 1& & 92&${\scriptscriptstyle(52)}$ & 706&${\scriptscriptstyle(294)}$ & 5848&${\scriptscriptstyle(4283)}$ & 44638&${\scriptscriptstyle(28997)}$ & \multicolumn{2}{@{}c|@{}}{3.0e5${\scriptscriptstyle(2e5)}$} & 5 & /5\\\hline
${\bf f_{4}}$ & 1& & 105&${\scriptscriptstyle(60)}$ & 573&${\scriptscriptstyle(282)}$ & 2722&${\scriptscriptstyle(1313)}$ & 30360&${\scriptscriptstyle(6471)}$ & \multicolumn{2}{@{}c|@{}}{3.0e5${\scriptscriptstyle(2e5)}$} & 5 & /5\\\hline
${\bf f_{5}}$ & 1& & 107&${\scriptscriptstyle(26)}$ & 1246&${\scriptscriptstyle(346)}$ & 26575&${\scriptscriptstyle(19553)}$ & \multicolumn{2}{@{}c@{}}{1.2e5${\scriptscriptstyle(57570)}$} & \multicolumn{2}{@{}c|@{}}{6.8e5${\scriptscriptstyle(2e5)}$} & 5 & /5\\\hline
${\bf f_{6}}$ & 1& & 55&${\scriptscriptstyle(26)}$ & 698&${\scriptscriptstyle(424)}$ & 3951&${\scriptscriptstyle(1002)}$ & 47723&${\scriptscriptstyle(12374)}$ & \multicolumn{2}{@{}c|@{}}{2.9e5${\scriptscriptstyle(60080)}$} & 5 & /5\\\hline
${\bf f_{7}}$ & 1& & 698&${\scriptscriptstyle(538)}$ & \multicolumn{2}{@{}c@{}}{1.1e5${\scriptscriptstyle(1e5)}$} & \multicolumn{2}{@{}c@{}}{4.9e6${\scriptscriptstyle(5e6)}$} & \multicolumn{2}{@{}c@{}}{$\infty$} & \multicolumn{2}{@{}c|@{}}{$\infty$\textit{5.8e6}} & 0 & /5\\\hline
${\bf f_{8}}$ & 2 & .8${\scriptscriptstyle(4)}$ & 2622&${\scriptscriptstyle(4874)}$ & \multicolumn{2}{@{}c@{}}{1.7e5${\scriptscriptstyle(1e5)}$} & \multicolumn{2}{@{}c@{}}{2.0e6${\scriptscriptstyle(2e6)}$} & \multicolumn{2}{@{}c@{}}{2.7e7${\scriptscriptstyle(4e7)}$} & \multicolumn{2}{@{}c|@{}}{$\infty$\textit{5.8e6}} & 0 & /5\\\hline
${\bf f_{9}}$ & 1& & 104&${\scriptscriptstyle(67)}$ & 525&${\scriptscriptstyle(114)}$ & 2004&${\scriptscriptstyle(377)}$ & 20926&${\scriptscriptstyle(2309)}$ & \multicolumn{2}{@{}c|@{}}{4.4e5${\scriptscriptstyle(2e5)}$} & 5 & /5\\\hline
${\bf f_{10}}$ & 1& & 390&${\scriptscriptstyle(344)}$ & 46136&${\scriptscriptstyle(82956)}$ & \multicolumn{2}{@{}c@{}}{1.2e5${\scriptscriptstyle(88081)}$} & \multicolumn{2}{@{}c@{}}{1.6e5${\scriptscriptstyle(78123)}$} & \multicolumn{2}{@{}c|@{}}{4.1e5${\scriptscriptstyle(33848)}$} & 5 & /5\\\hline
${\bf f_{11}}$ & 1& & 56&${\scriptscriptstyle(36)}$ & 519&${\scriptscriptstyle(598)}$ & 9095&${\scriptscriptstyle(14876)}$ & \multicolumn{2}{@{}c@{}}{1.3e5${\scriptscriptstyle(1e5)}$} & \multicolumn{2}{@{}c|@{}}{9.7e5${\scriptscriptstyle(7e5)}$} & 5 & /5\\\hline
${\bf f_{12}}$ & 1& & 53&${\scriptscriptstyle(62)}$ & 584&${\scriptscriptstyle(493)}$ & 6412&${\scriptscriptstyle(13086)}$ & 28731&${\scriptscriptstyle(51373)}$ & \multicolumn{2}{@{}c|@{}}{1.6e5${\scriptscriptstyle(2e5)}$} & 5 & /5\\\hline
${\bf f_{13}}$ & 2 & .0${\scriptscriptstyle(2)}$ & 26&${\scriptscriptstyle(17)}$ & 300&${\scriptscriptstyle(608)}$ & 3749&${\scriptscriptstyle(7841)}$ & 26314&${\scriptscriptstyle(44084)}$ & 86884&${\scriptscriptstyle(81208)}$ & 5 & /5\\\hline
${\bf f_{14}}$ & 1& & 283&${\scriptscriptstyle(244)}$ & 2094&${\scriptscriptstyle(454)}$ & 15271&${\scriptscriptstyle(4918)}$ & \multicolumn{2}{@{}c@{}}{1.3e5${\scriptscriptstyle(1e5)}$} & \multicolumn{2}{@{}c|@{}}{3.9e5${\scriptscriptstyle(1e5)}$} & 5 & /5\\\hline
${\bf f_{15}}$ & 4 & .8${\scriptscriptstyle(5)}$ & 101&${\scriptscriptstyle(82)}$ & 874&${\scriptscriptstyle(612)}$ & 6392&${\scriptscriptstyle(4252)}$ & 51584&${\scriptscriptstyle(36287)}$ & \multicolumn{2}{@{}c|@{}}{3.1e5${\scriptscriptstyle(1e5)}$} & 5 & /5\\\hline
${\bf f_{16}}$ & 1& & 3283&${\scriptscriptstyle(7912)}$ & 98937&${\scriptscriptstyle(1e5)}$ & \multicolumn{2}{@{}c@{}}{5.1e6${\scriptscriptstyle(5e6)}$} & \multicolumn{2}{@{}c@{}}{2.6e7${\scriptscriptstyle(2e7)}$} & \multicolumn{2}{@{}c|@{}}{2.6e7${\scriptscriptstyle(3e7)}$} & 1 & /5\\\hline
${\bf f_{17}}$ & 48&${\scriptscriptstyle(57)}$ & 8237&${\scriptscriptstyle(7098)}$ & \multicolumn{2}{@{}c@{}}{1.1e5${\scriptscriptstyle(18825)}$} & \multicolumn{2}{@{}c@{}}{3.0e6${\scriptscriptstyle(6e6)}$} & \multicolumn{2}{@{}c@{}}{2.5e7${\scriptscriptstyle(3e7)}$} & \multicolumn{2}{@{}c|@{}}{2.7e7${\scriptscriptstyle(6e7)}$} & 1 & /5\\\hline
${\bf f_{18}}$ & 1& & 62&${\scriptscriptstyle(12)}$ & 881&${\scriptscriptstyle(1333)}$ & 7694&${\scriptscriptstyle(9780)}$ & \multicolumn{2}{@{}c@{}}{1.2e5${\scriptscriptstyle(1e5)}$} & \multicolumn{2}{@{}c|@{}}{2.0e6${\scriptscriptstyle(3e5)}$} & 4 & /5\\\hline
${\bf f_{19}}$ & 4 & .8${\scriptscriptstyle(4)}$ & 1449&${\scriptscriptstyle(1467)}$ & 51687&${\scriptscriptstyle(50191)}$ & \multicolumn{2}{@{}c@{}}{1.5e5${\scriptscriptstyle(2e5)}$} & \multicolumn{2}{@{}c@{}}{2.7e5${\scriptscriptstyle(4e5)}$} & \multicolumn{2}{@{}c|@{}}{1.9e6${\scriptscriptstyle(8e6)}$} & 4 & /5\\\hline
${\bf f_{20}}$ & 2 & .0${\scriptscriptstyle(2)}$ & 43&${\scriptscriptstyle(48)}$ & 587&${\scriptscriptstyle(512)}$ & 5553&${\scriptscriptstyle(6016)}$ & 33785&${\scriptscriptstyle(32218)}$ & \multicolumn{2}{@{}c|@{}}{2.4e5${\scriptscriptstyle(1e5)}$} & 5 & /5\\\hline
${\bf f_{21}}$ & 1& & 149&${\scriptscriptstyle(198)}$ & 13829&${\scriptscriptstyle(33214)}$ & 34787&${\scriptscriptstyle(77474)}$ & 60168&${\scriptscriptstyle(70988)}$ & \multicolumn{2}{@{}c|@{}}{2.2e5${\scriptscriptstyle(35280)}$} & 5 & /5\\\hline
${\bf f_{22}}$ & 1& & 95&${\scriptscriptstyle(45)}$ & 1275&${\scriptscriptstyle(446)}$ & 9521&${\scriptscriptstyle(2799)}$ & 75476&${\scriptscriptstyle(23468)}$ & \multicolumn{2}{@{}c|@{}}{3.9e5${\scriptscriptstyle(1e5)}$} & 5 & /5\\\hline
${\bf f_{23}}$ & 1& & 61&${\scriptscriptstyle(43)}$ & 662&${\scriptscriptstyle(405)}$ & 3948&${\scriptscriptstyle(1617)}$ & 38675&${\scriptscriptstyle(5208)}$ & \multicolumn{2}{@{}c|@{}}{2.8e5${\scriptscriptstyle(31801)}$} & 5 & /5\\\hline
${\bf f_{24}}$ & 2 & .2 & 536&${\scriptscriptstyle(180)}$ & \multicolumn{2}{@{}c@{}}{1.3e5${\scriptscriptstyle(33890)}$} & \multicolumn{2}{@{}c@{}}{3.4e6${\scriptscriptstyle(5e5)}$} & \multicolumn{2}{@{}c@{}}{$\infty$} & \multicolumn{2}{@{}c|@{}}{$\infty$\textit{5.8e6}} & 0 & /5\\\hline
${\bf f_{25}}$ & 2 & .4${\scriptscriptstyle(4)}$ & 10730&${\scriptscriptstyle(10578)}$ & \multicolumn{2}{@{}c@{}}{1.3e5${\scriptscriptstyle(1e5)}$} & \multicolumn{2}{@{}c@{}}{2.2e6${\scriptscriptstyle(2e6)}$} & \multicolumn{2}{@{}c@{}}{1.2e7${\scriptscriptstyle(9e6)}$} & \multicolumn{2}{@{}c|@{}}{2.6e7${\scriptscriptstyle(3e7)}$} & 1 & /5\\\hline
${\bf f_{26}}$ & 2 & .4${\scriptscriptstyle(4)}$ & 102&${\scriptscriptstyle(124)}$ & 537&${\scriptscriptstyle(452)}$ & 1672&${\scriptscriptstyle(551)}$ & \multicolumn{2}{@{}c@{}}{3.8e5${\scriptscriptstyle(4e5)}$} & \multicolumn{2}{@{}c|@{}}{4.1e6${\scriptscriptstyle(6e6)}$} & 3 & /5\\\hline
${\bf f_{27}}$ & 1& & 2443&${\scriptscriptstyle(2837)}$ & 83976&${\scriptscriptstyle(59564)}$ & \multicolumn{2}{@{}c@{}}{1.9e5${\scriptscriptstyle(1e5)}$} & \multicolumn{2}{@{}c@{}}{3.2e5${\scriptscriptstyle(2e5)}$} & \multicolumn{2}{@{}c|@{}}{7.4e5${\scriptscriptstyle(6e5)}$} & 5 & /5\\\hline
${\bf f_{28}}$ & 1& & 24&${\scriptscriptstyle(10)}$ & 152&${\scriptscriptstyle(46)}$ & 1177&${\scriptscriptstyle(459)}$ & 90425&${\scriptscriptstyle(26397)}$ & \multicolumn{2}{@{}c|@{}}{1.5e6${\scriptscriptstyle(1e6)}$} & 5 & /5\\\hline
${\bf f_{29}}$ & 1& & 84&${\scriptscriptstyle(68)}$ & 1774&${\scriptscriptstyle(1312)}$ & 16753&${\scriptscriptstyle(12610)}$ & \multicolumn{2}{@{}c@{}}{1.3e5${\scriptscriptstyle(49966)}$} & \multicolumn{2}{@{}c|@{}}{4.3e5${\scriptscriptstyle(1e5)}$} & 5 & /5\\\hline
${\bf f_{30}}$ & 1& & 34&${\scriptscriptstyle(11)}$ & 476&${\scriptscriptstyle(322)}$ & 2851&${\scriptscriptstyle(896)}$ & 86136&${\scriptscriptstyle(2e5)}$ & \multicolumn{2}{@{}c|@{}}{4.0e5${\scriptscriptstyle(6e5)}$} & 5 & /5\\\hline
${\bf f_{31}}$ & 1& & 1371&${\scriptscriptstyle(2668)}$ & 67643&${\scriptscriptstyle(66756)}$ & \multicolumn{2}{@{}c@{}}{3.0e6${\scriptscriptstyle(3e6)}$} & \multicolumn{2}{@{}c@{}}{2.8e7${\scriptscriptstyle(4e7)}$} & \multicolumn{2}{@{}c|@{}}{$\infty$\textit{5.9e6}} & 0 & /5\\\hline
${\bf f_{32}}$ & 2 & .4${\scriptscriptstyle(4)}$ & 5032&${\scriptscriptstyle(7490)}$ & 93375&${\scriptscriptstyle(53306)}$ & \multicolumn{2}{@{}c@{}}{2.0e6${\scriptscriptstyle(2e6)}$} & \multicolumn{2}{@{}c@{}}{1.3e7${\scriptscriptstyle(8e6)}$} & \multicolumn{2}{@{}c|@{}}{2.9e7${\scriptscriptstyle(3e7)}$} & 1 & /5\\\hline
${\bf f_{33}}$ & 2 & .0 & 11&${\scriptscriptstyle(4)}$ & 70&${\scriptscriptstyle(53)}$ & 491&${\scriptscriptstyle(340)}$ & \multicolumn{2}{@{}c@{}}{3.6e5${\scriptscriptstyle(8e5)}$} & \multicolumn{2}{@{}c|@{}}{5.2e6${\scriptscriptstyle(1e7)}$} & 3 & /5\\\hline
${\bf f_{34}}$ & 1& & 222&${\scriptscriptstyle(184)}$ & 43144&${\scriptscriptstyle(54731)}$ & \multicolumn{2}{@{}c@{}}{2.5e5${\scriptscriptstyle(70626)}$} & \multicolumn{2}{@{}c@{}}{2.9e5${\scriptscriptstyle(1e5)}$} & \multicolumn{2}{@{}c|@{}}{8.2e5${\scriptscriptstyle(1e6)}$} & 5 & /5\\\hline
${\bf f_{35}}$ & 1& & 102&${\scriptscriptstyle(34)}$ & 2666&${\scriptscriptstyle(353)}$ & 55515&${\scriptscriptstyle(72979)}$ & \multicolumn{2}{@{}c@{}}{3.5e5${\scriptscriptstyle(3e5)}$} & \multicolumn{2}{@{}c|@{}}{1.5e6${\scriptscriptstyle(2e6)}$} & 5 & /5\\\hline
${\bf f_{36}}$ & 1& & 279&${\scriptscriptstyle(201)}$ & 7123&${\scriptscriptstyle(7264)}$ & 40761&${\scriptscriptstyle(58182)}$ & \multicolumn{2}{@{}c@{}}{1.3e5${\scriptscriptstyle(99680)}$} & \multicolumn{2}{@{}c|@{}}{4.8e5${\scriptscriptstyle(2e5)}$} & 5 & /5\\\hline
${\bf f_{37}}$ & 1& & 932&${\scriptscriptstyle(313)}$ & \multicolumn{2}{@{}c@{}}{1.7e5${\scriptscriptstyle(69116)}$} & \multicolumn{2}{@{}c@{}}{3.3e6${\scriptscriptstyle(5e6)}$} & \multicolumn{2}{@{}c@{}}{2.7e7${\scriptscriptstyle(2e7)}$} & \multicolumn{2}{@{}c|@{}}{$\infty$\textit{5.9e6}} & 0 & /5\\\hline
${\bf f_{38}}$ & 1& & 27778&${\scriptscriptstyle(50620)}$ & \multicolumn{2}{@{}c@{}}{3.1e5${\scriptscriptstyle(1e5)}$} & \multicolumn{2}{@{}c@{}}{1.2e7${\scriptscriptstyle(2e7)}$} & \multicolumn{2}{@{}c@{}}{$\infty$} & \multicolumn{2}{@{}c|@{}}{$\infty$\textit{5.9e6}} & 0 & /5\\\hline
${\bf f_{39}}$ & 1& & 213&${\scriptscriptstyle(152)}$ & 1043&${\scriptscriptstyle(398)}$ & 89093&${\scriptscriptstyle(1e5)}$ & \multicolumn{2}{@{}c@{}}{2.6e5${\scriptscriptstyle(2e5)}$} & \multicolumn{2}{@{}c|@{}}{7.0e5${\scriptscriptstyle(5e5)}$} & 5 & /5\\\hline
${\bf f_{40}}$ & 1& & 1505&${\scriptscriptstyle(1634)}$ & 18105&${\scriptscriptstyle(20769)}$ & \multicolumn{2}{@{}c@{}}{1.8e5${\scriptscriptstyle(47613)}$} & \multicolumn{2}{@{}c@{}}{3.7e5${\scriptscriptstyle(2e5)}$} & \multicolumn{2}{@{}c|@{}}{1.6e6${\scriptscriptstyle(4e5)}$} & 5 & /5\\\hline
${\bf f_{41}}$ & 1& & 38&${\scriptscriptstyle(38)}$ & 633&${\scriptscriptstyle(277)}$ & 5997&${\scriptscriptstyle(2780)}$ & 42241&${\scriptscriptstyle(18982)}$ & \multicolumn{2}{@{}c|@{}}{2.6e5${\scriptscriptstyle(1e5)}$} & 5 & /5\\\hline
${\bf f_{42}}$ & 1& & 648&${\scriptscriptstyle(629)}$ & \multicolumn{2}{@{}c@{}}{2.4e5${\scriptscriptstyle(2e5)}$} & \multicolumn{2}{@{}c@{}}{4.8e6${\scriptscriptstyle(5e6)}$} & \multicolumn{2}{@{}c@{}}{$\infty$} & \multicolumn{2}{@{}c|@{}}{$\infty$\textit{5.9e6}} & 0 & /5\\\hline
${\bf f_{43}}$ & 1 & .8${\scriptscriptstyle(2)}$ & 25272&${\scriptscriptstyle(52432)}$ & \multicolumn{2}{@{}c@{}}{2.0e5${\scriptscriptstyle(2e5)}$} & \multicolumn{2}{@{}c@{}}{4.9e6${\scriptscriptstyle(1e7)}$} & \multicolumn{2}{@{}c@{}}{2.6e7${\scriptscriptstyle(5e7)}$} & \multicolumn{2}{@{}c|@{}}{2.7e7${\scriptscriptstyle(2e7)}$} & 1 & /5\\\hline
${\bf f_{44}}$ & 1& & 51&${\scriptscriptstyle(34)}$ & 540&${\scriptscriptstyle(181)}$ & 1733&${\scriptscriptstyle(597)}$ & 36645&${\scriptscriptstyle(48512)}$ & \multicolumn{2}{@{}c|@{}}{9.6e5${\scriptscriptstyle(5e5)}$} & 5 & /5\\\hline
${\bf f_{45}}$ & 1& & 256&${\scriptscriptstyle(121)}$ & 56060&${\scriptscriptstyle(58192)}$ & \multicolumn{2}{@{}c@{}}{1.3e5${\scriptscriptstyle(98941)}$} & \multicolumn{2}{@{}c@{}}{1.8e5${\scriptscriptstyle(1e5)}$} & \multicolumn{2}{@{}c|@{}}{4.3e5${\scriptscriptstyle(1e5)}$} & 5 & /5\\\hline
${\bf f_{46}}$ & 1& & 14236&${\scriptscriptstyle(5438)}$ & \multicolumn{2}{@{}c@{}}{4.7e5${\scriptscriptstyle(2e5)}$} & \multicolumn{2}{@{}c@{}}{$\infty$} & \multicolumn{2}{@{}c@{}}{$\infty$} & \multicolumn{2}{@{}c|@{}}{$\infty$\textit{6.0e6}} & 0 & /5\\\hline
${\bf f_{47}}$ & 1& & 14416&${\scriptscriptstyle(9934)}$ & \multicolumn{2}{@{}c@{}}{2.7e5${\scriptscriptstyle(2e5)}$} & \multicolumn{2}{@{}c@{}}{6.4e6${\scriptscriptstyle(8e6)}$} & \multicolumn{2}{@{}c@{}}{2.7e7${\scriptscriptstyle(4e7)}$} & \multicolumn{2}{@{}c|@{}}{2.8e7${\scriptscriptstyle(3e7)}$} & 1 & /5\\\hline
${\bf f_{48}}$ & 1& & 11373&${\scriptscriptstyle(24814)}$ & 95569&${\scriptscriptstyle(90196)}$ & \multicolumn{2}{@{}c@{}}{1.0e6${\scriptscriptstyle(9e5)}$} & \multicolumn{2}{@{}c@{}}{9.7e6${\scriptscriptstyle(2e7)}$} & \multicolumn{2}{@{}c|@{}}{1.1e7${\scriptscriptstyle(1e7)}$} & 2 & /5\\\hline
${\bf f_{49}}$ & 1& & 2650&${\scriptscriptstyle(3074)}$ & \multicolumn{2}{@{}c@{}}{1.8e5${\scriptscriptstyle(84915)}$} & \multicolumn{2}{@{}c@{}}{5.9e5${\scriptscriptstyle(3e5)}$} & \multicolumn{2}{@{}c@{}}{1.2e7${\scriptscriptstyle(1e7)}$} & \multicolumn{2}{@{}c|@{}}{2.7e7${\scriptscriptstyle(3e7)}$} & 1 & /5\\\hline
${\bf f_{50}}$ & 1& & 11162&${\scriptscriptstyle(13848)}$ & \multicolumn{2}{@{}c@{}}{1.6e5${\scriptscriptstyle(30361)}$} & \multicolumn{2}{@{}c@{}}{2.9e6${\scriptscriptstyle(3e6)}$} & \multicolumn{2}{@{}c@{}}{$\infty$} & \multicolumn{2}{@{}c|@{}}{$\infty$\textit{5.9e6}} & 0 & /5\\\hline
${\bf f_{51}}$ & 1& & 7873&${\scriptscriptstyle(13724)}$ & \multicolumn{2}{@{}c@{}}{1.6e5${\scriptscriptstyle(1e5)}$} & \multicolumn{2}{@{}c@{}}{4.8e6${\scriptscriptstyle(3e6)}$} & \multicolumn{2}{@{}c@{}}{2.7e7${\scriptscriptstyle(2e7)}$} & \multicolumn{2}{@{}c|@{}}{2.7e7${\scriptscriptstyle(1e7)}$} & 1 & /5\\\hline
${\bf f_{52}}$ & 1& & 10582&${\scriptscriptstyle(7518)}$ & \multicolumn{2}{@{}c@{}}{1.5e5${\scriptscriptstyle(94938)}$} & \multicolumn{2}{@{}c@{}}{3.6e6${\scriptscriptstyle(3e6)}$} & \multicolumn{2}{@{}c@{}}{1.3e7${\scriptscriptstyle(8e6)}$} & \multicolumn{2}{@{}c|@{}}{$\infty$\textit{5.8e6}} & 0 & /5\\\hline
${\bf f_{53}}$ & 2 & .4${\scriptscriptstyle(2)}$ & 16&${\scriptscriptstyle(11)}$ & 50&${\scriptscriptstyle(19)}$ & 38574&${\scriptscriptstyle(96032)}$ & 88948&${\scriptscriptstyle(1e5)}$ & \multicolumn{2}{@{}c|@{}}{2.3e6${\scriptscriptstyle(2e6)}$} & 4 & /5\\\hline
${\bf f_{54}}$ & 1& & 89&${\scriptscriptstyle(31)}$ & 44791&${\scriptscriptstyle(84865)}$ & \multicolumn{2}{@{}c@{}}{1.3e5${\scriptscriptstyle(1e5)}$} & \multicolumn{2}{@{}c@{}}{1.5e5${\scriptscriptstyle(1e5)}$} & \multicolumn{2}{@{}c|@{}}{2.6e5${\scriptscriptstyle(71106)}$} & 5 & /5\\\hline
${\bf f_{55}}$ & 1 & .2${\scriptscriptstyle(0.2)}$ & 16706&${\scriptscriptstyle(21985)}$ & \multicolumn{2}{@{}c@{}}{1.3e5${\scriptscriptstyle(95226)}$} & \multicolumn{2}{@{}c@{}}{1.7e5${\scriptscriptstyle(45766)}$} & \multicolumn{2}{@{}c@{}}{2.4e5${\scriptscriptstyle(2e5)}$} & \multicolumn{2}{@{}c|@{}}{3.1e5${\scriptscriptstyle(1e5)}$} & 5 & /5
\end{tabular}}%
\parbox{0.499\textwidth}{\centering
   {\small 20-D}\\
   \begin{tabular}{@{}c@{}|*{6}{@{}r@{}@{}l@{}}|@{}r@{}@{}l@{}}
$\Delta f$ & \multicolumn{2}{c}{1e+0} & \multicolumn{2}{c}{1e-1} & \multicolumn{2}{c}{1e-2} & \multicolumn{2}{c}{1e-3} & \multicolumn{2}{c}{1e-4} & \multicolumn{2}{c}{1e-5} & \multicolumn{2}{|@{}r@{}}{\#succ}\\\hline
${\bf f_{1}}$ & 1& & 173&${\scriptscriptstyle(48)}$ & 1414&${\scriptscriptstyle(188)}$ & 8267&${\scriptscriptstyle(598)}$ & 76089&${\scriptscriptstyle(3218)}$ & \multicolumn{2}{@{}c|@{}}{3.9e5${\scriptscriptstyle(21373)}$} & 5 & /5\\\hline
${\bf f_{2}}$ & 1& & 176&${\scriptscriptstyle(74)}$ & 2284&${\scriptscriptstyle(1026)}$ & 23672&${\scriptscriptstyle(7342)}$ & \multicolumn{2}{@{}c@{}}{2.7e5${\scriptscriptstyle(1e5)}$} & \multicolumn{2}{@{}c|@{}}{3.1e6${\scriptscriptstyle(2e6)}$} & 5 & /5\\\hline
${\bf f_{3}}$ & 1& & 271&${\scriptscriptstyle(195)}$ & 2628&${\scriptscriptstyle(702)}$ & 11178&${\scriptscriptstyle(2747)}$ & 88645&${\scriptscriptstyle(25527)}$ & \multicolumn{2}{@{}c|@{}}{4.6e5${\scriptscriptstyle(42719)}$} & 5 & /5\\\hline
${\bf f_{4}}$ & 1& & 175&${\scriptscriptstyle(75)}$ & 1537&${\scriptscriptstyle(312)}$ & 9688&${\scriptscriptstyle(1302)}$ & 92783&${\scriptscriptstyle(5814)}$ & \multicolumn{2}{@{}c|@{}}{9.8e5${\scriptscriptstyle(6e5)}$} & 5 & /5\\\hline
${\bf f_{5}}$ & 1& & 229&${\scriptscriptstyle(36)}$ & 2991&${\scriptscriptstyle(297)}$ & 30127&${\scriptscriptstyle(14637)}$ & \multicolumn{2}{@{}c@{}}{1.9e5${\scriptscriptstyle(40187)}$} & \multicolumn{2}{@{}c|@{}}{8.0e5${\scriptscriptstyle(4e5)}$} & 5 & /5\\\hline
${\bf f_{6}}$ & 1& & 288&${\scriptscriptstyle(125)}$ & 2139&${\scriptscriptstyle(458)}$ & 13912&${\scriptscriptstyle(1273)}$ & \multicolumn{2}{@{}c@{}}{1.2e5${\scriptscriptstyle(15424)}$} & \multicolumn{2}{@{}c|@{}}{5.6e5${\scriptscriptstyle(60140)}$} & 5 & /5\\\hline
${\bf f_{7}}$ & 1& & 15922&${\scriptscriptstyle(35042)}$ & \multicolumn{2}{@{}c@{}}{4.7e5${\scriptscriptstyle(5e5)}$} & \multicolumn{2}{@{}c@{}}{1.5e7${\scriptscriptstyle(1e7)}$} & \multicolumn{2}{@{}c@{}}{3.3e7${\scriptscriptstyle(8e7)}$} & \multicolumn{2}{@{}c|@{}}{1.1e8${\scriptscriptstyle(1e8)}$} & 1 & /5\\\hline
${\bf f_{8}}$ & 1& & 40384&${\scriptscriptstyle(22527)}$ & \multicolumn{2}{@{}c@{}}{1.4e6${\scriptscriptstyle(2e6)}$} & \multicolumn{2}{@{}c@{}}{1.4e7${\scriptscriptstyle(2e7)}$} & \multicolumn{2}{@{}c@{}}{1.9e7${\scriptscriptstyle(7e6)}$} & \multicolumn{2}{@{}c|@{}}{1.9e7${\scriptscriptstyle(2e6)}$} & 4 & /5\\\hline
${\bf f_{9}}$ & 1& & 163&${\scriptscriptstyle(70)}$ & 1720&${\scriptscriptstyle(586)}$ & 6034&${\scriptscriptstyle(906)}$ & 47190&${\scriptscriptstyle(1876)}$ & \multicolumn{2}{@{}c|@{}}{5.7e5${\scriptscriptstyle(2e5)}$} & 5 & /5\\\hline
${\bf f_{10}}$ & 1& & 634&${\scriptscriptstyle(540)}$ & \multicolumn{2}{@{}c@{}}{2.3e5${\scriptscriptstyle(9849)}$} & \multicolumn{2}{@{}c@{}}{7.3e5${\scriptscriptstyle(5e5)}$} & \multicolumn{2}{@{}c@{}}{9.3e5${\scriptscriptstyle(8e5)}$} & \multicolumn{2}{@{}c|@{}}{1.3e6${\scriptscriptstyle(7e5)}$} & 5 & /5\\\hline
${\bf f_{11}}$ & 1& & 258&${\scriptscriptstyle(244)}$ & 3511&${\scriptscriptstyle(3284)}$ & \multicolumn{2}{@{}c@{}}{1.1e5${\scriptscriptstyle(1e5)}$} & \multicolumn{2}{@{}c@{}}{3.2e6${\scriptscriptstyle(2e6)}$} & \multicolumn{2}{@{}c|@{}}{1.8e7${\scriptscriptstyle(1e7)}$} & 4 & /5\\\hline
${\bf f_{12}}$ & 1& & 139&${\scriptscriptstyle(69)}$ & 1524&${\scriptscriptstyle(1066)}$ & 11420&${\scriptscriptstyle(10100)}$ & \multicolumn{2}{@{}c@{}}{1.4e5${\scriptscriptstyle(1e5)}$} & \multicolumn{2}{@{}c|@{}}{1.9e6${\scriptscriptstyle(2e6)}$} & 5 & /5\\\hline
${\bf f_{13}}$ & 1& & 175&${\scriptscriptstyle(72)}$ & 1419&${\scriptscriptstyle(399)}$ & 23144&${\scriptscriptstyle(15251)}$ & \multicolumn{2}{@{}c@{}}{9.8e6${\scriptscriptstyle(2e7)}$} & \multicolumn{2}{@{}c|@{}}{4.6e7${\scriptscriptstyle(4e7)}$} & 2 & /5\\\hline
${\bf f_{14}}$ & 1& & 229&${\scriptscriptstyle(114)}$ & 6010&${\scriptscriptstyle(1470)}$ & 53814&${\scriptscriptstyle(13114)}$ & \multicolumn{2}{@{}c@{}}{4.7e5${\scriptscriptstyle(3e5)}$} & \multicolumn{2}{@{}c|@{}}{2.2e6${\scriptscriptstyle(1e6)}$} & 5 & /5\\\hline
${\bf f_{15}}$ & 1& & 173&${\scriptscriptstyle(56)}$ & 1775&${\scriptscriptstyle(567)}$ & 13552&${\scriptscriptstyle(2336)}$ & \multicolumn{2}{@{}c@{}}{2.3e5${\scriptscriptstyle(63258)}$} & \multicolumn{2}{@{}c|@{}}{1.0e7${\scriptscriptstyle(2e7)}$} & 4 & /5\\\hline
${\bf f_{16}}$ & 1& & 31269&${\scriptscriptstyle(6709)}$ & \multicolumn{2}{@{}c@{}}{5.4e5${\scriptscriptstyle(5e5)}$} & \multicolumn{2}{@{}c@{}}{5.1e6${\scriptscriptstyle(4e6)}$} & \multicolumn{2}{@{}c@{}}{1.4e7${\scriptscriptstyle(2e7)}$} & \multicolumn{2}{@{}c|@{}}{1.4e7${\scriptscriptstyle(1e7)}$} & 4 & /5\\\hline
${\bf f_{17}}$ & 1& & 48714&${\scriptscriptstyle(39244)}$ & \multicolumn{2}{@{}c@{}}{6.0e5${\scriptscriptstyle(6e5)}$} & \multicolumn{2}{@{}c@{}}{4.9e6${\scriptscriptstyle(6e6)}$} & \multicolumn{2}{@{}c@{}}{1.5e7${\scriptscriptstyle(1e7)}$} & \multicolumn{2}{@{}c|@{}}{1.6e7${\scriptscriptstyle(1e7)}$} & 4 & /5\\\hline
${\bf f_{18}}$ & 1& & 164&${\scriptscriptstyle(26)}$ & 1567&${\scriptscriptstyle(528)}$ & 17826&${\scriptscriptstyle(14184)}$ & \multicolumn{2}{@{}c@{}}{5.3e5${\scriptscriptstyle(5e5)}$} & \multicolumn{2}{@{}c|@{}}{4.4e6${\scriptscriptstyle(2e6)}$} & 5 & /5\\\hline
${\bf f_{19}}$ & 1& & 6392&${\scriptscriptstyle(13525)}$ & \multicolumn{2}{@{}c@{}}{2.5e6${\scriptscriptstyle(3e6)}$} & \multicolumn{2}{@{}c@{}}{7.4e6${\scriptscriptstyle(8e5)}$} & \multicolumn{2}{@{}c@{}}{8.2e6${\scriptscriptstyle(2e7)}$} & \multicolumn{2}{@{}c|@{}}{1.1e7${\scriptscriptstyle(2e7)}$} & 4 & /5\\\hline
${\bf f_{20}}$ & 1& & 145&${\scriptscriptstyle(98)}$ & 2113&${\scriptscriptstyle(2334)}$ & 9529&${\scriptscriptstyle(12326)}$ & 40822&${\scriptscriptstyle(62295)}$ & \multicolumn{2}{@{}c|@{}}{4.8e5${\scriptscriptstyle(46844)}$} & 5 & /5\\\hline
${\bf f_{21}}$ & 1& & 339&${\scriptscriptstyle(360)}$ & 1886&${\scriptscriptstyle(2756)}$ & 9899&${\scriptscriptstyle(7236)}$ & \multicolumn{2}{@{}c@{}}{6.0e6${\scriptscriptstyle(2e7)}$} & \multicolumn{2}{@{}c|@{}}{6.9e6${\scriptscriptstyle(2e7)}$} & 4 & /5\\\hline
${\bf f_{22}}$ & 1& & 236&${\scriptscriptstyle(76)}$ & 3727&${\scriptscriptstyle(542)}$ & 16569&${\scriptscriptstyle(2956)}$ & \multicolumn{2}{@{}c@{}}{1.5e5${\scriptscriptstyle(13422)}$} & \multicolumn{2}{@{}c|@{}}{5.9e5${\scriptscriptstyle(2e5)}$} & 5 & /5\\\hline
${\bf f_{23}}$ & 1& & 236&${\scriptscriptstyle(104)}$ & 2311&${\scriptscriptstyle(722)}$ & 13117&${\scriptscriptstyle(9686)}$ & 94667&${\scriptscriptstyle(34764)}$ & \multicolumn{2}{@{}c|@{}}{6.2e5${\scriptscriptstyle(2e5)}$} & 5 & /5\\\hline
${\bf f_{24}}$ & 1& & 15684&${\scriptscriptstyle(20622)}$ & \multicolumn{2}{@{}c@{}}{6.6e5${\scriptscriptstyle(7e5)}$} & \multicolumn{2}{@{}c@{}}{7.7e6${\scriptscriptstyle(4e6)}$} & \multicolumn{2}{@{}c@{}}{2.7e7${\scriptscriptstyle(2e7)}$} & \multicolumn{2}{@{}c|@{}}{4.3e7${\scriptscriptstyle(5e7)}$} & 2 & /5\\\hline
${\bf f_{25}}$ & 1& & 49579&${\scriptscriptstyle(42675)}$ & \multicolumn{2}{@{}c@{}}{6.9e5${\scriptscriptstyle(5e5)}$} & \multicolumn{2}{@{}c@{}}{7.1e6${\scriptscriptstyle(6e6)}$} & \multicolumn{2}{@{}c@{}}{1.0e7${\scriptscriptstyle(6e6)}$} & \multicolumn{2}{@{}c|@{}}{1.1e7${\scriptscriptstyle(8e6)}$} & 5 & /5\\\hline
${\bf f_{26}}$ & 1& & 107&${\scriptscriptstyle(110)}$ & 766&${\scriptscriptstyle(844)}$ & 2602&${\scriptscriptstyle(322)}$ & \multicolumn{2}{@{}c@{}}{1.9e5${\scriptscriptstyle(21400)}$} & \multicolumn{2}{@{}c|@{}}{8.0e6${\scriptscriptstyle(7e6)}$} & 4 & /5\\\hline
${\bf f_{27}}$ & 1& & \multicolumn{2}{@{}c@{}}{2.1e5${\scriptscriptstyle(18317)}$} & \multicolumn{2}{@{}c@{}}{1.5e6${\scriptscriptstyle(3e6)}$} & \multicolumn{2}{@{}c@{}}{5.4e6${\scriptscriptstyle(6e6)}$} & \multicolumn{2}{@{}c@{}}{7.3e6${\scriptscriptstyle(2e7)}$} & \multicolumn{2}{@{}c|@{}}{8.0e6${\scriptscriptstyle(1e7)}$} & 4 & /5\\\hline
${\bf f_{28}}$ & 1& & 56&${\scriptscriptstyle(26)}$ & 353&${\scriptscriptstyle(246)}$ & 7009&${\scriptscriptstyle(4906)}$ & \multicolumn{2}{@{}c@{}}{2.9e5${\scriptscriptstyle(3e5)}$} & \multicolumn{2}{@{}c|@{}}{1.0e7${\scriptscriptstyle(2e7)}$} & 4 & /5\\\hline
${\bf f_{29}}$ & 1& & 407&${\scriptscriptstyle(464)}$ & 4107&${\scriptscriptstyle(426)}$ & 28043&${\scriptscriptstyle(5576)}$ & \multicolumn{2}{@{}c@{}}{2.1e5${\scriptscriptstyle(36604)}$} & \multicolumn{2}{@{}c|@{}}{1.4e6${\scriptscriptstyle(6e5)}$} & 5 & /5\\\hline
${\bf f_{30}}$ & 1& & 117&${\scriptscriptstyle(40)}$ & 1880&${\scriptscriptstyle(860)}$ & 11138&${\scriptscriptstyle(4617)}$ & \multicolumn{2}{@{}c@{}}{1.7e5${\scriptscriptstyle(1e5)}$} & \multicolumn{2}{@{}c|@{}}{8.2e6${\scriptscriptstyle(9e6)}$} & 4 & /5\\\hline
${\bf f_{31}}$ & 1& & 28458&${\scriptscriptstyle(10950)}$ & \multicolumn{2}{@{}c@{}}{4.6e5${\scriptscriptstyle(2e5)}$} & \multicolumn{2}{@{}c@{}}{1.8e7${\scriptscriptstyle(4e7)}$} & \multicolumn{2}{@{}c@{}}{2.5e7${\scriptscriptstyle(5e7)}$} & \multicolumn{2}{@{}c|@{}}{4.0e7${\scriptscriptstyle(6e7)}$} & 2 & /5\\\hline
${\bf f_{32}}$ & 1& & 32778&${\scriptscriptstyle(21806)}$ & \multicolumn{2}{@{}c@{}}{4.7e5${\scriptscriptstyle(3e5)}$} & \multicolumn{2}{@{}c@{}}{6.7e6${\scriptscriptstyle(2e6)}$} & \multicolumn{2}{@{}c@{}}{1.6e7${\scriptscriptstyle(6e6)}$} & \multicolumn{2}{@{}c|@{}}{2.1e7${\scriptscriptstyle(2e7)}$} & 4 & /5\\\hline
${\bf f_{33}}$ & 1& & 29&${\scriptscriptstyle(7)}$ & 212&${\scriptscriptstyle(18)}$ & 3520&${\scriptscriptstyle(1271)}$ & \multicolumn{2}{@{}c@{}}{5.7e5${\scriptscriptstyle(6e5)}$} & \multicolumn{2}{@{}c|@{}}{1.7e7${\scriptscriptstyle(7e6)}$} & 4 & /5\\\hline
${\bf f_{34}}$ & 1& & 540&${\scriptscriptstyle(750)}$ & \multicolumn{2}{@{}c@{}}{7.5e6${\scriptscriptstyle(7e6)}$} & \multicolumn{2}{@{}c@{}}{9.3e6${\scriptscriptstyle(2e7)}$} & \multicolumn{2}{@{}c@{}}{9.6e6${\scriptscriptstyle(1e7)}$} & \multicolumn{2}{@{}c|@{}}{2.1e7${\scriptscriptstyle(3e7)}$} & 3 & /5\\\hline
${\bf f_{35}}$ & 1& & 410&${\scriptscriptstyle(165)}$ & 4748&${\scriptscriptstyle(846)}$ & 41160&${\scriptscriptstyle(7250)}$ & \multicolumn{2}{@{}c@{}}{2.4e5${\scriptscriptstyle(43336)}$} & \multicolumn{2}{@{}c|@{}}{7.7e5${\scriptscriptstyle(5e5)}$} & 5 & /5\\\hline
${\bf f_{36}}$ & 1& & 569&${\scriptscriptstyle(335)}$ & 4542&${\scriptscriptstyle(1043)}$ & 34402&${\scriptscriptstyle(2092)}$ & \multicolumn{2}{@{}c@{}}{2.8e5${\scriptscriptstyle(1e5)}$} & \multicolumn{2}{@{}c|@{}}{1.5e6${\scriptscriptstyle(2e6)}$} & 5 & /5\\\hline
${\bf f_{37}}$ & 1& & 7210&${\scriptscriptstyle(6640)}$ & \multicolumn{2}{@{}c@{}}{4.7e5${\scriptscriptstyle(3e5)}$} & \multicolumn{2}{@{}c@{}}{1.1e7${\scriptscriptstyle(8e6)}$} & \multicolumn{2}{@{}c@{}}{1.0e8${\scriptscriptstyle(6e7)}$} & \multicolumn{2}{@{}c|@{}}{1.0e8${\scriptscriptstyle(9e7)}$} & 1 & /5\\\hline
${\bf f_{38}}$ & 1& & 84291&${\scriptscriptstyle(79832)}$ & \multicolumn{2}{@{}c@{}}{1.5e6${\scriptscriptstyle(6e5)}$} & \multicolumn{2}{@{}c@{}}{1.1e7${\scriptscriptstyle(5e6)}$} & \multicolumn{2}{@{}c@{}}{2.7e7${\scriptscriptstyle(4e7)}$} & \multicolumn{2}{@{}c|@{}}{2.7e7${\scriptscriptstyle(1e7)}$} & 3 & /5\\\hline
${\bf f_{39}}$ & 1& & 488&${\scriptscriptstyle(276)}$ & 4413&${\scriptscriptstyle(1142)}$ & 27408&${\scriptscriptstyle(11268)}$ & \multicolumn{2}{@{}c@{}}{1.6e5${\scriptscriptstyle(23316)}$} & \multicolumn{2}{@{}c|@{}}{9.1e5${\scriptscriptstyle(4e5)}$} & 5 & /5\\\hline
${\bf f_{40}}$ & 1& & 42097&${\scriptscriptstyle(62674)}$ & \multicolumn{2}{@{}c@{}}{6.4e6${\scriptscriptstyle(1e7)}$} & \multicolumn{2}{@{}c@{}}{1.7e7${\scriptscriptstyle(6e6)}$} & \multicolumn{2}{@{}c@{}}{1.7e7${\scriptscriptstyle(2e7)}$} & \multicolumn{2}{@{}c|@{}}{1.7e7${\scriptscriptstyle(2e7)}$} & 3 & /5\\\hline
${\bf f_{41}}$ & 1& & 285&${\scriptscriptstyle(252)}$ & 2579&${\scriptscriptstyle(1007)}$ & 16752&${\scriptscriptstyle(5982)}$ & \multicolumn{2}{@{}c@{}}{1.2e5${\scriptscriptstyle(42772)}$} & \multicolumn{2}{@{}c|@{}}{6.3e5${\scriptscriptstyle(4e5)}$} & 5 & /5\\\hline
${\bf f_{42}}$ & 1& & 48328&${\scriptscriptstyle(59068)}$ & \multicolumn{2}{@{}c@{}}{2.9e6${\scriptscriptstyle(3e6)}$} & \multicolumn{2}{@{}c@{}}{4.3e7${\scriptscriptstyle(1e7)}$} & \multicolumn{2}{@{}c@{}}{4.6e7${\scriptscriptstyle(8e7)}$} & \multicolumn{2}{@{}c|@{}}{4.6e7${\scriptscriptstyle(5e7)}$} & 2 & /5\\\hline
${\bf f_{43}}$ & 1& & 96614&${\scriptscriptstyle(71367)}$ & \multicolumn{2}{@{}c@{}}{2.0e6${\scriptscriptstyle(1e6)}$} & \multicolumn{2}{@{}c@{}}{1.2e7${\scriptscriptstyle(5e6)}$} & \multicolumn{2}{@{}c@{}}{2.2e7${\scriptscriptstyle(9e6)}$} & \multicolumn{2}{@{}c|@{}}{2.3e7${\scriptscriptstyle(1e7)}$} & 4 & /5\\\hline
${\bf f_{44}}$ & 1& & 171&${\scriptscriptstyle(48)}$ & 1490&${\scriptscriptstyle(394)}$ & 6162&${\scriptscriptstyle(1838)}$ & 40596&${\scriptscriptstyle(13032)}$ & \multicolumn{2}{@{}c|@{}}{7.4e5${\scriptscriptstyle(5e5)}$} & 5 & /5\\\hline
${\bf f_{45}}$ & 1& & 54016&${\scriptscriptstyle(80498)}$ & \multicolumn{2}{@{}c@{}}{1.4e6${\scriptscriptstyle(9e5)}$} & \multicolumn{2}{@{}c@{}}{1.0e7${\scriptscriptstyle(1e7)}$} & \multicolumn{2}{@{}c@{}}{2.1e7${\scriptscriptstyle(2e7)}$} & \multicolumn{2}{@{}c|@{}}{2.1e7${\scriptscriptstyle(2e7)}$} & 3 & /5\\\hline
${\bf f_{46}}$ & 1& & 48335&${\scriptscriptstyle(21389)}$ & \multicolumn{2}{@{}c@{}}{7.4e5${\scriptscriptstyle(3e5)}$} & \multicolumn{2}{@{}c@{}}{1.2e7${\scriptscriptstyle(3e7)}$} & \multicolumn{2}{@{}c@{}}{1.7e7${\scriptscriptstyle(1e7)}$} & \multicolumn{2}{@{}c|@{}}{1.9e7${\scriptscriptstyle(1e7)}$} & 4 & /5\\\hline
${\bf f_{47}}$ & 1& & 60949&${\scriptscriptstyle(27074)}$ & \multicolumn{2}{@{}c@{}}{1.1e6${\scriptscriptstyle(4e5)}$} & \multicolumn{2}{@{}c@{}}{7.2e6${\scriptscriptstyle(6e6)}$} & \multicolumn{2}{@{}c@{}}{8.7e6${\scriptscriptstyle(5e6)}$} & \multicolumn{2}{@{}c|@{}}{9.2e6${\scriptscriptstyle(7e6)}$} & 5 & /5\\\hline
${\bf f_{48}}$ & 1& & 37048&${\scriptscriptstyle(16124)}$ & \multicolumn{2}{@{}c@{}}{3.8e5${\scriptscriptstyle(2e5)}$} & \multicolumn{2}{@{}c@{}}{6.9e6${\scriptscriptstyle(8e6)}$} & \multicolumn{2}{@{}c@{}}{2.4e7${\scriptscriptstyle(2e7)}$} & \multicolumn{2}{@{}c|@{}}{2.5e7${\scriptscriptstyle(3e7)}$} & 3 & /5\\\hline
${\bf f_{49}}$ & 1& & \multicolumn{2}{@{}c@{}}{2.5e5${\scriptscriptstyle(2e5)}$} & \multicolumn{2}{@{}c@{}}{1.9e7${\scriptscriptstyle(5e7)}$} & \multicolumn{2}{@{}c@{}}{$\infty$} & \multicolumn{2}{@{}c@{}}{$\infty$} & \multicolumn{2}{@{}c|@{}}{$\infty$\textit{2.3e7}} & 0 & /5\\\hline
${\bf f_{50}}$ & 1& & 71298&${\scriptscriptstyle(36105)}$ & \multicolumn{2}{@{}c@{}}{6.3e5${\scriptscriptstyle(2e5)}$} & \multicolumn{2}{@{}c@{}}{5.0e6${\scriptscriptstyle(1e6)}$} & \multicolumn{2}{@{}c@{}}{7.3e6${\scriptscriptstyle(4e6)}$} & \multicolumn{2}{@{}c|@{}}{7.4e6${\scriptscriptstyle(4e6)}$} & 5 & /5\\\hline
${\bf f_{51}}$ & 1& & 46809&${\scriptscriptstyle(31196)}$ & \multicolumn{2}{@{}c@{}}{7.1e5${\scriptscriptstyle(5e5)}$} & \multicolumn{2}{@{}c@{}}{6.4e6${\scriptscriptstyle(6e6)}$} & \multicolumn{2}{@{}c@{}}{1.5e7${\scriptscriptstyle(1e7)}$} & \multicolumn{2}{@{}c|@{}}{1.6e7${\scriptscriptstyle(4e6)}$} & 4 & /5\\\hline
${\bf f_{52}}$ & 1& & 87993&${\scriptscriptstyle(97027)}$ & \multicolumn{2}{@{}c@{}}{4.4e6${\scriptscriptstyle(3e6)}$} & \multicolumn{2}{@{}c@{}}{2.6e7${\scriptscriptstyle(3e7)}$} & \multicolumn{2}{@{}c@{}}{2.7e7${\scriptscriptstyle(3e7)}$} & \multicolumn{2}{@{}c|@{}}{2.7e7${\scriptscriptstyle(5e7)}$} & 3 & /5\\\hline
${\bf f_{53}}$ & 1& & 29&${\scriptscriptstyle(8)}$ & 160&${\scriptscriptstyle(21)}$ & 1905&${\scriptscriptstyle(986)}$ & 17127&${\scriptscriptstyle(12449)}$ & \multicolumn{2}{@{}c|@{}}{1.9e7${\scriptscriptstyle(3e7)}$} & 3 & /5\\\hline
${\bf f_{54}}$ & 1& & 5005&${\scriptscriptstyle(5875)}$ & \multicolumn{2}{@{}c@{}}{7.4e5${\scriptscriptstyle(8e5)}$} & \multicolumn{2}{@{}c@{}}{1.5e7${\scriptscriptstyle(1e7)}$} & \multicolumn{2}{@{}c@{}}{2.2e7${\scriptscriptstyle(2e7)}$} & \multicolumn{2}{@{}c|@{}}{4.2e7${\scriptscriptstyle(5e7)}$} & 2 & /5\\\hline
${\bf f_{55}}$ & 1& & \multicolumn{2}{@{}c@{}}{2.3e5${\scriptscriptstyle(2e5)}$} & \multicolumn{2}{@{}c@{}}{6.4e6${\scriptscriptstyle(3e7)}$} & \multicolumn{2}{@{}c@{}}{7.0e6${\scriptscriptstyle(2e7)}$} & \multicolumn{2}{@{}c@{}}{7.1e6${\scriptscriptstyle(1e7)}$} & \multicolumn{2}{@{}c|@{}}{7.1e6${\scriptscriptstyle(2e7)}$} & 4 & /5
\end{tabular}}}%
\caption[Table of aRTs]{\label{tab:aRTs}\bbobpptablecaption{} 
}
\end{sidewaystable*}

%%%%%%%%%%%%%%%%%%%%%%%%%%%%%%%%%%%%%%%%%%%%%%%%%%%%%%%%%%%%%%%%%%%%%%%%%%%%%%%
% REFERENCES

% You must have a proper ".bib" file and remember to run:
% to create the ~.bbl file.  Insert that ~.bbl file into
% the .tex source file and comment out
% the command \texttt{{\char'134}thebibliography}.
%
% ACM needs 'a single self-contained file'!
%
\clearpage % otherwise the last figure might be missing

% Please uncomment for final version to fit paper to 8 pages.
%\end{document}

% Please uncomment for final version to fit paper to 8 pages.
%\end{document}

%%%%%%%%%%%%%%%%%%%%%%%%%%%%%%%%%%%%%%%%%%%%%%%%%%%%%%%%%%%%%%%%%%%%%%%%%%%%%%%
%%%%%%%%%%%%%%%%%%%%%%%%%%%%%%%%%%%%%%%%%%%%%%%%%%%%%%%%%%%%%%%%%%%%%%%%%%%%%%%

% Scaling of aRT with dimension

\end{document}